\documentclass[14pt]{article}
\usepackage{graphicx}
\usepackage{titlesec}
\usepackage[hyperfootnotes=true]{hyperref}
\hypersetup{colorlinks, citecolor=blue, linkcolor=blue, urlcolor=blue}
\usepackage{makecell}

\usepackage{hyperref}%
\usepackage{color,soul}
\usepackage{amsmath}
\usepackage{tcolorbox}
\usepackage{makecell}
\usepackage{multirow} 
\usepackage{booktabs}
\usepackage{amsmath}
\usepackage{caption}
\usepackage{svg}

\usepackage{geometry}
\graphicspath{{figs/}}
\usepackage{caption}
\usepackage[utf8]{inputenc}
\usepackage[numbers]{natbib}
\usepackage{listings}
\usepackage{minted}
\usepackage{xcolor}
\usepackage{amssymb}
\usepackage{pifont}

\usepackage{graphicx}
\usepackage{subcaption}

\usepackage[british]{babel}
\usepackage{hhline}
\usepackage{multirow}
\usepackage{authblk}
\usepackage{soulpos}

\definecolor{bg}{rgb}{0.95,0.95,0.95}
\definecolor{codegray}{rgb}{0.4,0.4,0.4}
\definecolor{codepurple}{rgb}{0.58,0,0.82}
\title{\textbf{In Transformer We Trust? \\ A Perspective on Transformer Architecture Failure Modes}}

\author[]{Trishit Mondal}
\author[]{Ameya D. Jagtap\thanks{Corresponding author: Ameya D. Jagtap (ajagtap@wpi.edu, ameyadjagtap@gmail.com)}}

\affil[]{\textit{\small{Aerospace Engineering Department, Worcester Polytechnic Institute, Worcester, MA 01609, USA.}}}
\date{}

\begin{document}

\maketitle
\begin{abstract}Transformer architectures have revolutionized machine learning across a wide range of domains, from natural language processing to scientific computing. However, their growing deployment in high-stakes applications, such as computer vision, natural language processing, healthcare, autonomous systems, and critical areas of scientific computing including climate modeling, materials discovery, drug discovery, nuclear science, and robotics, necessitates a deeper and more rigorous understanding of their trustworthiness. In this work, we critically examine the foundational question: \textit{How trustworthy are transformer models?} We evaluate their reliability through a comprehensive review of interpretability, explainability, robustness against adversarial attacks, fairness, and privacy. We systematically examine the trustworthiness of transformer-based models in safety-critical applications spanning natural language processing, computer vision, and science and engineering domains, including robotics, medicine, earth sciences, materials science, fluid dynamics, nuclear science, and automated theorem proving; highlighting high-impact areas where these architectures are central and analyzing the risks associated with their deployment. By synthesizing insights across these diverse areas, we identify recurring structural vulnerabilities, domain-specific risks, and open research challenges that limit the reliable deployment of transformers. %
\end{abstract}

\vspace{0.2cm}

 \begin{small}Keywords: \textit{Transformer Architecture}; \textit{Transformer Trustworthiness}; \textit{Safety-critical AI}
\end{small}

\tableofcontents
\section{Introduction}
In all scientific and technological advancement, history shows a recurring pattern: the emergence of engineering artifacts long before the formalization of the scientific principles that govern them. From ancient water wheels and Roman concrete to early steam engines and flight, researchers and engineers often developed functional systems through empirical knowledge, intuition, and iterative experimentation, only later did scientists construct the theoretical frameworks that explained their efficacy. Today, artificial intelligence (AI), and in particular transformer-based architectures, stands at a similar crossroads. Transformer models have become the cornerstone of modern deep learning, achieving groundbreaking results across natural language processing (NLP), computer vision, code generation, and even scientific discovery. Yet, despite their widespread success, transformers largely remain black-box artifacts: effective but opaque, powerful yet poorly understood. This disjunction raises a pressing question: how can we move from merely building and using these models to truly understanding and trusting them? Transformers were first introduced in 2017 by Vaswani et al. \cite{vaswani2017attention} with the seminal paper Attention Is All You Need, which reimagined sequence modeling using a self-attention mechanism. Since then, their adoption has accelerated rapidly, leading to architectures such as BERT \cite{devlin2019bert}, GPT \cite{brown2020language}, ViT \cite{dosovitskiy2020image}, and many others. These models have achieved superhuman performance on benchmark datasets, led to dramatic improvements in machine translation \cite{wu2016google}, content generation \cite{liu2021deep}, summarization \cite{adhikari2020nlp}, protein folding \cite{jumper2021highly} , and even theorem proving \cite{trinh2024solving}. Yet, their success has outpaced our ability to explain or guarantee their behavior. Many foundational questions remain unanswered: What inductive biases do transformers encode? Why are they so adept at in-context learning? How do emergent behaviors arise as model size scales? Under what conditions do they fail, and how can such failures be predicted or mitigated? The lack of transparency and theoretical grounding has significant consequences, especially as these models are deployed in real-world settings where trust is important. Transformers are being used to summarize legal documents, assist in medical diagnoses, provide financial advice, and control interactive agents in virtual and physical environments. In these contexts, performance alone is insufficient. Users demand \textit{trustworthiness}, a broad umbrella that includes robustness to adversarial inputs, resistance to spurious correlations, interpretability of decisions, calibration of uncertainty, fairness across demographic groups, and alignment with human values. A transformer that outputs impressive text but hallucinates factual content, or one that encodes and perpetuates harmful social biases, cannot be considered trustworthy, regardless of its benchmark scores. In the absence of rigorous theory, much of current practice relies on heuristics: prompt engineering, fine-tuning strategies, architectural tweaks, and massive scaling. These approaches, though often effective, leave practitioners vulnerable to failure modes that are poorly characterized and difficult to anticipate. For example, recent work has shown that transformers can be highly brittle under small perturbations \cite{neerudu2023robustness}, can generate toxic or biased content even after safety fine-tuning \cite{gehman2020realtoxicityprompts}, and can be susceptible to training data leakage or inversion attacks \cite{li2024seeing}. Without a principled understanding of why these behaviors arise, efforts to mitigate them remain ad hoc and reactive.
\begin{figure}[htpb]
    \centering
\includegraphics[scale=0.2,clip=true]{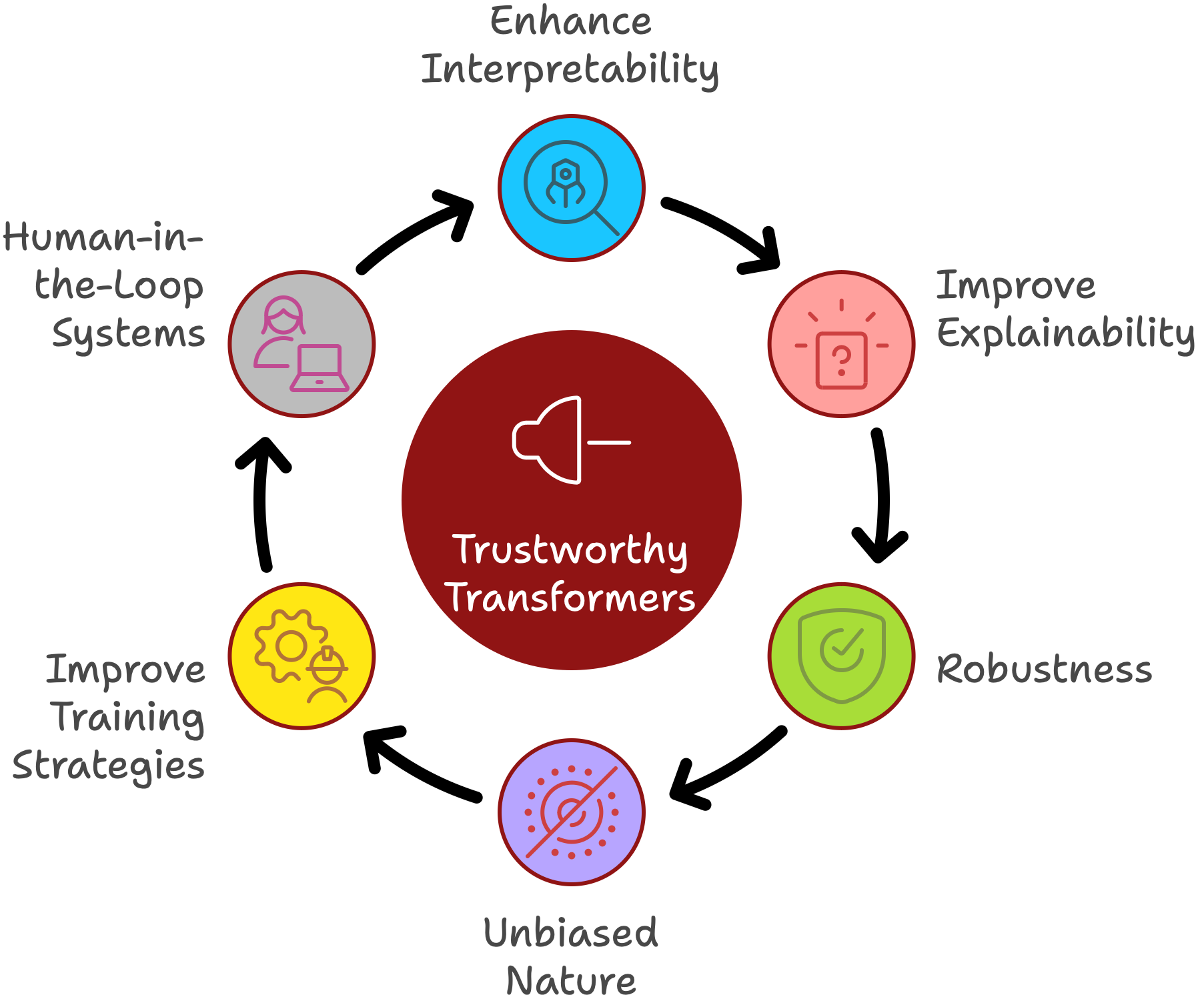} 
    \caption{Key dimensions of trustworthy transformer architectures.}
    \label{fig:trustTrans}
\end{figure}

In this work, we examine the key dimensions of trustworthy transformer architectures, as illustrated in Fig. \ref{fig:trustTrans}. These dimensions encompass explainability and interpretability, robustness against adversarial attacks, mitigation of bias to ensure fairness, and improved training strategies that enhance transparency and model understanding. Moreover, we highlight the role of human-in-the-loop frameworks for oversight and accountability, collectively fostering the development of reliable, ethical, and trustworthy transformer-based systems. Several promising research directions are emerging to enhance the trustworthiness of Transformer architectures. Information-theoretic \cite{clark2019does} and geometric frameworks \cite{marks2023geometry} aim to demystify attention mechanisms and the geometry of high-dimensional representations, while mechanistic interpretability \cite{rai2024practical} seeks to decompose network behaviors into discernible computational circuits and patterns. Research on robustness investigates model responses under adversarial attacks and distributional shifts, and causal inference approaches attempt to distinguish causality from correlation within learned representations. Aligning model behavior with human intent and values has driven advances in reinforcement learning from human feedback (RLHF) \cite{lambert2025reinforcement}, constitutional AI \cite{bai2022constitutional}, and cooperative AI \cite{dafoe2021cooperative}.

Trustworthiness is recognized as a multi-layered concept. Huang et al. \cite{huang2024trustllm} argue that models must not only provide correct outputs but also behave fairly, securely, robustly, and transparently, emphasizing ethical consistency across unpredictable situations. Complementing this, Chen et al. \cite{chen2024trustworthy} situate trustworthiness within a broader AI safety framework, highlighting technical correctness, operational reliability, and systemic risks arising from complex, interdependent AI components. Ferdaus et al. \cite{ferdaus2026towards} further show that traditional static benchmarks are insufficient for evaluating LLMs in practice. Their study shows important weaknesses, including high sensitivity to prompt variations and bias in human or automated evaluators. Security considerations, including software- and hardware-level vulnerabilities, are discussed by Latibari et al. \cite{latibari2024transformers}, while Tadi et al. \cite{tadi2025trustformer} stress privacy, governance, and user-rights challenges inherent in large-scale, distributed Transformer training. Collectively, these perspectives converge on a key insight: trustworthiness in Transformers is multi-dimensional, encompassing ethics, fairness, robustness, transparency, and security, and is essential for their safe and responsible deployment. Against this backdrop, our work connects high-level safety ideas with practical, domain-specific engineering by expanding the analysis beyond natural language to include Computer Vision, Robotics, and, importantly, Scientific Machine Learning. We argue that in scientific applications, trustworthiness requires more than accurate predictions. It demands strict adherence to physical conservation laws and reliable uncertainty estimation. To support this, we highlight concrete architectural solutions, such as neuro-symbolic integration for verifiable reasoning and Bayesian-based uncertainty quantification, that are critical for dependable decision-making in high-stakes scientific settings.

This Perspective is organized to progressively examine the foundations and broader implications of transformer architectures. Section~2 explores interpretability and explainability, highlighting emerging tools for understanding model behavior. Section~3 investigates robustness under distributional shifts and adversarial conditions, followed by Section~4, which discusses fairness considerations and strategies for bias mitigation. In Section~5, we examine trustworthiness within scientific machine learning, emphasizing reliability of the predicted solution. Section~6 discuss additional metrics for assessing trustworthy AI systems. Safety-critical applications are analyzed in depth in Section~7, highlighting practical deployment challenges. Section~8 offers a forward-looking overview of Agentic AI and its implications. We conclude in Section~9 by synthesizing these themes and outlining key insights and future directions.

\section{Interpretability and Explainability in Transformers}

Transformers have entirely revolutionized the machine learning landscape, particularly across applications such as NLP, computer vision, and even scientific computing. From driving LLM models \cite{brown2020language} and language translators \cite{conneau2020unsupervised} to enabling complex image recognition systems \cite{dosovitskiy2020image}, these models are now the foundational component of several cutting-edge AI applications. Their capacity for handling sequential data, learning contextual relationships, and scaling up to large datasets has made them remarkably useful and pervasive. Yet, notwithstanding their impressive performance, transformer architectures retain a fundamental limitation: they largely operate as black boxes. Once trained, the precise mechanisms by which they arrive at specific predictions are often opaque. Comprising stacked layers of attention mechanisms and dense nonlinear transformations, their architectural depth and mathematical complexity, while enabling exceptional expressive capacity, make it difficult to trace internal reasoning or to attribute influence to particular inputs. This opacity becomes a critical issue when Transformers are deployed in sensitive or high-risk domains like scientific or medical systems, where trusting and comprehending a model's decision is equally critical as its accuracy \cite{computers13040092}. This is where the discipline of interpretability and explainability is crucial. Interpretability is concerned with making the inner mechanisms of a model clearer, such as revealing what different attention heads do or how data passes between the layers. Explainability is all about explaining why a model took a particular decision in a manner that makes sense to human beings, even if they are not technical individuals. Much work has been done in the last few years to create tools and techniques to investigate Transformers, to uncover where in the input the model is paying attention, how it is processing that information, and how it is transforming it into an end output. These efforts aim to not only increase transparency but also to establish trust, improve debugging, and guaranteeing ethical use of AI systems. As Transformers continue to grow in popularity and complexity, the importance of understanding their inner workings will only become more critical \cite{rai2024practical}.

\subsection{Attention Visualization}
The interpretation of Transformer models has traditionally centered on the attention mechanism, under the intuitive assumption that attention weights serve as a proxy for input feature importance. Consequently, visualizing these attention maps became a standard for explaining model behavior. However, this premise that \textit{attention equals explanation} has been rigorously scrutinized, shifting the field from passive visualization to a critical evaluation of causal reliability. A pivotal critique was introduced by Jain et al. \cite{jain2019attention}, who argued that for attention to be a faithful explanation, it must be both consistent with feature importance measures and exclusive. Through extensive experiments across multiple NLP tasks, the authors demonstrated a systemic failure on both counts. First, they observed that learned attention weights often poorly align with established feature importance metrics, such as gradient-based attribution, suggesting that what the model looks at is not necessarily what it uses to make a decision. More distinctively, the study introduced the concept of \textit{adversarial attention}. Jain et al. successfully constructed alternative attention distributions that maximized the divergence from the original weights, effectively shifting focus to entirely different input tokens while inducing virtually no change in the model's output prediction. This counterfactual analysis revealed that standard attention weights are rarely unique; rather, they represent just one of many possible configurations that could result in the same decision. This disconnect fundamentally undermines the reliability of raw attention visualization as a faithful causal explanation for decision-making. 

\begin{figure}[htpb]
    \centering
\includegraphics[scale=0.35,clip=true]{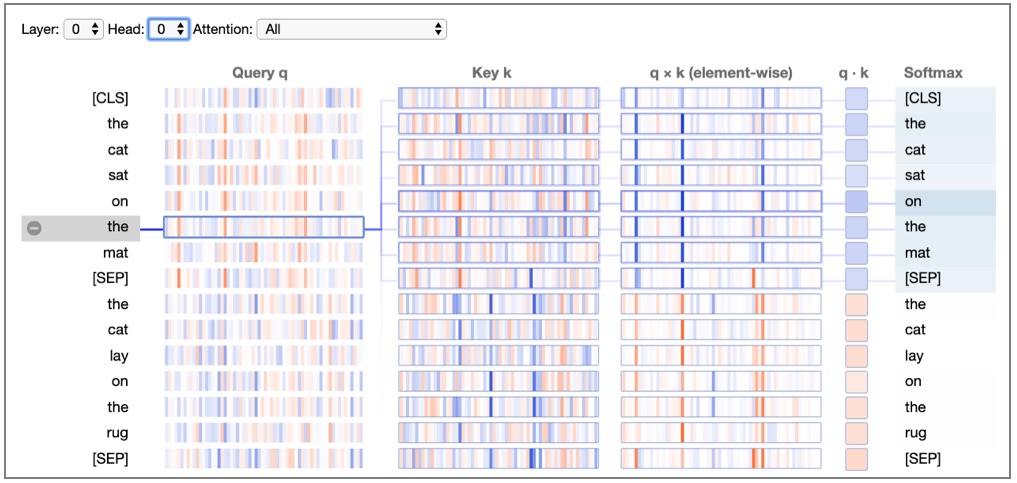} 
    \caption{Visualization of BERT’s neuron activations for layer 0, head 0. Positive and negative values are represented using blue and orange hues, respectively, with color intensity indicating the magnitude. Similar to the attention-head view, the connecting lines reflect attention strength between words, with line thickness proportional to the attention weights. Adapted from Vig \cite{vig2019multiscale}}
    \label{fig:vig}
\end{figure}

Despite ongoing debates about their causal validity, attention weights are still widely used as a practical tool for debugging and analyzing model behavior. To manage the complexity of multi-layer, multi-head Transformer architectures, Vig \cite{vig2019multiscale} introduced a structured multiscale analysis framework that breaks attention into three levels. The \textit{Attention-Head View} focuses on individual heads and reveals localized patterns such as subject–verb agreement. The \textit{Model View} provides a global matrix representation that helps identify heads with specific functions, such as paraphrase detection. Finally, the \textit{Neuron View} examines the underlying query–key interactions at the neuron level. As illustrated in Figure~\ref{fig:vig}, this hierarchical visualization connects individual neuron activations to higher-level functional behaviors, including effects similar to a sliding context window created by attention decay.

\begin{figure}[htpb]
    \centering
\includegraphics[scale=0.3,clip=true]{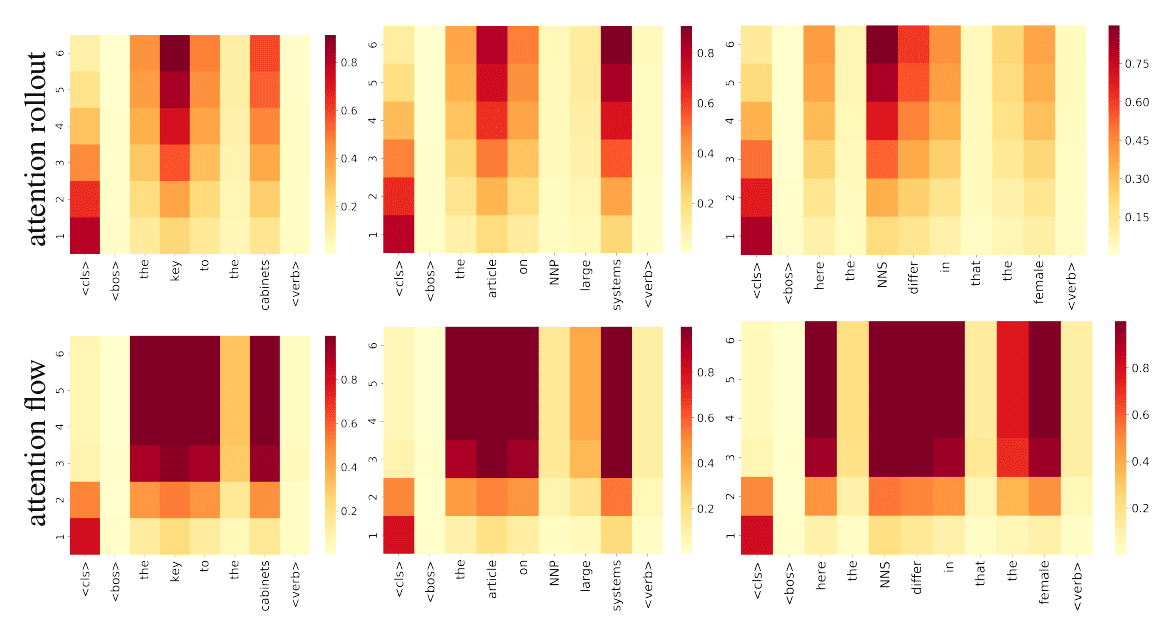} 
    \caption{Attribution maps from attention rollout (top) and attention flow (bottom) across six Transformer layers for three example sentences. Both methods highlight influential tokens (e.g., subject vs. distractors) more effectively than raw attention. Rollout produces sharper focus, while flow reveals broader influence patterns, aiding interpretability in deeper layers. Adapted from Abnar et al. \cite{abnar2020quantifying}.}
    \label{fig:attn_rollout_flow}
\end{figure}

A major limitation of visualizing raw attention is the entanglement problem: as information propagates through deeper layers, attention weights become increasingly diffuse, making it difficult to trace final predictions back to specific input tokens. To address this issue, Abnar et al. \cite{abnar2020quantifying} proposed two methods for reconstructing the flow of information: \textit{Attention Rollout} and \textit{Attention Flow}. Attention Rollout assumes that attention weights represent proportional mixing of token representations and computes attributions by recursively multiplying attention matrices across layers. In contrast, Attention Flow models the Transformer as a directed acyclic graph and applies a maximum-flow algorithm to identify the strongest information pathways. Comparative results show that both methods achieve higher Spearman correlation with model predictions than raw attention (Figure \ref{fig:attn_rollout_flow}). In general, Rollout yields sharper and more localized attribution maps, whereas Attention Flow captures broader, more distributed influence patterns that better reflect the model’s internal redundancy.

While attention visualization provides insight into token interaction, it does not fully explain how the content of these representations evolves layer by layer. Moving beyond attention weights, Voita et al. \cite{voita2019bottom} utilize information-theoretic measures, such as mutual information and canonical correlation analysis, to track the flow of information in models trained with different objectives: Machine Translation (MT), Language Modeling (LM), and Masked Language Modeling (MLM). Their analysis shows that the training objective fundamentally dictates the internal structure of the model. In standard LM, information about the past fades as the model focuses on predicting the future. Conversely, MLM (used in BERT) exhibits a distinct two-stage learning process: early layers encode rich contextual information while temporarily losing token identity, only to reconstruct it in the final layers. This \textit{diffuse-then-reconstruct} mechanism allows MLM-based models to generalize more effectively before finalizing their representations, explaining their superior performance in transfer learning tasks compared to standard LMs. Furthermore, the study highlights a frequency bias: representations of frequent tokens change significantly in lower layers, while rare tokens retain their influence longer, suggesting that the model prioritizes contextualizing common words early in the network.

\begin{figure}[htpb]
    \centering
    \includegraphics[scale=0.25,clip=true]{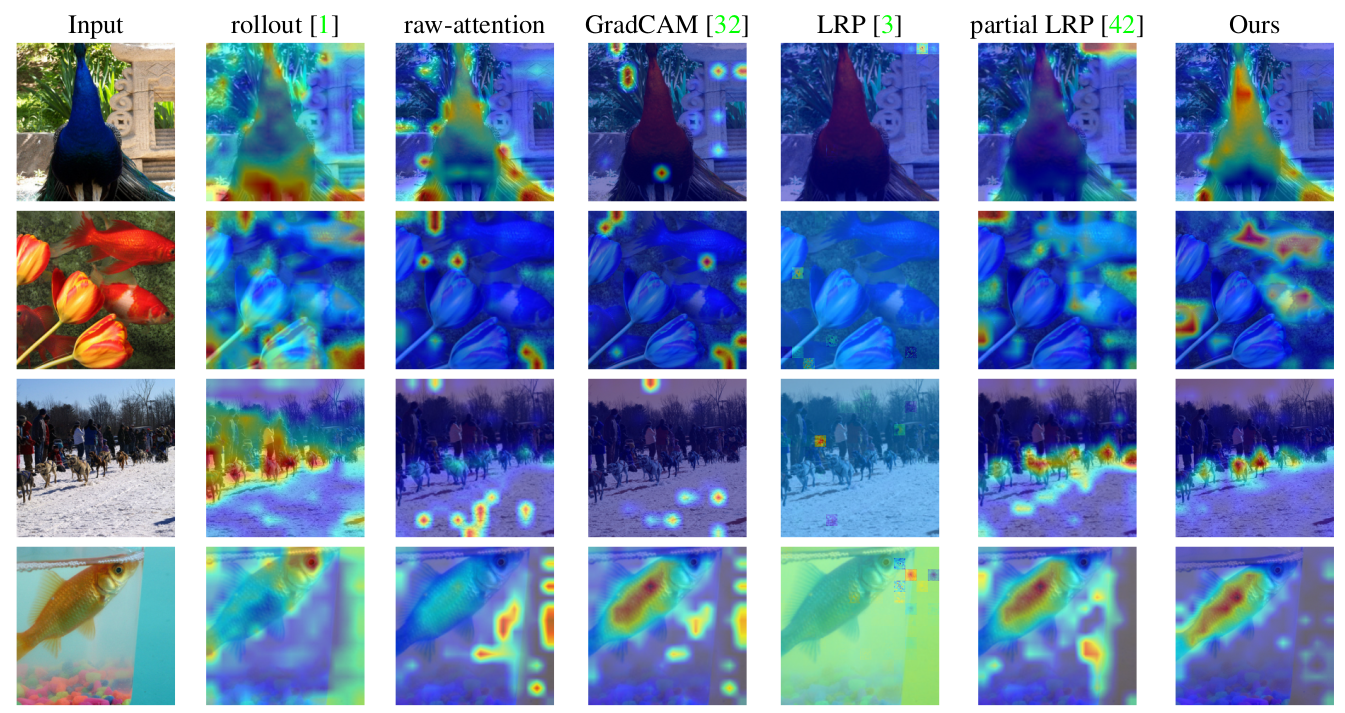} 
    \caption{Comparison of interpretability methods on visual tasks. The top row displays the input image containing multiple objects. Subsequent rows show attribution maps generated by Raw Attention, Rollout, GradCAM, and Chefer et al.'s Relevance Propagation (Ours). Note that the proposed method (bottom) successfully isolates class-specific features (e.g., distinguishing between the two dogs) with significantly less noise than heuristic methods like Rollout or GradCAM. Adapted from Chefer et al. \cite{chefer2021transformer}.}
    \label{fig:chefer_comparison}
\end{figure}

While the techniques discussed above focus largely on linguistic representations, Vision Transformers (ViTs) introduce unique challenges due to their reliance on patch-based image processing and complex non-linearities. Despite achieving state-of-the-art performance, standard ViTs remain opaque. Traditional linguistic interpretability methods, such as Attention Rollout, often fail in this domain because they ignore the contributions of skip connections and non-linear activations (e.g., GELU), which results in noisy, class-agnostic attention maps. To address this, Chefer et al. \cite{chefer2021transformer} proposed a method based on Deep Taylor Decomposition (DTD) to propagate relevance scores from the output class token back to the input layers. Unlike heuristic aggregation strategies, this approach explicitly handles skip connections and non-linear activations to maintain the \textit{conservation of relevance} throughout the network. This results in class-specific visualizations that can distinguish between multiple semantic objects in a single image. As demonstrated in Figure \ref{fig:chefer_comparison}, while methods like Rollout and GradCAM often produce noisy or diffuse highlights that cover unrelated regions (e.g., highlighting both dogs when predicting only one), the relevance propagation method cleanly segments the specific object corresponding to the target class.

Addressing interpretability at the fundamental architectural level, Qiang et al. \cite{qiang2023interpretability} propose an architectural solution: the Interpretable-Aware Vision Transformer (IA-ViT). Most existing explanation techniques for vision-based models are post hoc, applied after training, and often rely on attention weights or gradients to infer which parts of the input influenced the prediction. However, these methods have well-documented limitations, such as poor generalizability, limited faithfulness, and sensitivity to input variations. The pipeline begins with a Vision Transformer backbone that processes an input image by dividing it into patches and projecting them into embeddings through a linear projection layer, along with positional embeddings and a [CLS] token, where [CLS] is a classification token. These embeddings are passed through standard Transformer layers (multi-head attention and MLP blocks) to extract deep visual features. The predictor component uses the [CLS] token to produce the final prediction, as in conventional ViTs. In parallel, the interpreter receives all patch embeddings except the [CLS] token and applies a single-head self-attention (SSA) mechanism followed by an MLP to produce its own prediction. Its purpose is to simulate the behavior of the predictor and produce an alternative prediction based solely on the image patches. During training, the interpreter is optimized to mimic the predictor's output and generate attention maps that reveal which parts of the image were most influential in the model's decision. 

\begin{figure}[htpb]
    \centering
\includegraphics[scale=0.3,clip=true]{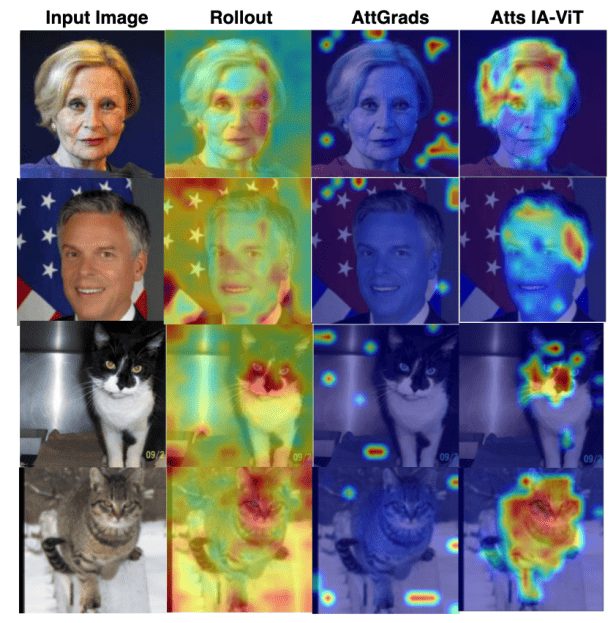} 
    \caption{Visual comparison of attribution maps generated by IA-ViT, Attention Rollout, and AttGrad methods. IA-ViT consistently highlights the most relevant image regions (e.g., the face of a cat or hair of a person) with high precision and minimal noise. In contrast, Rollout often produces diffuse attributions covering irrelevant background areas, while AttGrad occasionally emphasizes uninformative regions. Adapted from Qiang et al. \cite{qiang2023interpretability}}
    \label{fig:Qiang_result}
\end{figure}

 
These two components are trained together using a combined objective that includes (1) cross-entropy loss for prediction accuracy, (2) a knowledge distillation loss to align the interpreter’s outputs with the predictor’s, and (3) an attention regularization term to encourage similarity between the predictor’s multi-head attention and the interpreter’s single-head attention. Quantitative evaluations on datasets like CIFAR10 and CelebA show that IA-ViT significantly outperforms post-hoc methods in deletion and insertion scores. Qualitatively (Figure \ref{fig:Qiang_result}), IA-ViT generates high-precision attention maps that isolate semantic objects with minimal background noise, whereas baseline methods often produce dispersed or irrelevant attributions.

\subsection{Symbolic and Neuro-Symbolic Transformers}
Humans process information using a dual mechanism of \textit{perception} and \textit{cognition}. Perception involves converting raw sensory inputs into meaningful patterns, while cognition maps these patterns onto structured knowledge to enable abstraction, reasoning, and planning. Artificial intelligence aims to mimic this duality. While neural networks, particularly those driven by self-supervised learning, good at perception, they often function as opaque \textit{black-boxes}. Conversely, symbolic AI mirrors human cognition with structured logic for interpretable reasoning, though it often lacks scalability \cite{sheth2023neurosymbolic}. Neuro-symbolic AI bridges these paradigms, creating systems that learn from massive data while retaining the ability to reason, generalize, and explain their decisions. In the context of Transformers, this integration is typically achieved either by compressing symbolic knowledge (e.g., Knowledge Graphs) into neural embeddings or by converting neural outputs into symbolic structures for explicit reasoning. This synergy is critical for addressing failures in reasoning and out-of-distribution generalization, which represents an important path toward building trustworthy and safe AI \cite{hamilton2024neuro, zhang2024neuro}.

\begin{figure}[htpb]
    \centering
\includegraphics[scale=0.25,clip=true]{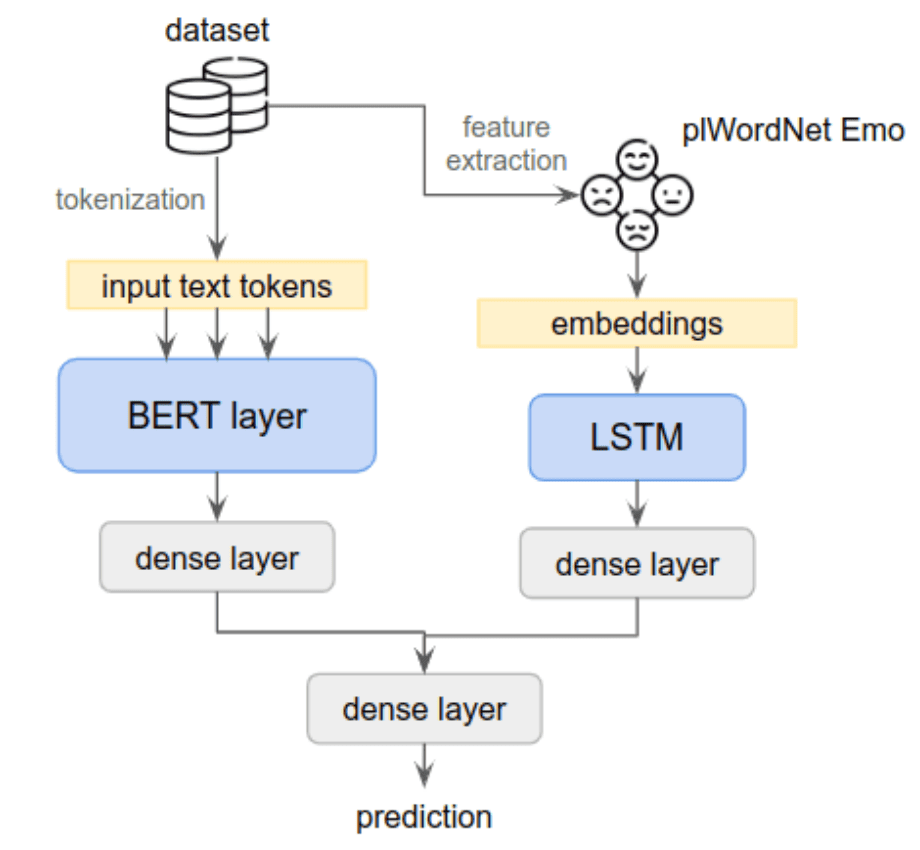} 
    \caption{HurtBERT Embedding Architecture: Sentiment features from the HurtLex lexicon are encoded using an LSTM and concatenated with contextual word embeddings from BERT. This combined representation is then used for downstream classification, enabling the model to utilize both symbolic sentiment cues and deep contextual information. Adapted from Baran et al.\cite{10031186}}
    \label{fig:Hurtbert}
\end{figure}

One effective approach to bridging the neuro-symbolic gap is to inject external symbolic knowledge directly into Transformer representations. Baran et al. \cite{10031186} explore this in the context of sentiment analysis, investigating whether deep learning models can be made more robust by grounding them in linguistic knowledge bases like WordNet and SentiWordNet. Their study contrasts two integration strategies: shallow fusion during fine-tuning versus deep fusion during pre-training. The shallow approach, exemplified by the HurtBERT architecture (Figure \ref{fig:Hurtbert}), processes lexical sentiment features using a parallel LSTM branch and concatenates them with BERT's contextual embeddings. However, experimental results indicate that this late-stage addition of knowledge often fails to improve and can even degrade performance compared to baselines. In contrast, the SentiLARE model adopts a deep fusion strategy by modifying the pre-training stage itself. It utilizes a Label-Aware Masked Language Modeling task to inject Part-of-Speech tags and polarity scores directly into RoBERTa’s weights. By including symbolic knowledge into the model's fundamental representation learning, SentiLARE consistently outperforms baselines, particularly on small-to-medium datasets. This suggests that for Transformers to truly benefit from symbolic knowledge, it must be integral to their learning objective rather than an auxiliary input. 

\begin{figure}[htpb]
    \centering
\includegraphics[scale=0.35,clip=true]{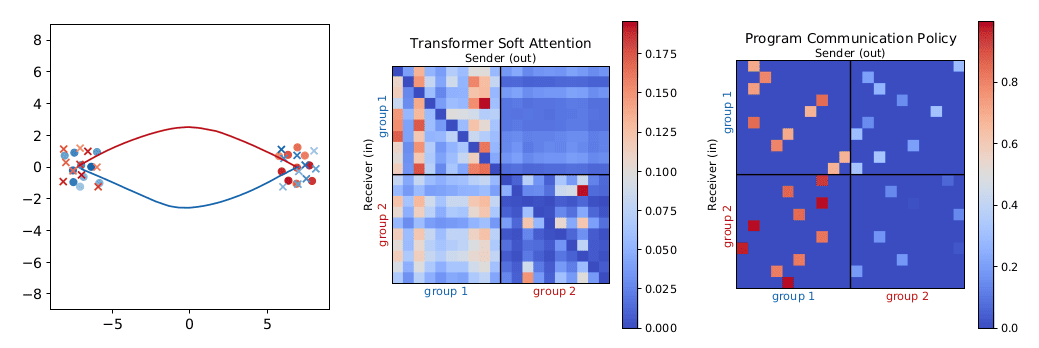} 
    \caption{Comparison of communication patterns in a two-group navigation task.
Left: Two groups of agents (blue and red) start from opposite sides and navigate toward their goals, requiring intra- and inter-group coordination to avoid collisions.
Middle: The soft attention weights from the trained transformer policy show that each agent communicates with all others, resulting in a dense communication graph.
Right: The programmatic communication policy synthesized by the neurosymbolic model produces a sparse communication graph by selecting only key agents to communicate with such as the nearest neighbor in the direction of travel and a random agent from the opposing group achieving similar coordination with reduced communication. Adapted from Inala et al. \cite{inala2020neurosymbolic}}
    \label{fig:ma_cooperative}
\end{figure}

In cooperative multi-agent systems, agents must not only make individual decisions but also coordinate effectively through communication. However, enabling every agent to communicate with all others is often infeasible due to bandwidth constraints and can lead to overfitting or poor generalization. The core challenge is to learn a communication policy that determines whom each agent should communicate with, such that the team can still achieve high task performance while minimizing the overall communication burden. Beyond static knowledge, neuro-symbolic methods can enhance trustworthiness by making the process of reasoning or coordination explicit. In cooperative multi-agent systems, Inala et al. \cite{inala2020neurosymbolic} address the challenge of efficient communication. Standard Transformer-based policies use \textit{soft attention} to weigh messages from all other agents, resulting in dense, opaque communication graphs (Figure \ref{fig:ma_cooperative}, Middle). To improve interpretability and bandwidth efficiency, the authors propose a neuro-symbolic architecture that \textit{hardens} this soft attention into discrete, programmatic rules. The system synthesizes a symbolic policy that approximates the neural attention, producing sparse and human-readable communication rules such as \textit{communicate only with the nearest neighbor in the direction of travel}. This transition from opaque weights to explicit code allows human operators to inspect and verify the emergent coordination strategy, a crucial feature for safety-critical systems. Advantage of using programmatic policies is that they offer inherent interpretability, each agent's communication decisions are governed by explicit rules that are easy to understand and analyze. For example, a learned rule might state that an agent should communicate with its nearest neighbor in a particular direction or randomly select a partner beyond a certain distance. Unlike neural attention weights, these rules can be read and reasoned about, enabling human insight into how coordination emerges among agents. This interpretability is particularly useful in safety-critical or bandwidth-constrained applications, where understanding the system's behavior is as important as achieving optimal performance. 

\begin{figure}[htpb]
    \centering
\includegraphics[scale=0.3,clip=true]{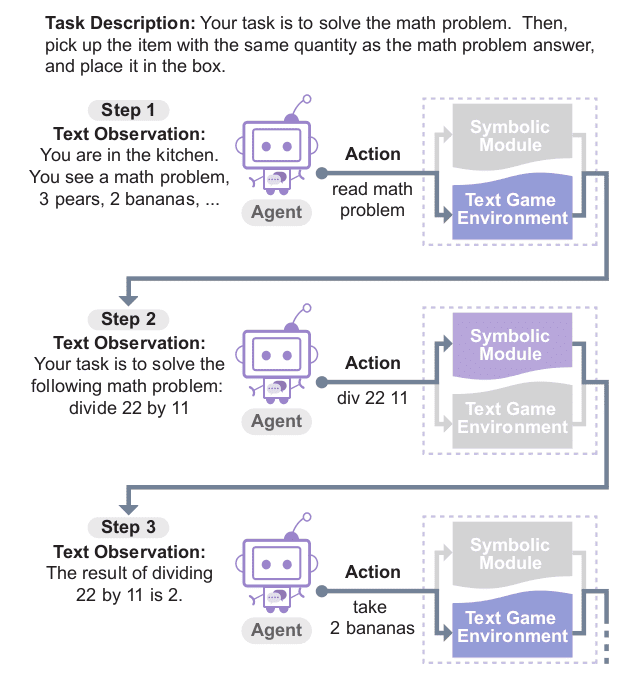} 
    \caption{An overview of the neurosymbolic reasoning architecture. At each time step, the behavior cloned transformer agent receives a textual observation from the environment and selects an action from a combined set of environment and symbolic module actions. Adapted from Wang et al. \cite{wang2022behavior}}
    \label{fig:behavior}
\end{figure}

Similarly, in the domain of interactive text-based games, Wang et al. \cite{wang2022behavior} enhance interpretability by integrating symbolic tools directly into the agent's action space. Rather than forcing a Transformer to implicitly learn arithmetic or sorting via massive scale, their architecture allows the model to explicitly invoke symbolic modules (e.g., a calculator). As shown in Figure \ref{fig:behavior}, complex tasks are decomposed into observable steps: the agent reads a problem, outputs a symbolic command (``div 22 11''), and acts on the result. This explicit tool usage renders the reasoning chain transparent, allowing observers to trace errors to specific logic failures rather than vague model hallucinations.

To this end, symbolic integration offers a mechanism to enforce logical consistency and safety in high-stakes environments like autonomous driving. Russo et al. \cite{10651426} introduce ESRA, a relation transformer for scene graph generation that incorporates Logic Tensor Networks (LTN). An important feature of ESRA is its enhanced interpretability, achieved through its neuro-symbolic architecture. Unlike traditional transformer models, ESRA integrates symbolic reasoning via LTNs to enforce logical consistency in predictions. Rather than relying solely on data-driven learning, ESRA is equipped with a set of human-understandable first-order logic constraints that encode commonsense knowledge about traffic scenarios, for example, '\textit{a person can ride a bicycle}' but '\textit{a traffic light cannot stand on a car}'. These constraints guide the model to favor semantically valid ⟨subject, predicate, object⟩ triplets during training, filtering out implausible combinations. When visual features alone are ambiguous; such as suggesting an unlikely relation like pedestrian overtaking a car, the symbolic component can downweight or correct such outputs based on prior knowledge. This fusion allows ESRA’s decisions to be traced back not only to visual attention patterns but also to the logical axioms that shaped its learning. ESRA grounds object and relation predictions in both perceptual cues and rule-based knowledge, offering a more explainable and reliable interpretation of complex traffic scenes.

\begin{figure}[htpb]
    \centering
\includegraphics[scale=0.24,clip=true]{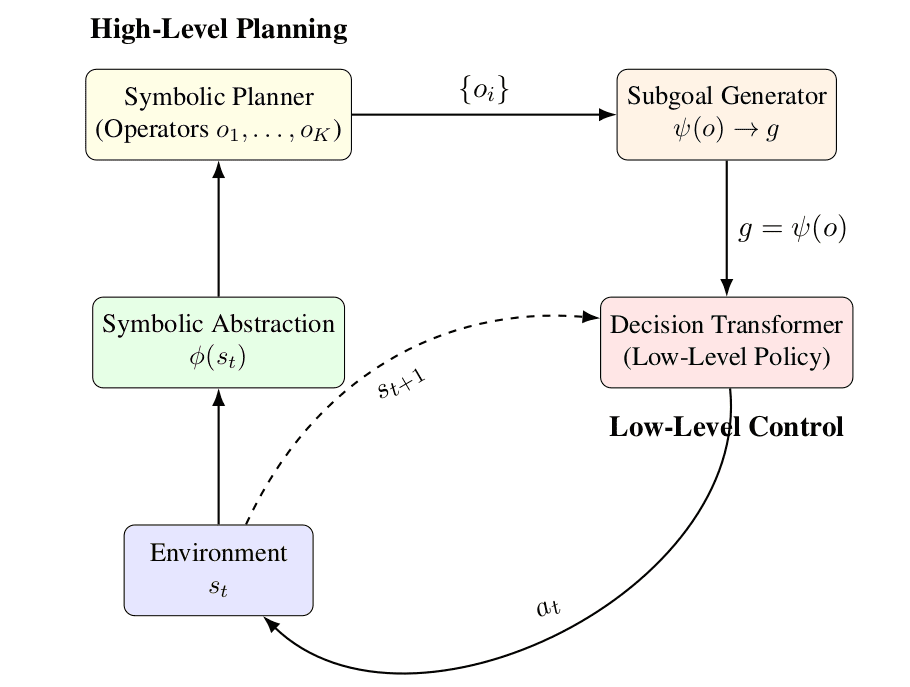} 
    \caption{Overview of the Hierarchical Neuro-Symbolic Control Architecture. The system first abstracts the environment state into symbolic predicates, which are used by a high-level symbolic planner to generate a sequence of logical operators (e.g., '\textit{PickKey}', '\textit{OpenDoor}'). Each operator is mapped to a sub-goal that conditions a low-level decision transformer. The transformer then generates fine-grained actions to achieve each sub-goal, enabling interpretable high-level planning combined with adaptive low-level control. Adapted from Ali Baheri and Cecilia O. Alm \cite{baheri2025hierarchical}}
    \label{fig:heirarchial_DT}
\end{figure}

For long-horizon tasks requiring both structure and adaptability, Baheri and Alm \cite{baheri2025hierarchical} propose a Hierarchical Neuro-Symbolic Decision Transformer. This architecture (Figure \ref{fig:heirarchial_DT}) decouples high-level planning from low-level control. A symbolic planner generates a transparent sequence of logical operators (e.g., '\textit{PickKey}', '\textit{OpenDoor}'), which are then translated into sub-goals for a neural Decision Transformer to execute. This separation of concerns ensures that the overall strategy is human-readable and verifiable, while the neural component handles the noisy, sensorimotor details of execution.

Integrating symbolic reasoning with Transformer architectures marks an important move away from purely data-driven, opaque models toward more transparent and logically grounded AI. By combining the pattern-recognition strength of neural networks with the structure of symbolic logic, neuro-symbolic approaches help address the \textit{black-box} nature of standard deep learning. These methods incorporate explicit knowledge, rule-based reasoning, or hierarchical constraints, making model behavior easier to interpret and verify. As AI systems are increasingly used in safety-critical applications, this hybrid approach provides a promising path toward decisions that are both reliable and explainable.
\begin{tcolorbox}[colback=orange!5!white,colframe=red!75!black]
\noindent Together, advances in attention visualization and neuro-symbolic integration highlight a broader shift toward building trustworthy AI. Although visualization tools and flow-based metrics help probe the internal behavior of so-called \textit{black-box} models, they also expose the limits of purely data-driven approaches. Neuro-symbolic architectures address these limits by structuring learning with explicit logic, enforcing consistency, and grounding model outputs in verifiable symbolic representations. As a result, true interpretability in Transformers is likely to come not only from improved post-hoc analysis, but from hybrid architectures where reasoning and transparency are built into the model design, bridging neural performance with human-understandable logic.
\end{tcolorbox}

\section{Robustness of Transformer}

Robustness, the ability of a model to behave consistently under noisy inputs, adversarial perturbations, or distributional shifts, is essential for building trust in AI systems. Although Transformers achieve impressive performance on standard benchmarks, their use in safety-critical applications such as autonomous driving and medical diagnostics demands reliability beyond average accuracy. Recent studies reveal a concerning contrast: Transformers can be highly robust in certain settings, such as in-context learning with noisy labels \cite{cheng2025exploring}, yet extremely fragile under adversarial conditions. For example, in reading comprehension tasks, adding semantically irrelevant \textit{distractor} sentences can reduce accuracy from 75\% to below 7\% \cite{jia2017adversarial}. This behavior, often described as overstability to spurious correlations, indicates that while Transformers are powerful pattern learners, they often lack strong semantic grounding. Addressing this issue requires a careful analysis of their failure modes, separating weaknesses in learned representations from structural limitations of the attention mechanism.

\subsection{The Fragility of Learned Representations} 
A major obstacle to the trustworthiness of Transformer models is the fragility of their learned representations. Unlike human understanding, which remains stable under meaning-preserving changes such as synonym substitution or syntactic variation, Transformers often depend on shallow, surface-level patterns. This weakness is clearly exposed by textual adversarial attacks, which exploit the discrete structure of language to separate model predictions from true semantic meaning. Jia et al. \cite{jia2017adversarial} show an important weakness in reading comprehension models, which is the difficulty in separating meaningful information from superficial distractors. In their study of sixteen Transformer-based models, the authors added adversarial sentences that shared keywords with the question but did not contain the correct answer. Although these sentences were irrelevant to human readers, model performance dropped sharply, with average F1 scores falling from 75\% to 36\%, and further to 7\% when ungrammatical sequences were used. This behavior highlights an overstability to surface-level cues, where models rely on spurious correlations such as word overlap rather than robust semantic reasoning, limiting their trustworthiness in real-world settings. While distractor attacks manipulate specific inputs, Wallace et al. \cite{wallace2019universal} identified a more fundamental vulnerability: universal adversarial triggers. These are fixed, input-independent token sequences (e.g., '\textit{zoning tapping fiennes}') that, when appended to virtually any text, can force a targeted model prediction. The authors showed that such triggers can reduce sentiment analysis accuracy from 85\% to below 30\% and even induce harmful outputs from generative models like GPT-2, independent of context. The effectiveness of a single, nonsensical phrase across diverse inputs suggests that this weakness originates in the way Transformers aggregate token embeddings. These results indicate that Transformer representations contain global blind spots, making them vulnerable to exploitation without any knowledge of the underlying input content. While universal triggers exploit systemic weaknesses, other attacks expose a different limitation: the inability of Transformers to maintain semantic invariance. Jin et al. introduced TextFooler \cite{jin2020bert}, a black-box attack that creates adversarial text by replacing words with semantically similar synonyms. By identifying important tokens and substituting them with nearby words in the embedding space, TextFooler can change model predictions while preserving the original meaning for human readers. Similarly, Li et al. proposed BERT-Attack \cite{li2020bert}, which uses attention-based gradients to locate vulnerable tokens and generate effective replacements. 

Together, these attacks reveal a significant semantic gap: Transformers are highly sensitive to changes that are minor or meaningless to humans. The ease with which synonym substitutions or nonsensical prefixes can override the intended meaning shows that many decisions are driven by fragile statistical patterns rather than a robust understanding of language.

\subsection{Architectural Robustness: Vision Transformers vs. CNNs} 

The rise of ViTs prompted early questions about their reliability compared to Convolutional Neural Networks (CNNs). It was initially thought that the global receptive field of self-attention might make ViTs more robust to perturbations. However, later studies present a more nuanced picture: while ViTs often generalize better to natural image corruptions, they remain as vulnerable as CNNs to adversarial attacks, though their failure modes differ and are linked to how they process spectral and geometric information.

Early claims about the robustness of Vision Transformers (ViTs) were often based on unfair comparisons, where large, heavily pre-trained Transformers were evaluated against smaller CNNs. Bai et al. \cite{bai2021transformers} addressed this issue by comparing models with matched capacity and training settings, such as ResNet-50 and DeiT-S. Under these controlled conditions, they showed that CNNs can match or even outperform ViTs against strong gradient-based attacks like PGD. They also found that ViTs are more fragile during adversarial training, often requiring careful warm-up and data augmentation strategies to avoid training collapse, an issue less common in CNNs. These findings are supported by Mahmood et al. \cite{mahmood2021robustness}, who evaluated ViTs under both white-box (where the attacker has full access to model gradients) and black-box attacks (where only model outputs are observed). Their results show that self-attention does not provide inherent protection against gradient-based attacks, with robust accuracy dropping to near zero under strong PGD (an iterative attack that maximizes model error within a fixed constraint) or C\&W attacks (an optimization-based attack designed to find the absolute minimum perturbation required to break a model), similar to CNNs. However, an important difference emerges in attack transferability: adversarial examples crafted for CNNs do not easily transfer to ViTs, and vice versa. This suggests that although both models are vulnerable, they rely on different features: CNNs emphasize local textures, while ViTs capture global relationships. This distinction can be exploited for defense; for example, combining ViT and BiT (a large-scale pre-trained ResNet architecture as a strong CNN baseline) models in an ensemble can significantly improve black-box robustness, as a single perturbation is unlikely to fool both architectures simultaneously.

\begin{figure}[htpb]
    \centering
\includegraphics[scale=0.34,clip=true]{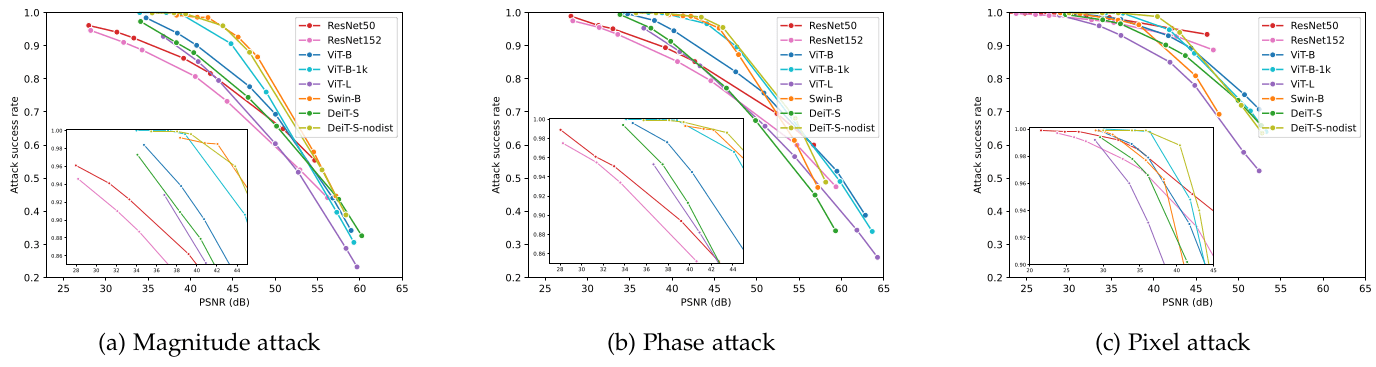} 
    \caption{Attack success rate vs. PSNR for various models under (a) magnitude perturbations, (b) phase perturbations, and (c) pixel-space perturbations. Vision Transformers (ViT, DeiT, Swin) are especially vulnerable to phase perturbations compared to CNNs (ResNet), showing a steeper drop in robustness under subtle spectral distortions. Adapted from Kim et al. \cite{kim2024exploring}.}
    \label{fig:kim}
\end{figure}

Although ViTs are more robust to local texture noise than CNNs, they have distinct weaknesses in the spectral domain. Kim et al. \cite{kim2024exploring} showed, using frequency-selective attacks, that ViTs rely strongly on low-frequency phase information. As shown in Figure \ref{fig:kim}, CNNs remain relatively stable under phase perturbations, whereas ViTs suffer significant accuracy drops even when these changes are imperceptible. This vulnerability arises from the self-attention mechanism, which emphasizes global structural information encoded in the phase rather than local pixel details.

Extending these findings, Tan et al.~\cite{tan2024transformation} identified transformation-dependent vulnerabilities in vision models. They proposed attacks in which perturbations remain harmless until a specific image transformation, such as scaling, gamma correction, or JPEG compression, is applied, after which they become adversarial. This shows that ViTs are vulnerable not only to pixel-level noise but also to the image processing pipeline itself. Moreover, these metamorphic attacks transfer across both ViTs and CNNs, challenging common defense strategies that assume adversarial perturbations remain stable under image transformations.

Despite these vulnerabilities, ViTs show strong performance in natural robustness, meaning their ability to generalize to unseen domains and naturally occurring noise. Paul and Chen \cite{paul2022vision} evaluated ViTs on benchmarks such as ImageNet-C (common corruptions), ImageNet-A (natural adversarial examples), and ImageNet-R (renditions). Their results show that ViTs consistently outperform CNNs in these out-of-distribution settings. In particular, on ImageNet-A, ViTs achieved a 4.3$\times$ improvement in top-1 accuracy (28.1\% versus 6.5\%) compared to comparable BiT models. This robustness is largely due to the shape bias of self-attention. Unlike CNNs, which often rely on high-frequency texture cues (such as identifying a cat by its fur pattern), ViTs focus more on global shape and structure. As a result, they can maintain performance even when textures are distorted by noise or contrast changes, as long as the overall geometry is preserved. This highlights an important tradeoff in ViT robustness: they are resilient to natural, semantic variations but remain vulnerable to targeted spectral and phase-based attacks.


\subsection{Enhancing Reliability: Training, Architecture, and Verification} 

Given the vulnerabilities of standard Transformers, ranging from spectral sensitivity to backdoor attacks, robustness cannot be treated as an afterthought. Instead, it must be built into the model’s training process, architecture, and verification pipeline. This section discusses important strategies for moving from fragile accuracy toward truly reliable and robust models.

A key strategy to improve Transformer robustness is diversifying the training distribution. Despite their capacity, Transformers often overfit to familiar data and fail catastrophically under out-of-distribution (OOD) shifts. Hendrycks et al. \cite{hendrycks2020pretrained} show that large-scale pretraining on diverse corpora, as in RoBERTa, greatly improves OOD performance, while aggressive model compression, such as DistilBERT, can remove redundant features and reduce robustness. Beyond dataset scale, training quality is critical. Shao et al. \cite{shao2021adversarial} demonstrate that standard training biases models toward high-frequency, non-robust features, but frequency-aware adversarial training can guide ViTs to focus on low-frequency, semantic content, boosting robust accuracy (59.8\% vs. 16.7\% for CNNs under AutoAttack). However, this can introduce gradient masking, where standard adversarial training causes underflow in attention blocks. Jain et al. \cite{10658195} address this with Adaptive Attention Scaling, dynamically adjusting pre-softmax outputs during training to prevent numerical issues and ensure the model learns from genuine adversarial signals. 

\begin{figure}[htpb]
    \centering
\includegraphics[scale=0.3,clip=true]{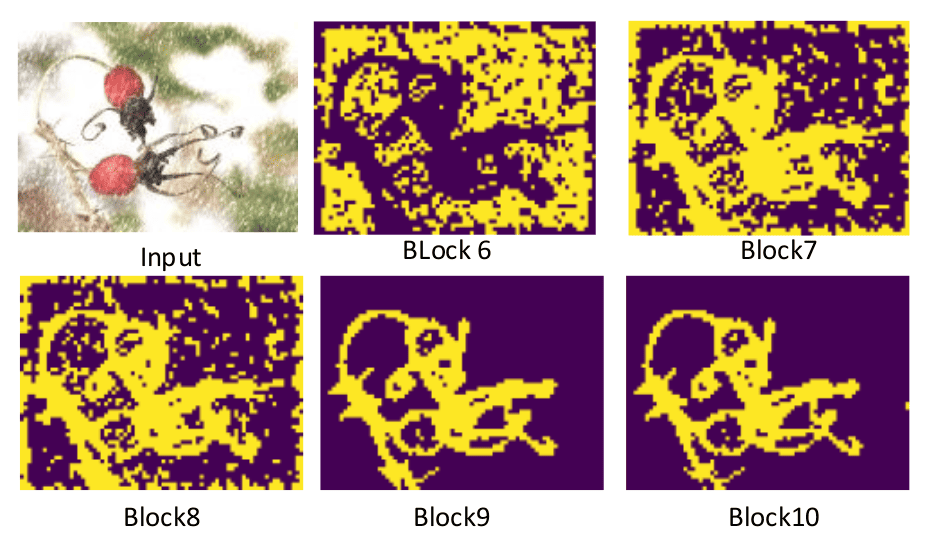}
    \caption{Clustering visualization of token features across different layers in the FAN-S model. As the network goes deeper, token representations become more grouped, demonstrating the emergence of meaningful visual segments. This progressive clustering highlights the role of self-attention in promoting robust mid-level abstractions. Adapted from Zhou et al. \cite{zhou2022understanding}}
    \label{fig:FAN}
\end{figure}

One might ask whether robustness can emerge naturally from the architecture itself. Zhou et al. \cite{zhou2022understanding} suggest that the self-attention mechanism acts as an information bottleneck, filtering out noise and encouraging semantically related tokens to cluster together. As shown in Figure \ref{fig:FAN}, as data passes through the layers, token representations spontaneously form meaningful visual segments, such as separating an object from its background. Building on this idea, the authors introduced Fully Attentional Networks (FAN), which replace standard MLP blocks with attentional channel processing. This enhances the token clustering effect, achieving state-of-the-art robustness on corrupted benchmarks like ImageNet-C without costly adversarial training. Their results suggest that the shape bias, focusing on objects rather than textures, can be reinforced directly through architectural design.

To this end, building trust requires verification. Even robust models can be compromised by Trojan or backdoor attacks, hidden triggers embedded during training that force specific failures. Harikumar et al \cite{harikumar2021scalable} address this with the Scalable Trojan Scanner, which can reverse-engineer potential triggers by identifying patterns that collapse the model’s output, allowing detection of compromised models without needing prior knowledge of the target class.
For applications requiring absolute certainty, empirical testing is not enough. Shi et al. \cite{shi2020robustness} introduced Formal Verification for Transformers, modeling self-attention non-linearities as bounded linear approximations to mathematically guarantee that predictions remain unchanged within a defined perturbation radius. While currently feasible only for smaller models, this approach represents the high standard for trust: moving from \textit{it didn’t fail in tests} to \textit{it cannot fail under these conditions}.

\begin{tcolorbox}[colback=orange!5!white,colframe=red!75!black]
\noindent Overall, the study of Transformer robustness shows a basic conflict between their strong representation ability and their structural weaknesses. Although these models perform very well on standard benchmarks, they remain vulnerable to small changes that do not change the true meaning, such as distracting text, universal triggers, or spectral phase noise. This shows that they often depend on fragile surface patterns instead of stable causal relationships. Because of this, building real trust requires more than just training on diverse or augmented data. It calls for a broader approach that combines frequency-aware training, stronger architectural biases (such as those used in FAN), and formal verification methods. Until Transformers are designed to remain reliable even under rare or extreme conditions, their use in safety-critical systems must be approached with caution.
\end{tcolorbox}

\section{Fairness and Bias Mitigation}
Deploying Transformer models in high-stakes domains assumes that their decisions are not only accurate but also fair. However, these models often mirror and amplify the biases present in their training data. Trustworthiness, therefore, goes beyond predictive performance to include fairness to ensure that models do not cause allocation harms (unequal access to resources) or representation harms (reinforcing negative stereotypes). Bias in Transformers is not just an incidental issue, it is a structural challenge arising from how the models extract and prioritize patterns from \textit{uncurated data}, which is a vast, 
unfiltered web corpora that inherently encodes historical inequities and societal stereotypes 
as statistical facts.
\subsection{Amplification of Systemic Bias} 
While Transformers have transformed feature extraction, their dependence on large, uncurated datasets introduces a major safety concern: not just reproducing, but amplifying societal biases. Unlike earlier models, Transformers act as \textit{over-parameterized sponges}, absorbing both meaningful patterns and spurious demographic correlations. This can lead to harms ranging from allocation harm (denying resources to protected groups) to representation harm (reinforcing negative stereotypes).

\begin{figure}[htpb]
    \centering
    \includegraphics[scale=0.3,clip=true]{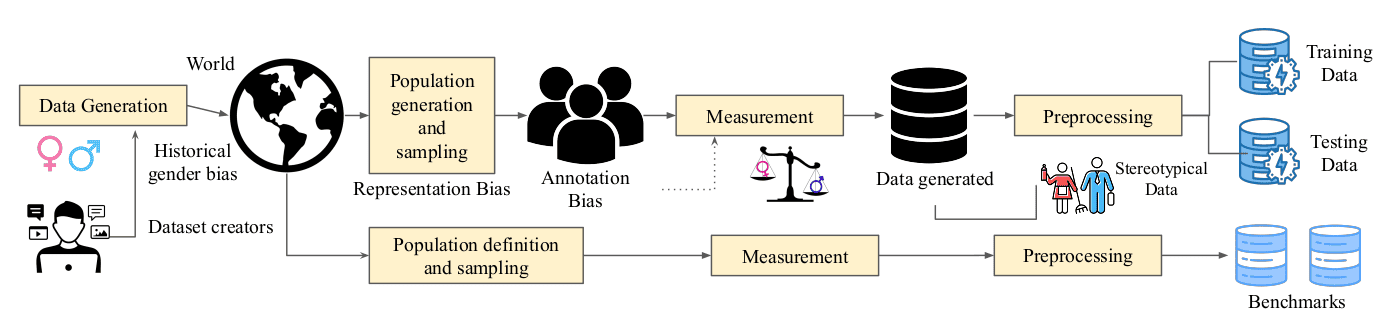} 
    \caption{Conceptual representation of how gender bias is introduced during the NLP data generation and modeling pipeline. Adapted from Nemani et al. \cite{nemani2023gender}}
    \label{fig:nemani}
\end{figure}

In NLP, bias often arises from models relying on statistical co-occurrence. Nemani et al. \cite{nemani2023gender} describe bias overamplification, where models exaggerate existing data imbalances to reduce training loss. For example, if the training data shows that \textit{Programmer} is 70\% male, a model may push this probability close to 90\%, which effectively erases the minority class. This problem is especially evident in machine translation. Gender-neutral sentences like '\textit{The doctor asked the nurse to help her}' are often translated with male pronouns for \textit{doctor} (e.g., Spanish '\textit{El doctor}'), ignoring the contextual cue '\textit{her}'. Such errors reflect a structural bias: the model’s learned prior (Doctor = Male) overrides the actual text. These biases, denigration, stereotyping, and underrepresentation, arise from a cascade of issues, from skewed data collection to loss functions that favor majority-class accuracy over fairness (Figure \ref{fig:nemani}).

While linguistic bias is largely driven by co-occurrence, visual bias in Transformers is amplified by the architecture itself. Mandal et al. \cite{mandal2023biased} show that ViTs are more prone to bias than CNNs. Using the Accuracy Difference ($\Delta$) metric, ViTs (e.g., ViT-B/32) show an average $\Delta$ of 0.17 compared to 0.11 for CNNs, which is a 54\% increase in bias impact. This is likely due to the global self-attention mechanism. Unlike CNNs, which focus on local textures (e.g., recognizing a doctor by a stethoscope), ViTs aggregate global context, often relying on background cues (e.g., a boardroom suggesting '\textit{male}'). As a result, occupations like '\textit{CEO}' or '\textit{Engineer}' show stronger male associations in ViTs. The risk is amplified in multimodal models like CLIP, where visual stereotypes reinforce linguistic ones, causing generative outputs to consistently portray female-associated roles (e.g., '\textit{housekeeper}') in stereotypical ways.

\subsection{Bias Mitigation Strategies}

Identifying the systemic amplification of bias is only the first step; the real challenge is preventing it. Because bias is embedded throughout the modeling pipeline from attention mechanisms that capture spurious correlations to loss functions that favor majority groups, mitigation must target these same components. Accordingly, bias-mitigation strategies can be described into two categories: \textit{Internal Alignment}, which guides the model toward fairer representations during learning, and \textit{Outcome Alignment}, which shapes the optimization process to ensure equitable performance across groups.

An important source of representational bias is the tendency of attention mechanisms to take shortcuts. Instead of focusing on meaningful content, attention often concentrates on superficial tokens. Gesi et al. \cite{gesi2024beyond} show this clearly in code models, where standard Transformers such as CodeBERT place excessive attention on special tokens like [CLS] and [SEP], even though these tokens carry little or no semantic information. As a result, important elements such as variable names and operators receive less attention than they should. To address this issue, the authors propose SyntaGuid, a method that explicitly guides attention during training. Using information from Abstract Syntax Trees, they add an auxiliary loss that penalizes the model when its attention does not align with known syntactic structure. This forces the model to attend to relevant features rather than relying on shortcuts. The approach leads to a 3.25\% improvement in overall performance and corrects 28.3\% of previously wrong predictions. The broader implication is important: attention can be explicitly constrained, reducing spurious learning and improving robustness and fairness.

While guided attention improves internal representations, it does not ensure that model decisions are fair across all groups. Zhang et al. \cite{zhang2024towards} highlight a key issue in robust training known as the robustness–fairness trade-off. They show that standard adversarial training often increases average robustness by disproportionately harming vulnerable or minority classes. To address this, they propose Fairness-Aware Adversarial Learning (FAAL). Instead of treating all errors equally, FAAL reweights the adversarial loss to prioritize classes that are more vulnerable to attacks. During training, these classes receive higher importance, forcing robustness gains to be distributed more evenly. Importantly, FAAL achieves this with minimal overhead, requiring only two fine-tuning epochs. This approach shifts robustness optimization from improving average performance to protecting worst-case groups, which is essential for safety-critical applications.

\subsection{Privacy-Preserving Architectures}
The trustworthiness of a Transformer depends not only on its predictions but also on how well it protects sensitive data. Large models are known to memorize and sometimes reproduce private information from their training data, such as personal or medical records. As a result, privacy must be built into the model rather than treated as an afterthought. This leads to the utility–privacy trade-off: stronger privacy guarantees often reduce the model’s ability to learn useful and generalizable patterns.

The method for protecting training data is Differential Privacy (DP), most commonly implemented through Differentially Private Stochastic Gradient Descent (DP-SGD) \cite{abadi2016deep}. By clipping per-sample gradients and injecting calibrated noise, DP-SGD provides formal guarantees that the inclusion of any single individual in the training set cannot be inferred from the final model parameters. Historically, however, this noise severely degraded the performance of large Transformer models. Anil et al. \cite{anil2021large} challenged this assumption by demonstrating that high-utility pretraining of BERT-Large is achievable even under strict privacy budgets ($\epsilon = 5.36$), provided the training dynamics are carefully redesigned. Their key insight was the use of mega-batches that exceed two million examples per update, which dramatically increases the signal-to-noise ratio of gradient estimates, effectively mitigating the impact of DP noise. Moreover, they observed that scale-invariant components such as LayerNorm interact poorly with DP perturbations, necessitating aggressive weight decay to maintain training stability. In settings where direct data sharing is infeasible, such as healthcare, Castellon et al. \cite{castellon2023dp} propose DP-TBART, a generative alternative to conventional privacy-preserving learning. Rather than training downstream models directly on sensitive records, DP-TBART employs an autoregressive Transformer trained with DP-SGD to generate synthetic tabular data. These synthetic datasets preserve high-order feature dependencies and statistical structure while preventing exposure of real individuals, enabling safe downstream analysis without compromising privacy.

While training-time privacy protects the dataset, inference privacy protects the user’s input. In cloud-based deployments, users typically send plaintext queries to model providers, creating serious privacy and surveillance risks. To enable computation without show inputs, Chen et al. \cite{chen2022x} proposed \textit{THE-X}, a framework for Transformer inference on Homomorphically Encrypted data. Since HE supports only addition and multiplication, it is incompatible with key Transformer operations such as Softmax, GELU, and LayerNorm. THE-X addresses this by replacing these non-linear functions with polynomial approximations that are HE-compatible. Despite the approximation error, the framework achieves accuracy within 1.5\% of standard plaintext inference. As a result, users can encrypt their queries locally, perform inference in the cloud, and receive encrypted outputs, ensuring that the model provider never accesses the raw input.

Finally, privacy loss is not uniformly distributed across populations. Recent studies show that underrepresented groups, often the same groups most affected by model bias, suffer greater privacy leakage. Because their data points are statistical outliers, models tend to memorize them more strongly to reduce training loss. This reveals a deep connection between privacy and fairness: protecting only the average case is insufficient. A trustworthy Transformer must incorporate adaptive gradient clipping and group-aware noise calibration so that privacy guarantees apply equally to both majority and marginalized groups.

\begin{tcolorbox}[colback=orange!5!white,colframe=red!75!black]
\noindent In conclusion, fairness and privacy are not peripheral ethical constraints but foundational prerequisites for the trustworthiness of Transformer models. Our analysis demonstrates that, absent deliberate intervention, these architectures systematically amplify existing inequities, reinforcing historical biases through statistical co-occurrence and global attention, while disproportionately exposing the privacy of marginalized subgroups. Trustworthy deployment, therefore, demands a comprehensive engineering paradigm: one that pairs the expressive power of self-attention with principled constraints on representation learning, decision-making, and data protection. Only by embedding fairness and privacy as intrinsic architectural objectives, rather than post hoc corrections, can Transformer models realize their transformative potential while serving all segments of society equitably.
\end{tcolorbox}

\section{Trustworthy Transformers for Scientific Machine Learning}

In recent years, transformer architectures, initially designed for language processing, have found a growing role in scientific computing. Their ability to model complex relationships and long-range dependencies makes them ideal for tackling problems in fluid dynamics, molecular modeling, and climate simulations. However, in scientific computing, trustworthiness extends far beyond predictive accuracy on a held-out validation set. A reliable model must have physical consistency (adherence to conservation laws and invariants), numerical stability, and interpretability (alignment with established mathematical structure). While standard Transformers good at discrete sequence modeling, they are poorly matched to the continuous, multi-scale, and highly nonlinear nature of physical systems. This mismatch has motivated a growing body of work aimed at grounding Transformers in physical reality, restructuring their architectures to explicitly encode the geometric, temporal, and symbolic structure of Partial Differential Equations (PDEs), rather than treating physical dynamics as generic token sequences.

\subsection{Physics-Informed Transformers for Scientific Computing}

\begin{figure}[htpb]
    \centering
    \includegraphics[scale=0.34,clip=true]{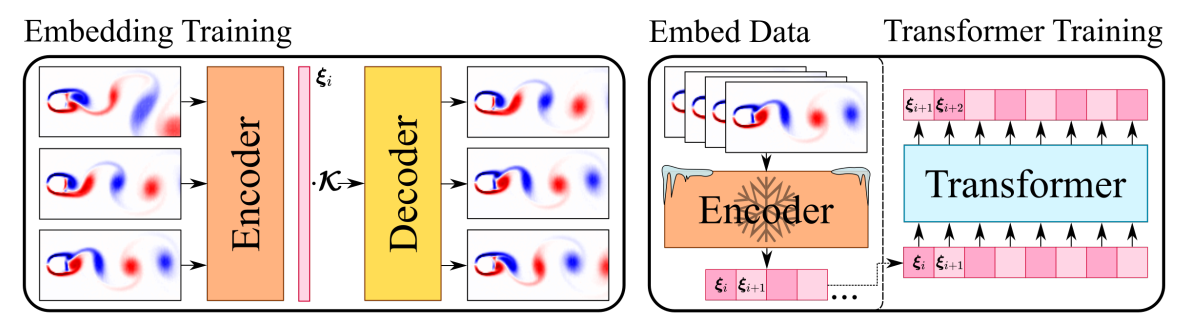}
    \caption{Two-stage training process using Koopman embeddings and a transformer decoder for modeling physical dynamics. Adapted from Geneva et al.~\cite{geneva2022transformers}}
    \label{fig:dyna}
\end{figure}

Recent advancements in Scientific Machine Learning (SciML) have demonstrated that transformer architectures are uniquely suited for modeling the solution operators of PDEs. While traditional numerical solvers are often computationally challenging for high-dimensional or multi-scale systems, neural operators \cite{
goswami2024learning,Zhang2025BubbleOKAN,peyvan2024riemannonets,lu2021learning} offer a faster alternative by approximating the mapping from input conditions to output solutions in a single forward pass. 
A major obstacle to trust in SciML is the difficulty standard neural networks face in modeling stiff temporal dynamics, where fast, high-frequency modes coexist with slow evolution. In such systems, conventional self-attention tends to behave like a low-pass filter, smoothing out precisely the oscillations that are essential for accurate physical behavior. To address this limitation, Geneva et al. \cite{geneva2022transformers} combine Koopman Operator Theory with Transformer architectures. Their approach uses an encoder to map nonlinear physical states, such as chaotic fluid flows, into a latent space where the dynamics evolve approximately linearly. In this transformed coordinate system, the Transformer predicts long-term evolution more stably, avoiding error accumulation and unphysical divergence (Figure \ref{fig:dyna}). By shifting prediction to a linearized dynamical space, the model preserves temporal fidelity and maintains physically consistent behavior over long horizons.

\begin{figure}[htpb]
    \centering
    \includegraphics[width=0.7\textwidth]{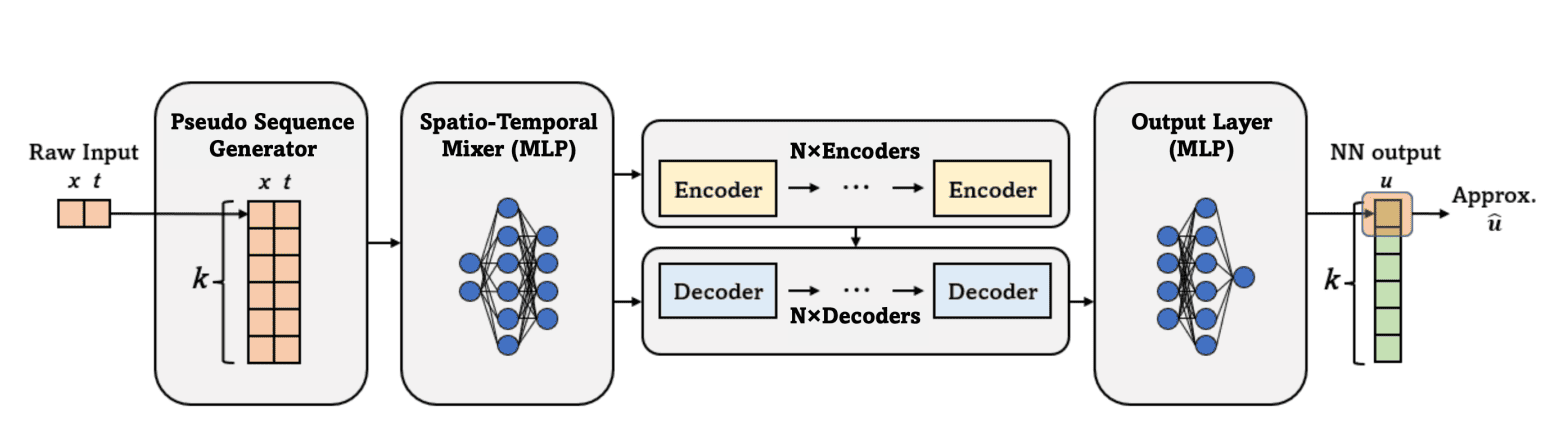}
    \caption{PINNsFormer architecture. The model projects spatio-temporal inputs into a pseudo sequence, embeds the same in an encoder-decoder Transformer, and predicts sequentially. Adapted from Zhao et al.~\cite{zhao2023pinnsformer}}
    \label{fig:pinnsformer_arch}
\end{figure}

Addressing the same challenge in the context of Physics-Informed Neural Networks (PINNs) \cite{raissi2019physics,jagtap2020adaptive,jagtap2022deepknn,abbasi2025history,jagtap2022deepWW,abbasi2025challenges,menon2025anant,jagtap2020extended,jagtap2023important,jagtap2020locally,hu2021extended,jagtap2022physics}, Zhao et al. \cite{zhao2023pinnsformer} propose the PINNSFormer. Unlike standard MLPs that struggle with spectral bias (learning low frequencies first), PINNSFormer employs wavelet-based activation functions within a Transformer encoder-decoder to explicitly capture high-frequency components. By converting spatio-temporal inputs into pseudo-sequences, it resolves stiff reaction-convection dynamics with orders-of-magnitude lower error than baseline solvers (Figure \ref{fig:pinnsformer_arch}). 

\begin{figure}[htpb] 
    \centering 
    \includegraphics[width=0.8\textwidth]{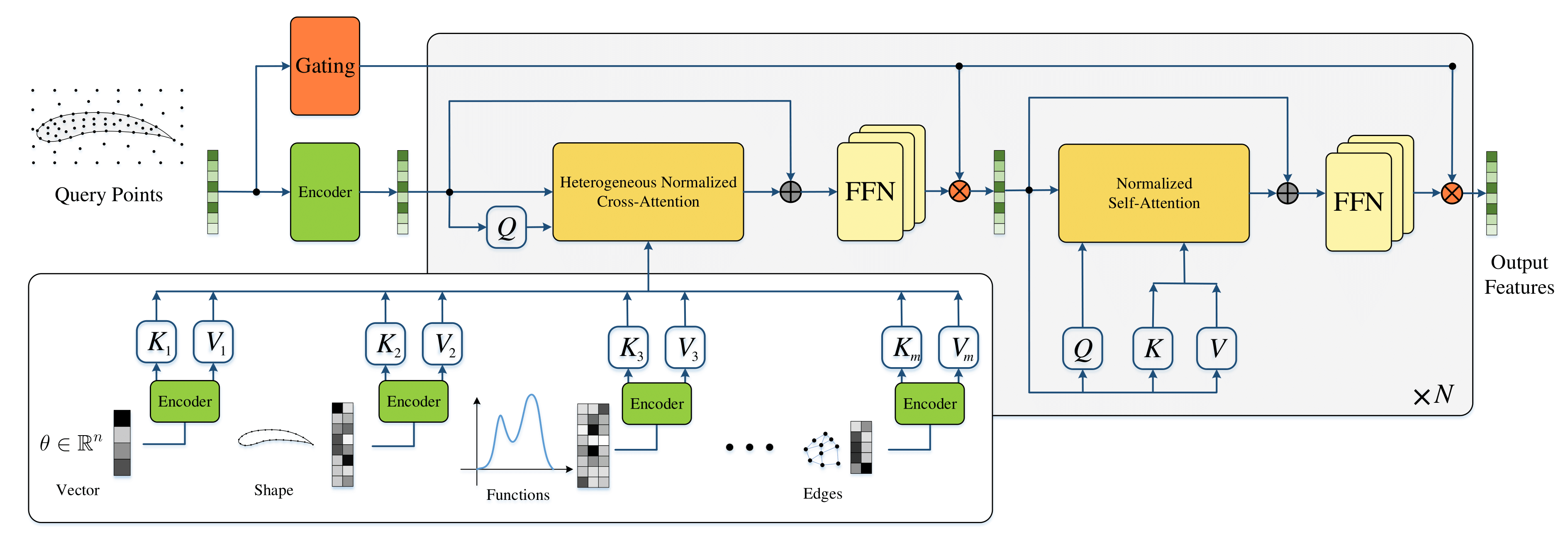}
    \caption{Diagram showing the main components of the GNOT framework, featuring normalized attention blocks for heterogeneous inputs and geometric gating for multi-scale processing. Adapted from Hao et al.~\cite{hao2023gnot}}
    \label{fig:gnot_framework}
\end{figure}

Trustworthy models must also function outside of idealized laboratory conditions, specifically on the irregular, non-uniform meshes found in real-world engineering. Hao et al. \cite{hao2023gnot} argue that standard attention mechanisms, which assume regular grids, fail to capture the topology of complex domains. They introduce the General Neural Operator Transformer (GNOT), which utilizes \textit{Geometric Gating}, a mechanism inspired by domain decomposition. This allows expert networks to specialize in specific geometric regions, enabling the model to generalize effectively across varying mesh densities and irregular shapes in elasticity and heat conduction tasks (Figure \ref{fig:gnot_framework}).

\begin{figure}[htpb]
    \centering
    \includegraphics[width=0.75\textwidth]{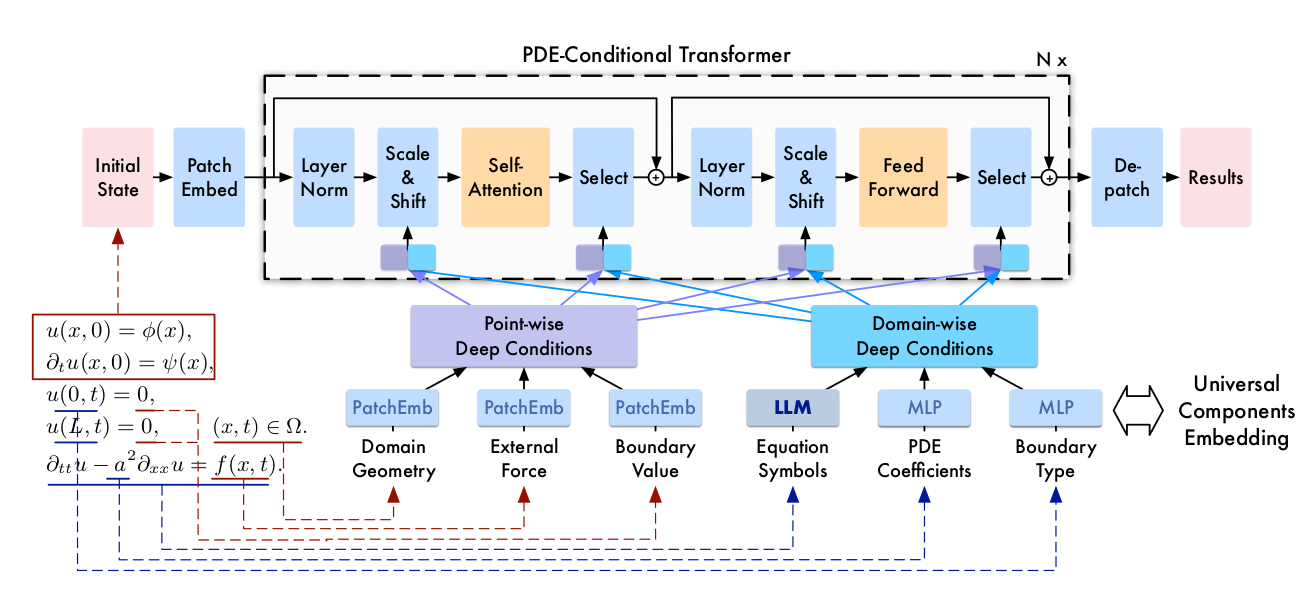}
    \caption{Overview of the Unisolver architecture. The model embeds complete PDE components—equation symbols, coefficients, and boundary types, into a decoupled Transformer backbone to enable generalized solving. Adapted from Zhou et al.~\cite{zhou2024unisolver}}
    \label{fig:unisolver_architecture}
\end{figure}

\begin{figure}[htpb]
    \centering
        \includegraphics[scale=0.35, clip=true]{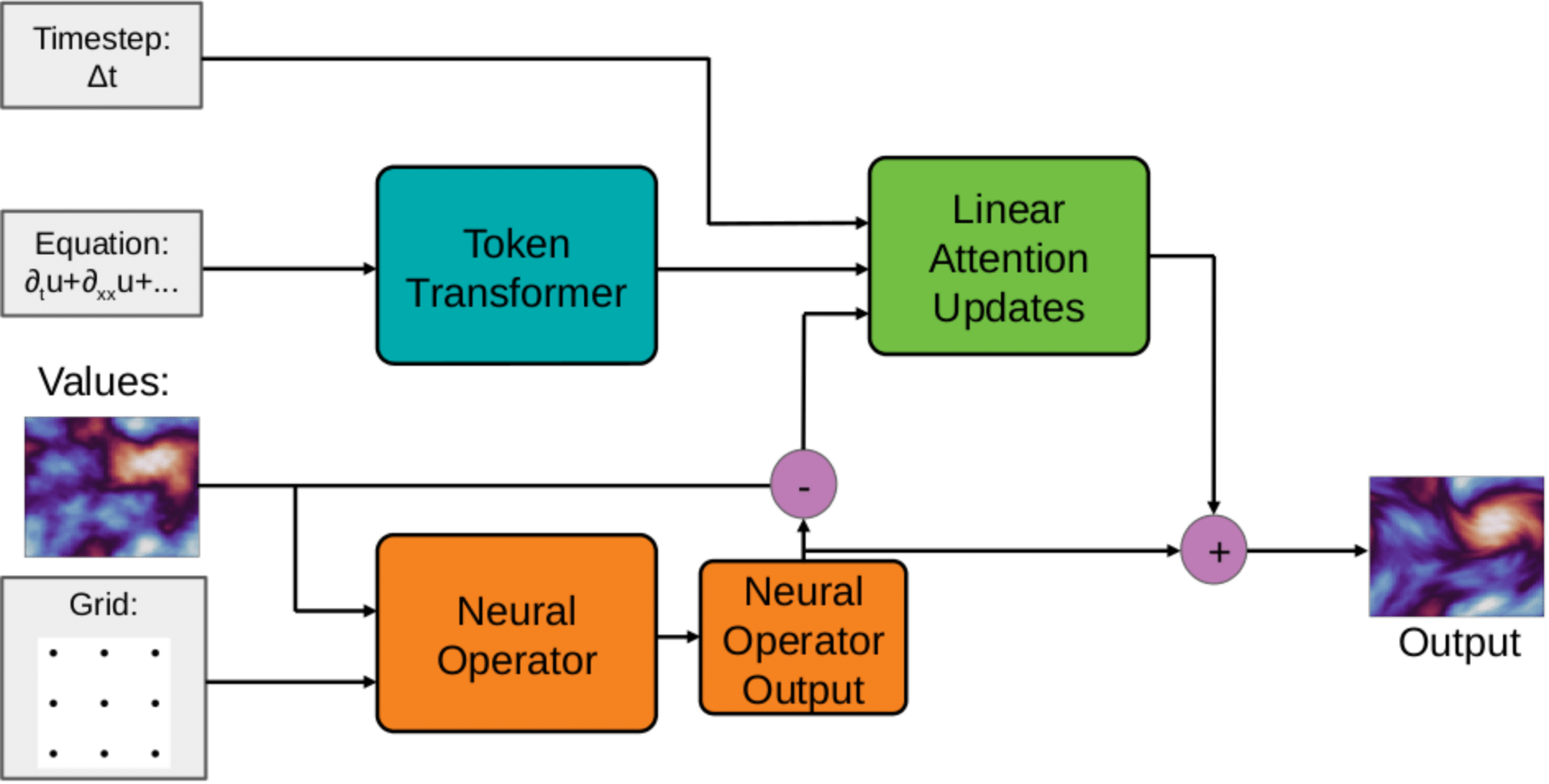}
    \caption{The Physics-Informed Token Transformer (PITT) framework. The model tokenizes PDEs to learn semantic representations of physical laws, which then guide numerical updates. Adapted from Lorsung et al.~\cite{lorsung2024physics}.}
    \label{fig:pitt_diagram}
\end{figure}

Moving towards universal applicability, recent works leverage the structured nature of PDE equations themselves. Zhou et al. \cite{zhou2024unisolver} propose Unisolver, a universal PDE solver that categorizes inputs into domain-wise (geometry) and point-wise (coefficients) conditions. Notably, it processes the equation symbols using a Large Language Model (LLaMA-3 \cite{grattafiori2024llama}) to generate physics-aware embeddings for the Transformer backbone (as shown in Fig \ref{fig:unisolver_architecture}). Complementing this symbolic approach, Lorsung et al. \cite{lorsung2024physics} introduce the Physics-Informed Token Transformer (PITT). PITT explicitly parses and tokenizes governing equations (e.g., differential operators) into symbolic sequences (shown in Fig \ref{fig:pitt_diagram}). These sequences guide a \textit{Numerical Update Module} that applies physics-consistent corrections to predictions. Both methods demonstrate that embedding symbolic mathematical structure directly into the network significantly enhances zero-shot generalization and interpretability.

\begin{figure}[htbp]
    \centering
\includegraphics[width=0.73\textwidth]{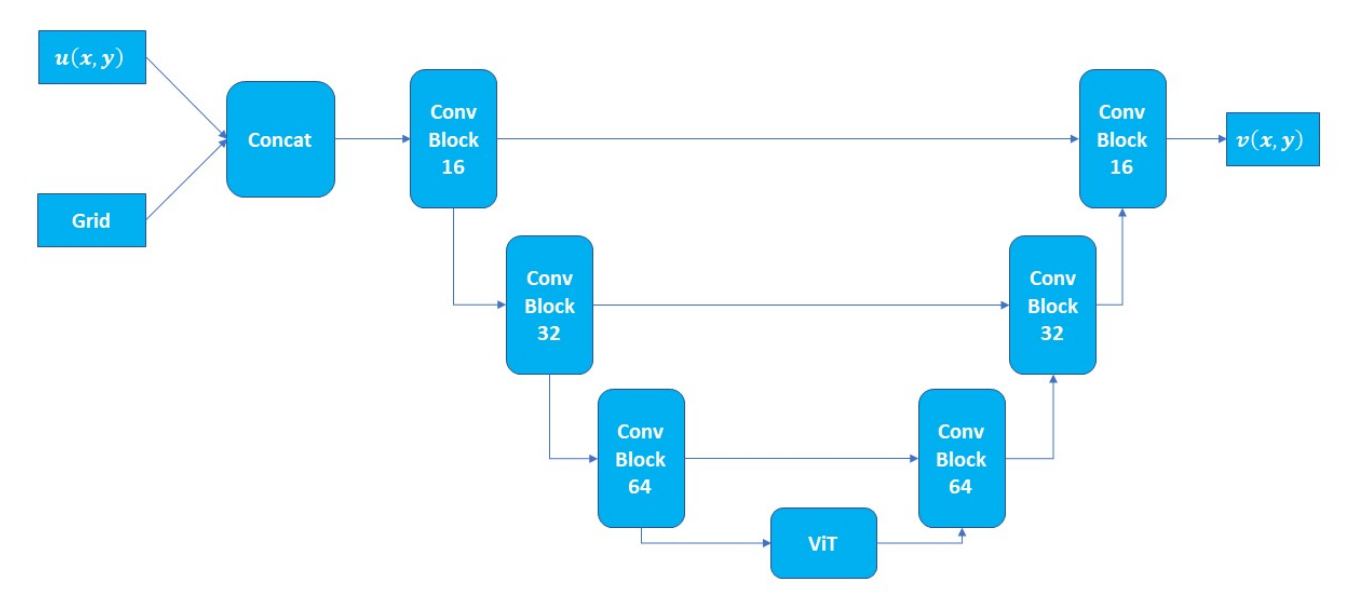}
    \caption{Architecture of ViTO. A hybrid U-Net and Vision Transformer design that encodes discretized solutions to perform inverse reconstruction and super-resolution. Adapted from Ovadia et al.~\cite{ovadia2024vito}}
    \label{fig:vito_architecture}
\end{figure}

While the mentioned models focus on forward problem solving, Transformers are also known for inverse problems, where unknown parameters (e.g., material properties) must be recovered from sparse observations. Ovadia et al. \cite{ovadia2024vito} introduce ViTO, a hybrid architecture combining U-Net encoders with Vision Transformers (as shown in Fig \ref{fig:vito_architecture}). Designed for ill-posed problems, ViTO utilizes relative positional embeddings to generalize across grid sizes, effectively reconstructing fine-scale features from coarse, noisy data in Darcy flow and wave equations. Similarly, Guo et al. \cite{guo2022transformer} tackle Electrical Impedance Tomography (EIT), a challenging boundary-value inverse problem. Their framework transforms 1D boundary measurements into 2D feature maps via harmonic extension, which are then processed by a modified attention module designed to mimic integral operators. This ensures mathematical consistency and allows the model to capture global dependencies essential for inferring internal conductivity distributions. Together, these approaches illustrate that Transformer architectures, when grounded in physical theory, whether through operator learning, symbolic tokenization, or integral formulations, offer a robust path toward reliable scientific computing.

\subsection{Uncertainty Quantification in Transformer}

Standard neural networks are typically trained as deterministic function approximators, $\hat{y} = f_\theta(x)$, producing point estimates that lack a measure of confidence. This limitation is important in high-stakes scientific applications where understanding the reliability of a prediction is as important as the prediction itself. While Bayesian inference offers a rigorous framework for quantifying uncertainty via the posterior distribution $p(\theta \mid \mathcal{D})$, calculating this analytically is often intractable for high-dimensional models like Transformers. Consequently, recent research has focused on two primary pathways to integrate Uncertainty Quantification (UQ) into these architectures: Bayesian approximations and Ensemble strategies.

\begin{figure}[htpb]
\centering
\includegraphics[width=0.65\textwidth]{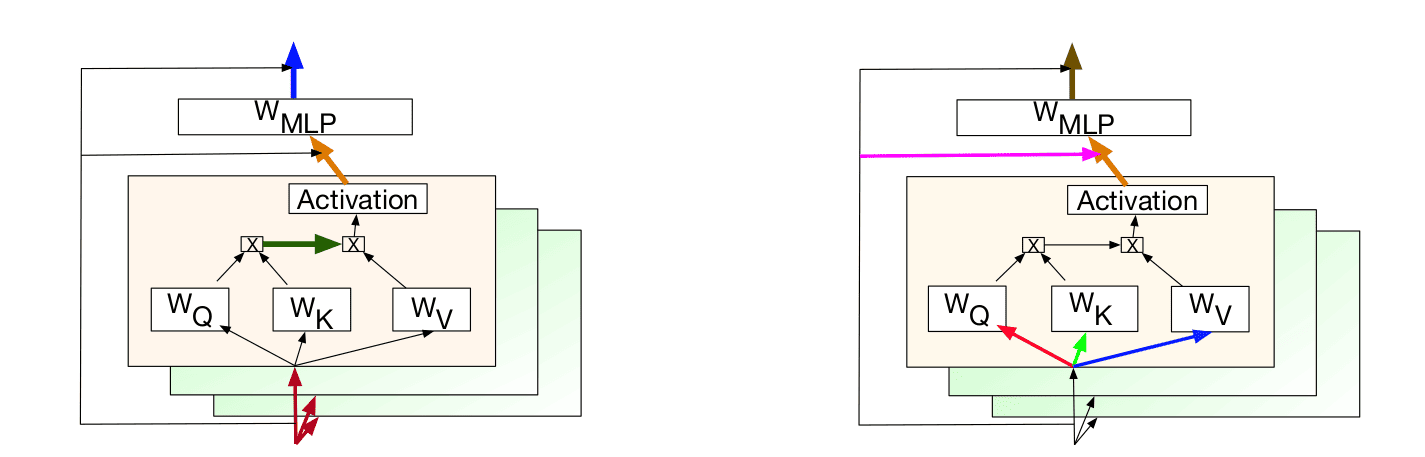}
\caption{Comparison of dropout placement in standard Transformer (left) versus BayesFormer (right). The modified design enables the architecture to simulate samples from a variational posterior. Adapted from Sankararaman et al. \cite{sankararaman2022bayesformer}.}
\label{fig:bayesformer}
\end{figure}

The path to UQ is Bayesian inference, which seeks to compute the posterior distribution of model parameters. Since calculating this analytically is intractable for high-dimensional Transformers, researchers use stochastic injection to simulate the posterior during inference. Sankararaman et al. \cite{sankararaman2022bayesformer} introduce BayesFormer, which utilizes structured dropout to approximate variational inference. Unlike standard dropout used for regularization, BayesFormer treats embedding matrices and attention weights as probabilistic random variables. By applying independent dropout masks during multiple inference passes (Monte Carlo sampling), the model generates a distribution of predictions rather than a single value, allowing for the estimation of predictive variance without the cost of training multiple models (Figure \ref{fig:bayesformer}). 

\begin{figure}[htpb]
    \centering
    \includegraphics[width=0.65\textwidth]{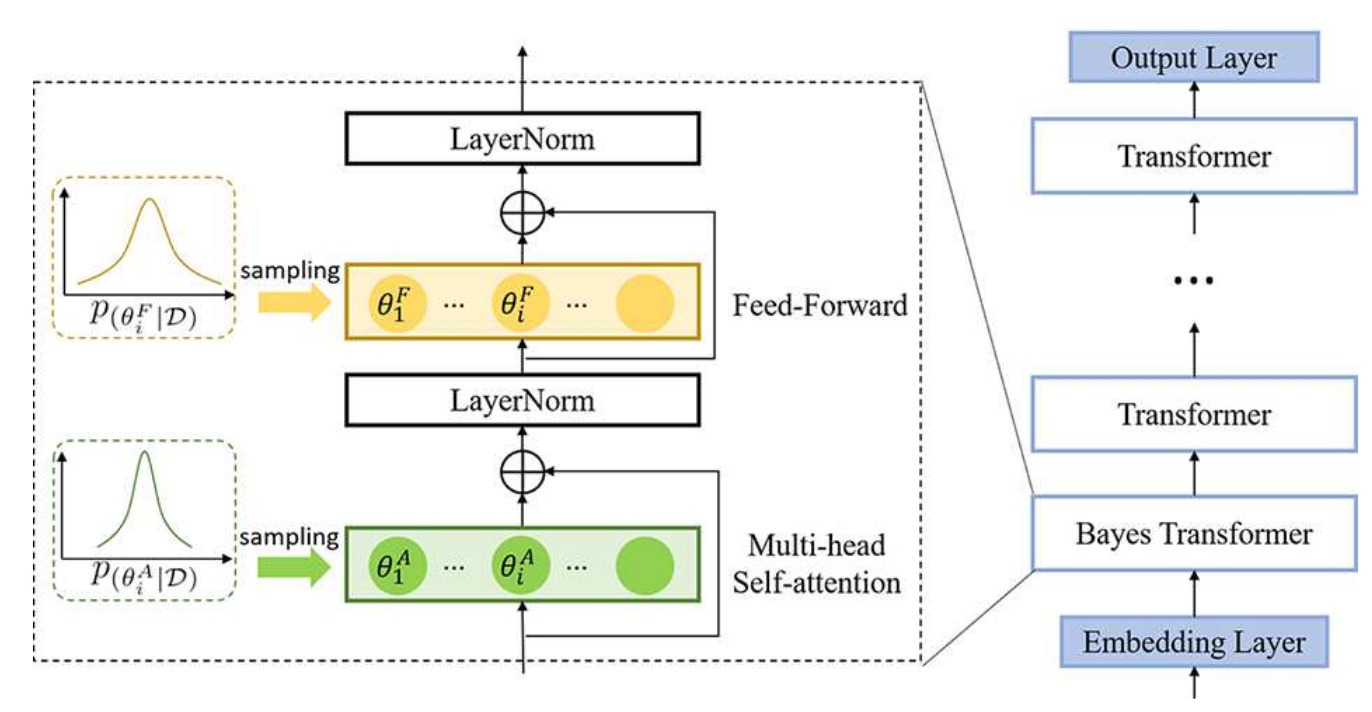}
    \caption{Bayesian Transformer Language Model Architecture. Bayesian learning is applied selectively to specific layers (highlighted) to reduce computational overhead while capturing model uncertainty. Adapted from Xue et al. \cite{xue2021bayesian}}
    \label{fig:xue}
\end{figure}

For specific domains, this stochasticity can be specified further. In industrial fault diagnosis, Xiao et al. \cite{xiao2024bayesian} embed lognormal distributions directly into the attention mechanism. This allows uncertainty to emerge naturally from the variability of the attention maps themselves, offering robustness in noisy, cross-domain settings. Conversely, for efficiency-critical tasks like speech recognition, Xue et al. \cite{xue2021bayesian} demonstrate that full-network stochasticity is unnecessary. Their Bayesian Transformer Language Model places Gaussian-distributed weights only in specific feed-forward layers (Figure \ref{fig:xue}), achieving an optimal trade-off that reduces error rates on benchmarks like DementiaBank while minimizing computational overhead. 

\begin{figure}[htpb]
\centering
\includegraphics[width=0.75\textwidth]{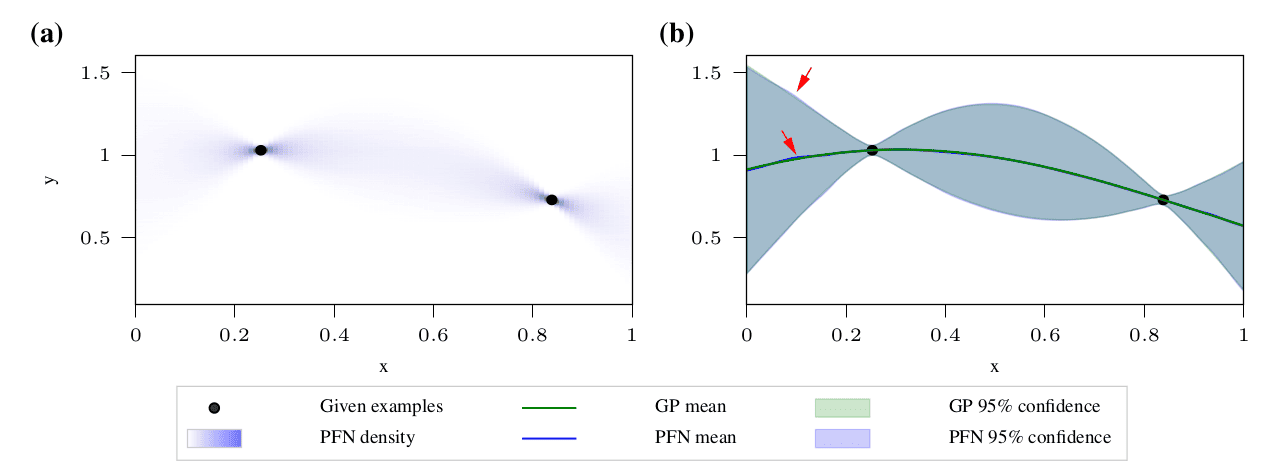}
\caption{Posterior predictive distributions from a PFN trained on GP samples. The PFN closely matches the GP's predictive mean and confidence intervals (blue vs. green), capturing uncertainty even in extrapolation regions. Adapted from Muller et al. \cite{muller2021transformers}}
\label{fig:gp-match}
\end{figure}

\begin{figure}[htpb]
    \centering
    \includegraphics[width=0.65\textwidth]{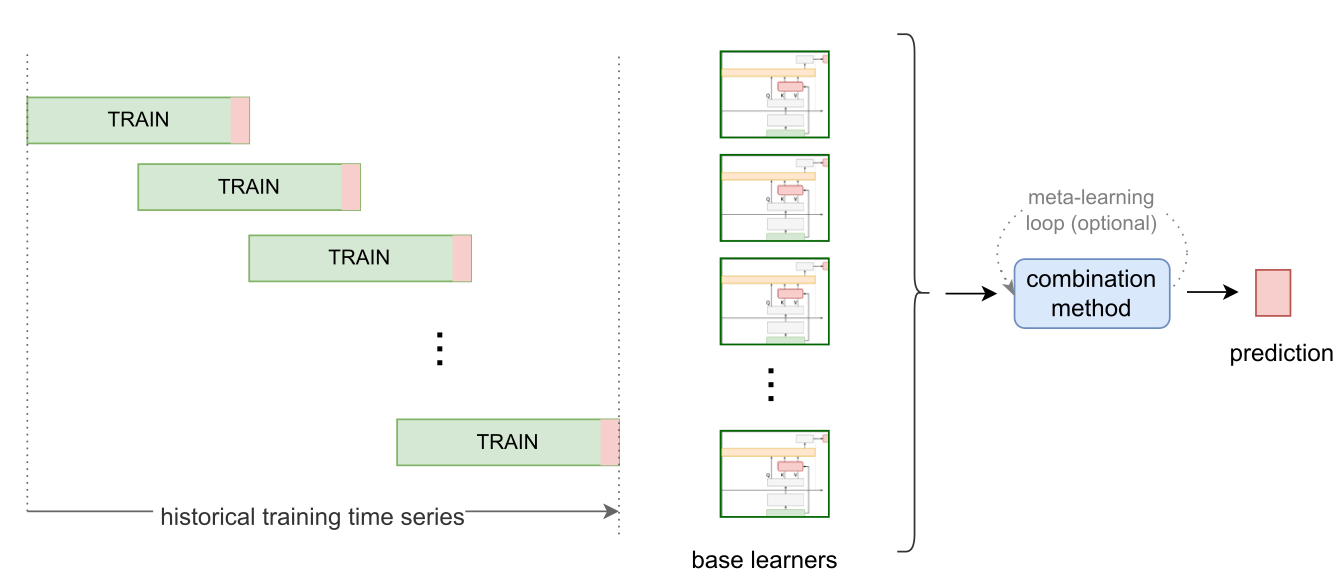}
    \caption{Architecture of WETT, showing base models trained on overlapping sliding windows aggregated via q-stack for uncertainty-aware financial forecasting. Adapted from Olorunnimbe et al. \cite{olorunnimbe2024ensemble}}
    \label{fig:ensemble}
\end{figure}

A radically different approach is to learn the posterior rather than approximate it. Muller et al. \cite{muller2021transformers} propose Prior-Data Fitted Networks (PFNs), which reframe Bayesian prediction as a supervised meta-learning task. By training on synthetic datasets generated from Gaussian Processes (GPs), PFNs learn to output a predictive distribution $q_\theta(y \mid x, \mathcal{D})$ that explicitly matches the true Bayesian posterior. This allows the Transformer to generalize from limited context sets with the mathematical rigor of a GP, even in extrapolation regions (Figure \ref{fig:gp-match}). 
An alternative to Bayesian modeling is Ensemble Learning, which uses diversity, either in data, time, or architecture as a proxy for epistemic uncertainty.
A strategy is to train identical architectures on different subsets of data. Hittawe et al. \cite{hittawe2024time} introduce StackPred for climate modeling, where Transformers trained on different cross-validation splits are stacked. The variance between these experts provides an implicit measure of uncertainty, achieving near-perfect correlation on Red Sea datasets. Similarly, Olorunnimbe et al. \cite{olorunnimbe2024ensemble} propose WETT for financial forecasting, which ensembles models trained on sliding temporal windows. By integrating these temporal experts via a quantile-regression meta-learner, WETT generates robust predictive intervals that hold up even during market volatility (Figure \ref{fig:ensemble}).


Transformers can also be used to validate existing ensembles. Bouallegue et al. \cite{bouallegue2024improving} present PoET, a hierarchical Transformer that processes the outputs of numerical weather prediction ensembles. By attending across ensemble members, PoET learns to correct systematic biases, significantly improving probabilistic forecasts (CRPS). Finally, to avoid the prohibitive cost of training multiple large models, recent architectures use Internal Ensembles. DMEformer \cite{liu2023dmeformer} dynamically weights different Transformer variants (Autoformer, Reformer, Informer) within a single framework. Similarly, IELT \cite{xu2023fine} treats individual attention heads as weak learners, employing multi-head voting to mimic an ensemble. These approaches capture structural uncertainty without the computational complexity of traditional methods.

\begin{tcolorbox}[colback=orange!5!white,colframe=red!75!black]
\noindent Transformers have become increasingly prevalent in scientific and physics-informed machine learning (SciFM) models \cite{menon2026scientific}, owing to their ability to capture complex, long-range dependencies and integrate diverse physical data. Yet, deploying these models in high-stakes scientific applications requires more than data-driven performance: it demands trust. Establishing such trust entails moving beyond purely empirical training toward architectures grounded in known physical principles. As discussed in Section~5.1, advances such as operator learning, geometric constraints, and symbolic reasoning enforce physical consistency, while Section~5.2 highlights the need to explicitly quantify uncertainty. By integrating physics-informed design with principled uncertainty estimation, Transformers can achieve not only computational efficiency but also theoretical soundness and probabilistic reliability. This dual emphasis, structural fidelity coupled with calibrated uncertainty, is important for safety-sensitive contexts, where overconfident yet erroneous predictions can have severe consequences.
\end{tcolorbox}

\section{Metrics and Benchmarks for Trustworthiness}

Evaluating the trustworthiness of language models requires more than just measuring accuracy on a test set. Modern applications demand models that are robust, fair, interpretable, and generalizable. However, standard evaluation practices often fail to capture these deeper properties. This section reviews four key works: Schaeffer et al.~\cite{schaeffer2023emergent}, CheckList~\cite{ribeiro2020beyond}, Robustness Gym~\cite{goel2021robustness}, and ARC Challenge~\cite{clark2018think}, which highlight new directions in trustworthiness evaluation. Together, they reveal how metric choice, testing methodology, and benchmark design critically shape our understanding of model behavior.

Schaeffer et al.~\cite{schaeffer2023emergent} challenge the commonly held belief that large language models (LLMs), such as GPT-3, demonstrate sudden \textit{emergent} abilities at specific scales. These capabilities, such as arithmetic reasoning or code generation, often appear abruptly in reported evaluations. The authors argue that this perceived emergence is often an artifact of using coarse, binary metrics like exact match or multiple-choice accuracy. They show that when smoother, continuous metrics, such as Brier score or token-level log-loss, are used instead, model performance improves gradually and predictably with scale. Their analysis, which includes re-evaluating BIG-Bench tasks, shows that over 90\% of previously reported emergent behaviors vanish under these improved metrics. This work highlights the importance of thoughtful metric selection, as it can fundamentally alter our perception of model capabilities.

Building on the need for more insightful evaluation, CheckList~\cite{ribeiro2020beyond} proposes a behavioral testing framework that treats NLP models as black boxes. Inspired by software testing, CheckList organizes evaluation through a matrix combining language capabilities (e.g., negation, vocabulary, coreference) with test types: Minimum Functionality Tests (MFT), Invariance Tests (INV), and Directional Expectation Tests (DIR). For example, MFTs check whether a sentiment model correctly labels '\textit{I love this movie}' as positive, while INVs verify that synonyms or small edits don’t change the model’s output. DIRs test whether adding strongly positive phrases shifts sentiment as expected. Applying CheckList to sentiment analysis and QA models revealed critical failures, such as over 70\% failure on negation tests. In a user study, CheckList helped users write more than twice as many tests and discover nearly three times more bugs compared to traditional methods. This approach encourages systematic diagnosis of model weaknesses beyond aggregate scores.

\begin{figure}[htpb]
    \centering
    \includegraphics[width=0.65\linewidth]{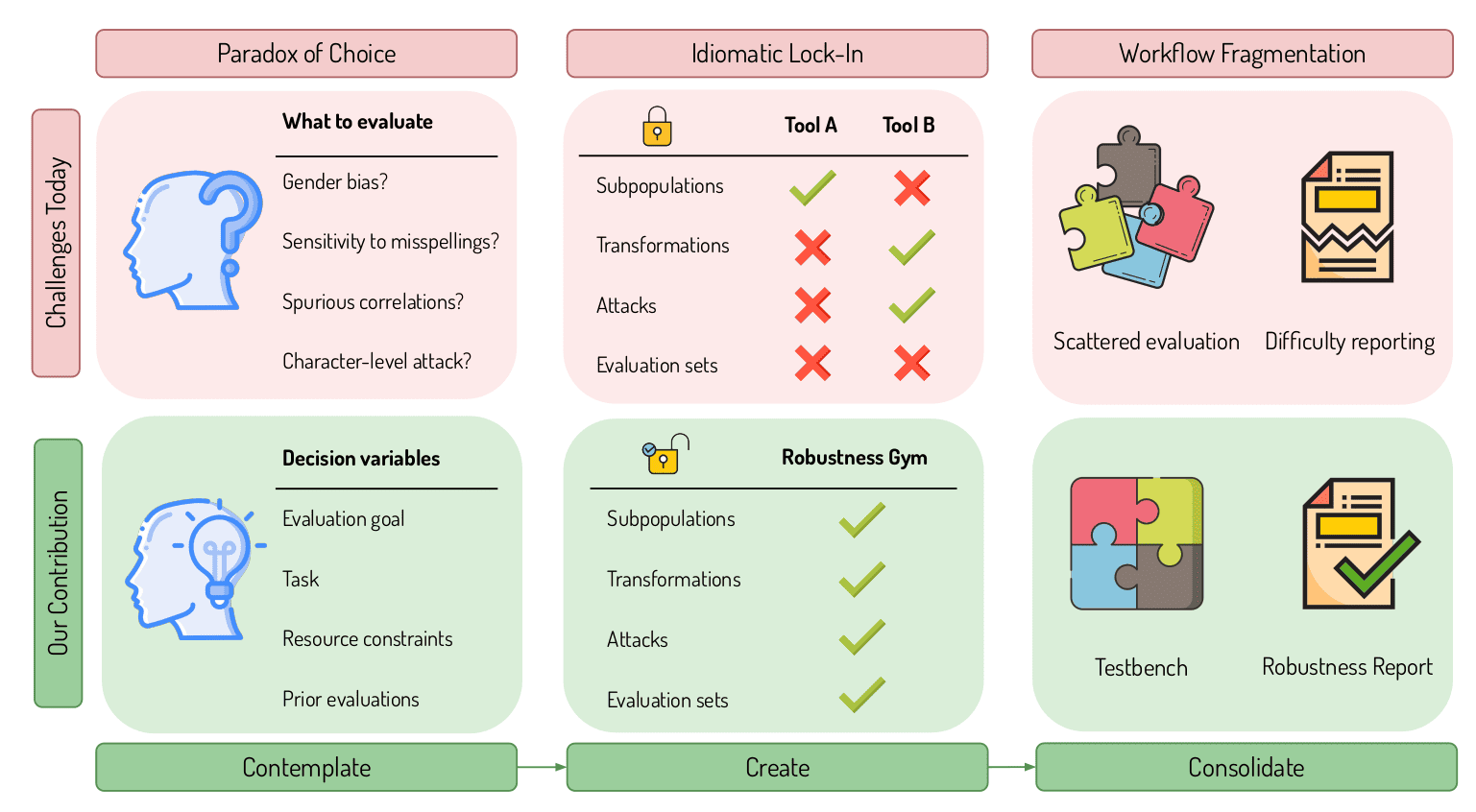}
    \caption{Robustness Gym’s core evaluation loop: Contemplate $\rightarrow$ Create $\rightarrow$ Consolidate. This framework helps users define what to test, construct test slices using transformations or subpopulations, and summarize results in reusable TestBenches. ~\cite{goel2021robustness}}
    \label{fig:robustnessgym_loop}
\end{figure}

While CheckList focuses on manual, structured testing, Robustness Gym~\cite{goel2021robustness} extends these ideas into an automated and modular evaluation framework. It supports four modes of robustness testing: (1) subpopulations (e.g., performance on underrepresented groups), (2) transformations (e.g., typos or rephrasings), (3) adversarial attacks, and (4) curated diagnostic sets. The toolkit follows a loop: Contemplate $\rightarrow$ Create $\rightarrow$ Consolidate. Users define robustness goals, build data slices using abstractions like SliceBuilder, and bundle them into reusable TestBenches. Integrated with platforms like HuggingFace and TextAttack, Robustness Gym simplifies robustness analysis and benchmarking. For example, it uncovered up to 18\% degradation on targeted slices in a sentiment analysis model and exposed weaknesses in named entity linking across commercial systems. This toolkit enables scalable, reproducible trustworthiness evaluations with minimal setup effort. The core loop—Contemplate → Create → Consolidate—is illustrated in Figure~\ref{fig:robustnessgym_loop}, showing how users define evaluation goals, build test slices, and consolidate results into reusable modules.

Complementing these metric and testing frameworks, the AI2 Reasoning Challenge (ARC)~\cite{clark2018think} introduces a benchmark that emphasizes reasoning-based trust. It consists of over 7,700 science questions from real standardized tests (grades 3–9), targeting skills like causal inference, scientific reasoning, and logical deduction. ARC is split into an Easy Set, solvable via shallow pattern-matching, and a Challenge Set, which requires deeper reasoning. Popular models like BiDAF and Decomposable Attention perform barely above random on the Challenge Set, demonstrating the difficulty of true reasoning. ARC also includes a large science corpus and open-source baselines to support further research. By requiring generalizable and knowledge-driven inference, ARC serves as a meaningful benchmark for assessing models in domains where reliable reasoning is important.

Overall, these works demonstrate that evaluating trustworthiness requires more than accuracy; it demands careful metric selection, structured behavioral testing, and reasoning-intensive benchmarks that reflect real-world complexity.

\section{Safety-Critical Applications}
\begin{figure}
    \centering
\includegraphics[width=0.65\linewidth]{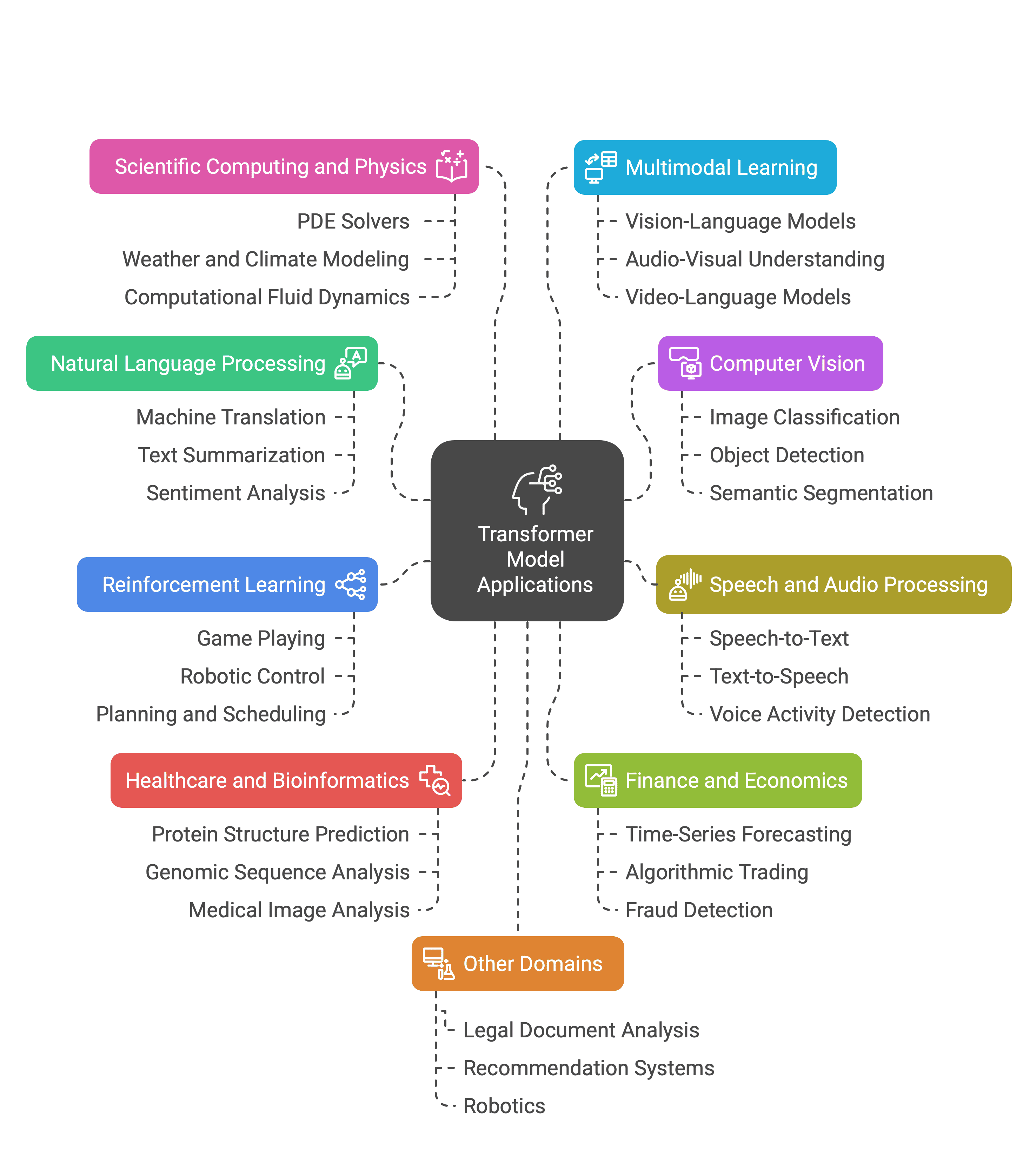}
    \caption{Representative applications across diverse domains where Transformer models have been successfully deployed.}
    \label{fig:TransAppli}
\end{figure}

Transformer architectures have become foundational models across a broad spectrum of safety-critical domains, demonstrating exceptional capacity to learn from high-dimensional, multimodal, and temporally structured data. Their aptitude for capturing long-range dependencies, integrating heterogeneous input modalities, and scaling with data complexity has led to significant advances in applications where precision, robustness, and interpretability are essential. This section examines the expanding role of transformers in domains of critical societal and scientific importance, including natural language processing, computer vision, robotics and autonomous systems, biomedicine, Earth and environmental sciences, materials science, fluid dynamics, and nuclear physics. Figure~\ref{fig:TransAppli} presents a schematic overview of the diverse application domains in which Transformer architectures have been deployed.
Each of these fields presents distinct challenges that demand rigorous evaluation of model generalization, uncertainty quantification, and reliability under real-world constraints.
\subsection{NLP, Vision and Multimodal Learning}

In \textbf{Natural Language Processing (NLP)}, recent progress has been driven by large autoregressive models such as PaLM \cite{chowdhery2023palm}, and LLaMA \cite{touvron2023llama}, which frame reasoning as next-token prediction at massive scale. Models like GPT-4 \cite{achiam2023gpt} use Reinforcement Learning from Human Feedback (RLHF) to better align outputs with human preferences. However, RLHF mainly acts as a post-training behavioral adjustment rather than a structural guarantee of correctness. As a result, these systems can still produce confident hallucinations due to the lack of built-in mechanisms for uncertainty estimation or factual verification. In addition, bias and memorization issues are often addressed only through retrospective audits, leaving the underlying models vulnerable to adversarial prompts that can bypass safety controls.
In \textbf{Computer Vision}, the shift from CNNs to ViTs \cite{dosovitskiy2020image} and hierarchical variants such as the Swin Transformer \cite{liu2021swin} replaces strong inductive biases with global contextual modeling. By representing images as sequences of fixed-size patches, approaches like Masked Autoencoders (MAE) \cite{he2022masked} learn effective representations through reconstruction-based pretraining. However, this patch-based design also introduces new vulnerabilities. Unlike CNNs, which are naturally translation-invariant, ViTs can be sensitive to adversarial perturbations in background regions that are visually irrelevant to humans. Moreover, despite strong benchmark performance, these models typically lack built-in mechanisms for out-of-distribution detection or formal robustness guarantees. Post-hoc explanation methods such as Grad-CAM are also less reliable for ViTs, often failing to accurately identify the features that truly influence model decisions.
A similar trend is observed in \textbf{Audio Processing}, where self-supervised models such as wav2vec 2.0 \cite{baevski2020wav2vec}, HuBERT \cite{hsu2021hubert}, and BEATs \cite{chen2022beats} learn strong speech representations through latent units or semantic features. The Audio Spectrogram Transformer \cite{gong2021ast} further applies the Vision Transformer framework to spectrogram patches. Models like Whisper \cite{radford2023robust} achieve high robustness to noise through large-scale weakly supervised training. However, these systems largely operate as black boxes, with limited evaluation of fairness across accents, dialects, and languages. They also lack reliable confidence calibration to indicate uncertain predictions. As a result, despite strong average performance, their use in high-stakes applications is limited by their inability to assess and communicate failure risk.

Across NLP, vision, and audio, the core Transformer-based advances share a pattern: they optimize heavily for predictive performance through larger models, better pre-training objectives, and architectural refinements, but they pay little attention to integrating trustworthiness into their design. Models like GPT-4 have taken meaningful steps by including RLHF, adversarial testing, and system cards, and some masked self-supervised approaches in vision and audio have incidentally improved robustness. Nevertheless, accuracy and even robustness alone are insufficient as metrics of trust. Few of these models incorporate uncertainty estimation, and even fewer address fairness or interpretability in a rigorous, built-in manner.

The central innovation enabling multimodal learning in Transformer architectures is the alignment and fusion of heterogeneous data types: images, text, audio, video, point clouds, and more into a unified representation space suitable for attention-based processing. This fusion is achieved through a variety of strategies across the literature. The frontier of Transformer research lies in \textbf{Multimodal Learning}, where different data types are mapped into a shared representation space. However, combining modalities also increases trustworthiness challenges by introducing complex interactions between visual, textual, and other biases.

\textbf{Fusion Strategies and Transparency:} Multimodal models differ in how they combine information. Two-stream architectures such as ViLBERT \cite{lu2019vilbert} and LXMERT \cite{tan2019lxmert} encode images and text separately and fuse them using co-attention, allowing partial interpretability through cross-modal alignment. In contrast, unified single-stream models like Flamingo \cite{alayrac2022flamingo} directly integrate visual tokens into language sequences. While this improves few-shot reasoning, it reduces transparency, making it difficult to identify the source of errors or hallucinations.

\textbf{Contrastive Alignment and Bias:} Models like CLIP \cite{radford2021learning} align image and text embeddings using contrastive learning, enabling strong zero-shot performance. However, training on large web-scale datasets leads to the inheritance and amplification of societal biases. Methods such as BLIP-2 \cite{li2023blip} attempt to bridge vision encoders and language models efficiently, but the use of frozen backbones limits the ability to correct underlying biases.

\textbf{Generative and Unified Risks:} Multimodal generative models, including AV-DiT \cite{wang2024av} and AV-Transformer \cite{lin2020audiovisual}, ensure cross-modal consistency but lack safeguards against misuse, such as deepfake generation. Fully unified systems like Meta-Transformer \cite{zhang2023meta} map multiple modalities into a shared token space. While flexible and scalable, this design creates a potential single point of failure, where biases or vulnerabilities in the shared encoder can affect multiple downstream applications.

\begin{figure}[htpb]
    \centering
    \includegraphics[width=0.65\textwidth]{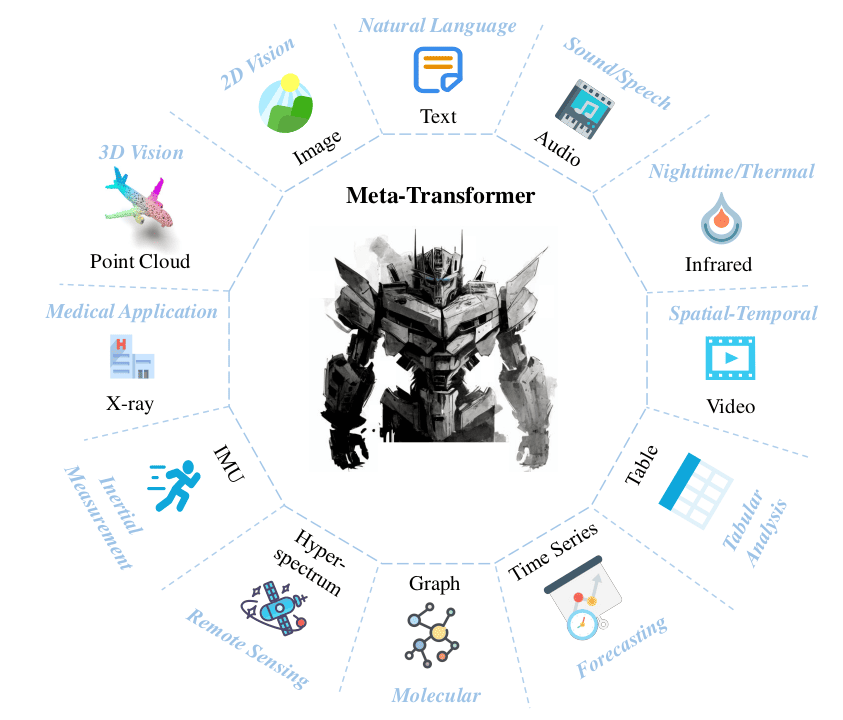}
    \caption{Illustration of the Meta-Transformer framework showcasing its capability to process and unify information across 12 diverse modalities, including text, image, video, audio, point cloud, graphs, and medical data. This conceptual depiction highlights the potential of a shared transformer backbone for achieving general-purpose multimodal intelligence. Adapted from \cite{zhang2023meta}}
    \label{fig:meta_transformer}
\end{figure}

\begin{tcolorbox}[colback=orange!5!white,colframe=red!75!black]

\noindent Despite significant architectural convergence, none of the reviewed multimodal works embed trustworthiness directly into their core training or evaluation paradigms.
\begin{itemize}
    \item \textbf{Uncertainty:} Models like Flamingo and CLIP lack epistemic uncertainty quantification, making them prone to confident misclassifications in zero-shot settings.
    \item \textbf{Bias:} Unified models like Meta-Transformer risk propagating latent biases across 12+ modalities without modality-specific auditing.
    \item \textbf{Robustness:} Frozen backbones in BLIP-2 and adapters in AV-DiT create rigid dependencies where upstream vulnerabilities (e.g., adversarial patches in the vision encoder) cannot be corrected by downstream alignment.
\end{itemize}
Without mechanisms for interpretability, uncertainty quantification, or robustness to adversarial inputs, even the most capable multimodal Transformers cannot be deemed fully trustworthy for safety-critical deployment.
\end{tcolorbox}

\subsection{Robotics \& Autonomous Driving}
Unlike virtual agents, autonomous systems operate in physical environments where failures have real-world consequences. Tasks such as autonomous driving or robotic surgery require real-time decision-making under uncertainty. Transformers are widely used in this domain due to their ability to fuse multiple sensor modalities (e.g., LiDAR, cameras, proprioception) and capture long-term dependencies. However, high predictive performance does not guarantee physical safety. Current models mainly focus on improving detection and control accuracy, while often lacking explicit safety mechanisms to handle sensor noise, adversarial inputs, or out-of-distribution scenarios.

\begin{figure}[htpb]
    \centering
    \includegraphics[width=0.7\textwidth]{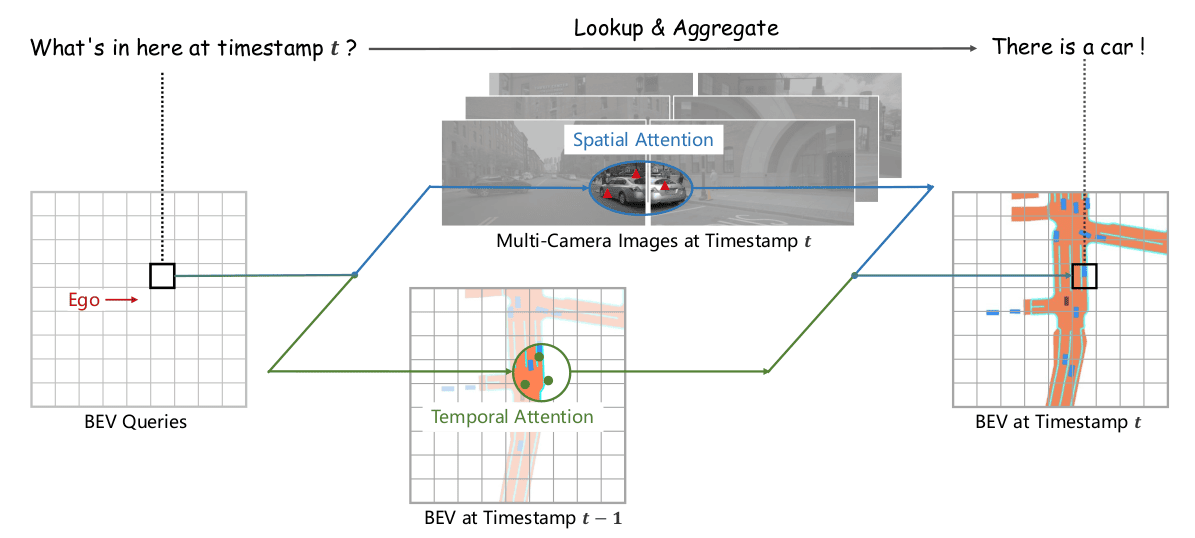}
    \caption{Overview of BEVFormer framework for generating bird’s-eye-view representations using spatial and temporal Transformer attention from multi-camera inputs. Adapted from Li et al. \cite{li2203bevformer}}
    \label{fig:bevformer}
\end{figure}

In autonomous driving, the key challenge is scene understanding, building a consistent 3D view from fragmented 2D sensor data. BEVFormer \cite{li2203bevformer} addresses this using spatial cross-attention to project multi-camera images into a unified Bird’s-Eye-View (BEV), along with temporal self-attention to capture motion over time. As shown in Figure \ref{fig:bevformer}, learnable BEV queries gather spatial features from camera views and align them with past frames. This design improves robustness to occlusions, allowing the model to infer object locations even when they are temporarily hidden. However, the BEV output is deterministic and lacks uncertainty estimates. Without confidence information, downstream planners cannot distinguish between reliable detections and artifacts caused by sensor noise or temporal misalignment.

This limitation becomes more critical in collaborative settings. V2X-ViT \cite{xu2022v2x} extends perception beyond a single vehicle by fusing information from other vehicles and roadside infrastructure using multi-agent attention. Although the model shows robustness to GPS errors and communication delays, this reliability is implicit. It operates as a black box, combining inputs without estimating the confidence of each agent’s information. In safety-critical situations, the planner cannot distinguish between a true obstacle confirmed by multiple agents and a false detection caused by a single noisy source, reducing the practical safety benefits of cooperative perception.

\begin{figure}[htpb]
    \centering
    \includegraphics[scale=0.25]{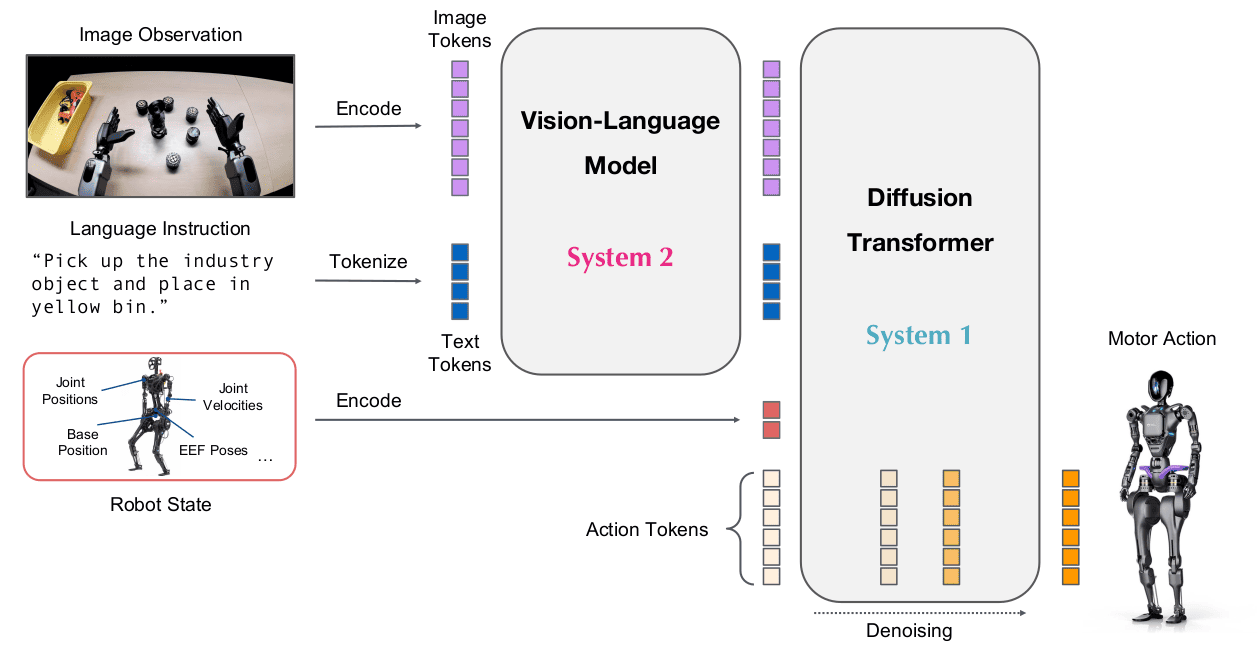}
    \caption{GR00T N1’s dual Transformer system combining a Vision-Language Model with a Diffusion Transformer for control. Adapted from Bjorck et al. \cite{bjorck2025gr00t} }
    \label{fig:groot}
\end{figure}

The move toward Generalist Robot Policies has led to the use of Transformers directly for control, where the model maps sensor observations straight to motor actions. DriveTransformer \cite{jia2025drivetransformer} is an example of this idea in autonomous driving. Instead of using separate modules for perception, prediction, and planning, it replaces the traditional pipeline (perception to prediction to planning) with a single end-to-end network. It uses sparse tokens and a streaming temporal memory so that it can process long sequences efficiently and plan over long time horizons. However, this fully unified design removes the intermediate outputs that normally help with debugging and interpretation. If the vehicle suddenly swerves, there is no separate object detection or prediction output to examine. This makes it very difficult to determine whether the error came from incorrect perception, poor prediction, or a flawed control decision.
In robotics, another approach to generalization has been to scale model size and data. GR00T N1 \cite{bjorck2025gr00t} follows this strategy and introduces a dual-system architecture inspired by human cognition, separating high-level reasoning from low-level motor control. As shown in Figure \ref{fig:groot}, the system is divided into two different Transformer components. First is System 1 (Action), where a Diffusion Transformer uses the tokens from the VLM to generate detailed, high-frequency motor commands such as joint positions or velocities. And second is System 2 (Reasoning), where a Vision-Language Model (VLM) that takes an image and a natural language instruction (for example, '\textit{Pick up the industrial object}'). It converts these inputs into tokens and produces a high-level plan. This separation helps GR00T generalize across different types of robots, such as humanoids and robotic arms. However, the connection between System 2 and System 1 creates an important risk. The VLM functions as a black-box semantic planner and does not provide explicit safety constraints. If the VLM misinterprets the scene or generates an unsafe instruction for example, confusing a human hand with an \textit{industrial object} the Diffusion Transformer will still execute the resulting motion. There is no symbolic safety check or formal verification layer between the reasoning and action stages to prevent such unsafe behavior.

\begin{tcolorbox}[colback=orange!5!white,colframe=red!75!black]
\noindent While Transformer architectures have enabled powerful capabilities in sensor fusion and long-term decision making, a closer look shows a clear lack of built-in safety and reliability mechanisms:
\begin{itemize}
\item \textbf{Unquantified Uncertainty:} Models such as \textit{BEVFormer} and \textit{RT-1} produce fixed, deterministic outputs. They do not estimate epistemic uncertainty (for example, Bayesian attention or conformal prediction) to indicate when the model is unsure or encountering out-of-distribution (OOD) inputs.
\item \textbf{Opaque Decision-Making:} End-to-end systems like \textit{DriveTransformer} and \textit{GR00T} hide the internal reasoning process between perception and action. Without interpretable intermediate outputs, it becomes very difficult to identify whether a failure was caused by a sensor issue, perception error, or poor decision logic.
\item \textbf{Implicit Instead of Explicit Safety:} Some architectures, such as \textit{V2X-ViT}, rely on large-scale data to handle noise and variability implicitly. However, none of the reviewed models include explicit safety mechanisms, such as formal verification layers, rule-based safety filters, or guaranteed safe action bounds to prevent dangerous behavior when the model makes an incorrect prediction or hallucination.
\end{itemize}
Overall, current robotic Transformer systems focus more on improving generalization and performance than on ensuring operational safety. For real-world deployment in human environments, future designs must move beyond black-box control and become \textit{auditable agents} systems that can explain their decisions, estimate their confidence, and reject actions that may be unsafe.
\end{tcolorbox}

\subsection{Biology and Medicine}
The adaptation of Transformers to biology represents a paradigm shift from manual feature engineering to data-driven representation learning. However, biology differs fundamentally from natural language: data is high-dimensional, strictly constrained by physical laws, and often sparse. Consequently, \textit{trustworthiness} in this domain is not just about prediction accuracy, but about biological plausibility and actionable confidence. A model that predicts a protein structure without a confidence score is clinically useless, regardless of its average accuracy. This section evaluates how Transformers navigate these challenges across three distinct biological modalities: 1D sequences, 3D geometric structures, and high-dimensional tabular profiles. 

\begin{figure}[htpb]
    \centering
    \includegraphics[width=1.0\linewidth]{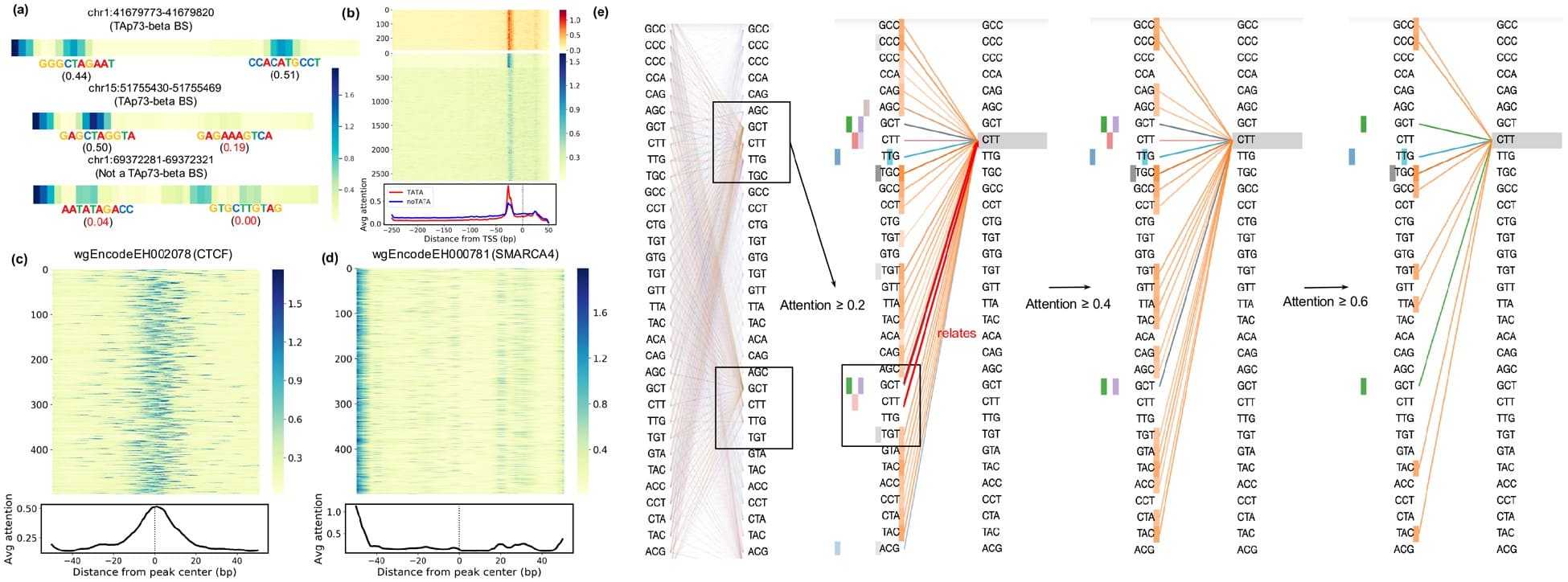}
    \caption{Interpretability of DNABERT via attention maps on genomic sequences:
(a–b) Heatmaps depict average attention scores across promoter and transcription start site regions.
(c–d) Aggregate attention signals over transcription factor, showing peak-centered activation.
(e) Attention pathways over 6-mer tokens in DNA sequences at increasing thresholds (0.2, 0.4, 0.6) highlight how DNABERT captures dependencies between nucleotide motifs (source: Ji et al.\cite{ji2021dnabert}).}
    \label{fig:dnabert}
\end{figure}

Biological sequences (DNA, RNA, Proteins) share linguistic properties, making them natural candidates for BERT-style architectures. DNABERT \cite{ji2021dnabert} adapts this by tokenizing DNA into overlapping $k$-mers ($k=3$ to 6). By pretraining on the human reference genome, it learns a \textit{grammar} of nucleotides that generalizes to promoter detection and splice site prediction. Similarly, ESM-1b \cite{rives2021biological} scales this to proteins, training on 250 million sequences to capture evolutionary patterns that correlate with secondary and tertiary structures. The primary trust mechanism in these sequence models is post-hoc interpretability via attention visualization. As shown in Figure \ref{fig:dnabert}, DNABERT's attention heads spontaneously align with known biological motifs (e.g., transcription factor binding sites), providing a form of validation that the model is learning \textit{real} biology rather than artifacts. However, this interpretability is implicit. Neither model incorporates explicit uncertainty quantification or robustness checks against sequencing noise. They rely on the assumption that high attention weights equal causal importance, a heuristic that can be misleading in safety-critical genomic editing. 

One of the most impactful applications of Transformers in biology is AlphaFold \cite{jumper2021highly}, which revolutionized protein structure prediction. Sequence models treat biological data like text, but real biological function happens in three-dimensional space. For models in this area to be trustworthy, they must follow geometric rules such as bond lengths, angles, and atomic distance constraints. AlphaFold addressed this challenge by going beyond standard attention mechanisms. Its specialized Evoformer block uses triangle multiplicative updates and axial attention to explicitly capture the spatial relationships between amino acid residues. Importantly, AlphaFold also tackles the \textit{Confidence Gap} by providing a confidence score for each residue, called pLDDT. Instead of giving only a single structural prediction, the model shows how reliable each part of the structure is. This makes the output actionable, allowing biologists to identify which regions are trustworthy and which may be flexible or uncertain. This type of built-in self-evaluation sets a strong benchmark for trustworthiness in scientific AI.

\begin{figure}[htpb]
    \centering
    \includegraphics[width=0.65\linewidth]{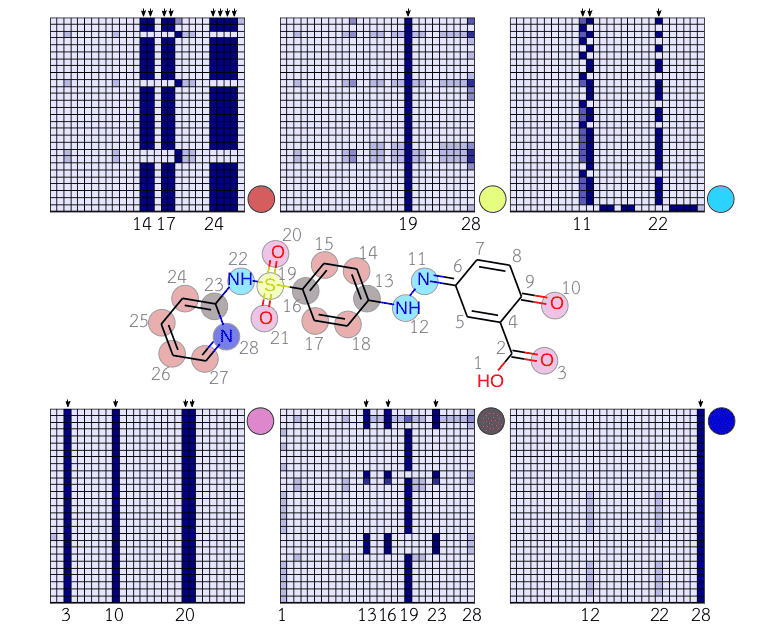}
    \caption{Attention visualization from MAT:
Attention maps for different heads in the Transformer model are shown around a molecule, highlighting the atomic interactions each attention head focuses on. Darker blue cells represent higher attention weights between atom pairs. This visualization demonstrates the interpretability of MAT, as chemically significant substructures (such as aromatic rings, polar groups, or connected functional units) are emphasized by the model’s attention mechanism (Source: Maziarka et al.\cite{maziarka2020molecule})}
    \label{fig:mat}
\end{figure}

In chemical informatics, Transformers have been adapted to graph-structured data to predict molecular properties. The Molecule Attention Transformer (MAT) \cite{maziarka2020molecule} is a prime example, incorporating distance and adjacency biases into the attention mechanism. MAT introduces \textit{structural bias} into the attention mechanism. Standard Transformers are graph-agnostic, treating atoms as a '\textit{bag of words}'. MAT injects inter-atomic distances and bond types directly into the attention computation, forcing the model to respect the molecular graph topology. As seen in Figure \ref{fig:mat}, this allows the model to focus on chemically significant substructures (e.g., aromatic rings) when predicting properties like toxicity. However, unlike AlphaFold, MAT lacks a formal uncertainty output, limiting its reliability in drug discovery pipelines where false positives are costly.

Gene expression data brings a different type of challenge. It is not a sequence or a graph, but a high-dimensional and unordered table containing expression levels for more than 20,000 genes. \textbf{T-GEM} \cite{zhang2022transformer} adapts Transformers to this setting by treating each gene as a token and its expression level as the corresponding embedding. The model shows what can be called emergent interpretability. Analysis of its attention patterns shows a meaningful hierarchy: early layers tend to focus on broadly expressed housekeeping genes, while deeper layers concentrate on phenotype-specific regulatory signals, such as cancer-related markers. This layered processing reflects the multi-level organization of biological systems. However, despite this promising interpretability, T-GEM does not include mechanisms to handle batch effects, distribution shifts that occur between different experiments or laboratories. Since such shifts are common in transcriptomic data, this lack of robustness creates a significant limitation for reliable clinical use.

\begin{tcolorbox}[colback=orange!5!white,colframe=red!75!black]
\noindent Transformer-based architectures are being actively explored across biological subfields, from molecular property prediction to protein folding and transcriptomic classification. Several models, such as MAT, T-GEM, and DNABERT, incorporate attention visualizations to support interpretability, contributing to trustworthiness in a limited scope. Models like AlphaFold include confidence scores wthat partially address uncertainty, while others like ESM-1b rely on pretrained embeddings with implicit generalizability. However, a gap in nearly all reviewed works is the lack of explicit trustworthiness mechanisms such as: Uncertainty quantification, Bias detection and correction, Robustness evaluation etc. As biology intersects increasingly with clinical and therapeutic applications, embedding these trustworthiness dimensions into Transformer architectures becomes not optional but necessary. Without them, even high-performing models remain opaque and potentially unsafe for translational use. Future work should thus aim to pair architectural innovations with systematic evaluation of model reliability, fairness, and interpretability.

\end{tcolorbox}

\subsection{Earth Science}
In Earth Science, the distinction between a correct prediction and a trustworthy one is often measured in human lives. Whether mapping post-disaster landslides or forecasting extreme weather, AI models operate in high-stakes environments where decision-makers require not just a prediction, but a guarantee of reliability. Transformers have rapidly been adopted in this domain for their ability to model hierarchical spatiotemporal features in high-dimensional data. However, a critical review shows a bifurcated landscape: while some models achieve \textit{interpretability by design} by aligning attention with physical signals, the majority function as opaque feature extractors that lack the uncertainty quantification necessary for disaster response.
In the domain of hazard assessment and resource exploration, Transformers are mainly used as strong but hard-to-interpret segmentation tools. For landslide detection, Wu et al. \cite{wu2024landslide} propose SCDUNet++, a hybrid model that combines CNNs and Transformers. The CNN part captures local texture details, while the Transformer captures long-range spatial relationships. Through a Global-Local Feature Extraction block, the model can accurately detect post-earthquake landslides from Sentinel-2 satellite images. While the use of Deep Transfer Learning improves its ability to adapt to new regions, which is very important for fast disaster response, the model does not provide any estimate of prediction uncertainty. In real rescue situations, this limitation is critical: a false negative (missing a landslide) can be dangerous, yet the system gives no confidence score to warn decision-makers about uncertain or ambiguous areas. Similarly, Bao et al. \cite{bao2022application} use the Swin Transformer for Landslide Susceptibility Mapping. Its shifted-window design helps the model capture terrain patterns at multiple scales, making it more effective than traditional CNNs with fixed kernels. However, the study mainly uses the Transformer to improve accuracy and does not investigate why a particular slope is labeled as high-risk. It also does not test how robust the model is to noisy or imperfect remote sensing data. A similar situation appears in subsurface exploration. Wang et al. \cite{wang2023seismic} introduce U-Segformer-Hyper for seismic facies classification. The model uses a Hypercolumn strategy to combine features from different scales, improving representation quality. However, the internal decision process is still difficult to interpret. In high-cost hydrocarbon exploration, using a model that cannot clearly explain how it combines and uses features reduces trust and limits its practical deployment.

\begin{figure}[htpb]
\centering
\includegraphics[width=0.65\linewidth]{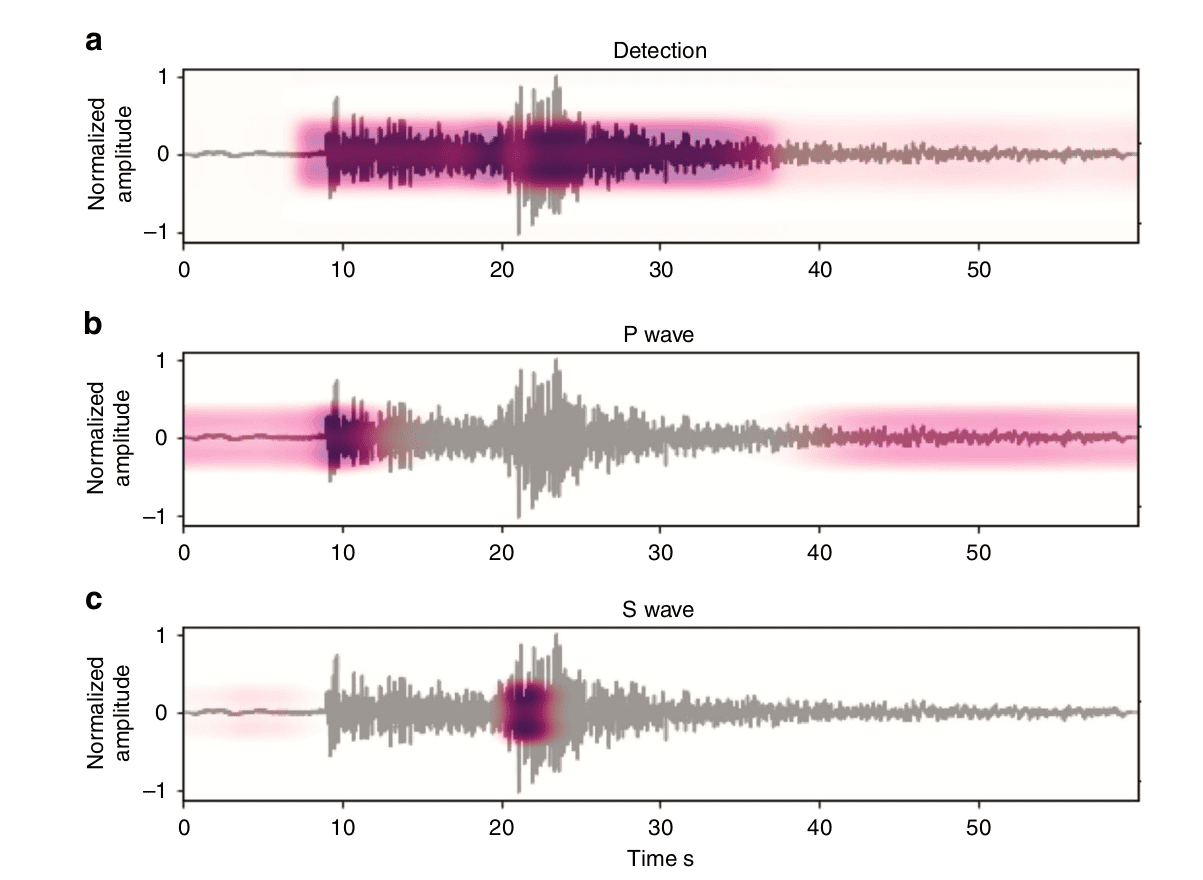}
    \caption{Attention weight visualizations from the Earthquake Transformer model for a single seismic waveform. (a) Detection task: global attention highlights the full earthquake signal region. (b) P-wave picking: attention sharply focuses on the early-arriving P-phase onset. (c) S-wave picking: attention localizes around the later S-phase arrival. These task-specific attention maps illustrate the model’s interpretability by revealing which parts of the waveform influence each prediction. Adapted from Mousavi et al. \cite{mousavi2020earthquake}}
    \label{fig:eq_transformer}
\end{figure}

In contrast to these opaque mapping models, the Earthquake Transformer proposed by Mousavi et al. \cite{mousavi2020earthquake} stands out as a strong example of a trustworthy design. Built for simultaneous event detection and phase picking, the model uses a hierarchical attention structure that reflects how human experts analyze seismic signals. The global attention first examines the entire waveform to determine whether an earthquake signal is present or if the data contains only background noise. Then local attention focuses on smaller regions of the waveform to accurately identify the arrival times of P-phases (primary waves) and S-phases (secondary waves). Most importantly, the model provides a form of self-validation. As illustrated in Figure \ref{fig:eq_transformer}, the attention weights are not only linked to model performance but also align with the physical structure of the seismic signal, showing sharp peaks exactly at the P and S wave arrival times. This physical consistency turns the attention mechanism from a purely mathematical operation into an interpretable signal analysis tool, increasing confidence and trust among seismologists.

The final frontier is spatiotemporal forecasting, where the objective is to predict how complex and chaotic systems like weather evolve over time. Earthformer \cite{gao2022earthformer} introduces a method called \textit{Cuboid Attention}, which divides volumetric data into small 3D blocks to capture both local patterns and large-scale interactions for tasks such as precipitation nowcasting. In a similar direction, Stormer \cite{nguyen2024scaling} uses \textit{randomized dynamics} along with a pressure-weighted loss function to produce accurate medium-range weather forecasts (5-10 days) while keeping computational cost low. Although these models are much faster than traditional numerical weather solvers, they face a major limitation known as the \textit{Confidence Gap}. Weather prediction is naturally uncertain, and providing a completely deterministic forecast is scientifically unrealistic. However, neither Earthformer nor Stormer includes explicit UQ or probabilistic ensemble outputs. In real-world decision-making situations, such as planning evacuations, the range of possible outcomes and their likelihoods is often more important than a single predicted scenario. Without this uncertainty information, these Transformer-based systems remain powerful and efficient, but they are not yet reliable enough for high-stakes operational decisions.

\begin{tcolorbox}[colback=orange!5!white,colframe=red!75!black]
\noindent In sum, transformer models have markedly advanced Earth-science tasks: from SCDUNet++ for landslide mapping and ViT/Swin architectures for susceptibility analysis to U-Segformer-Hyper in seismic facies classification, Earthformer in precipitation nowcasting, and Stormer in medium-range weather forecasting by effectively capturing long-range spatial and temporal dependencies. However, only the Earthquake Transformer explicitly incorporates interpretability via attention-weight visualizations that align with seismological expertise. None of the other studies integrate critical trustworthiness components, such as uncertainty quantification, bias assessment, or robustness checks, which are essential for high-stakes applications in disaster response, public safety, and resource exploration. Without these, even top-performing transformer models have the risk remaining and may not gain the operational confidence needed for deployment. Future work must therefore pair architectural innovation with explicit mechanisms for explainability, reliability, and accountability.
\end{tcolorbox}

\subsection{Material Science}
Materials science is going through a major shift, moving from heuristic discovery to data-driven design. Transformers are playing an important role in this change because they provide a single architecture that can handle different types of data, such as 3D crystal structures, 1D spectral signals, and symbolic chemical formulas. However, the risks in this field are very different from those in language tasks. In language models, a hallucination may only produce an incorrect or awkward sentence. In materials science, a hallucination could mean generating a physically unstable material or even a toxic compound. For this reason, the trustworthiness of these models depends on two critical factors: whether they follow fundamental physical constraints and whether they can explain the reasoning behind their design choices. At present, these reliability and interpretability capabilities are still inconsistent across different approaches in the field.

\begin{figure}[htpb]
    \centering
    \includegraphics[width=0.9\textwidth]{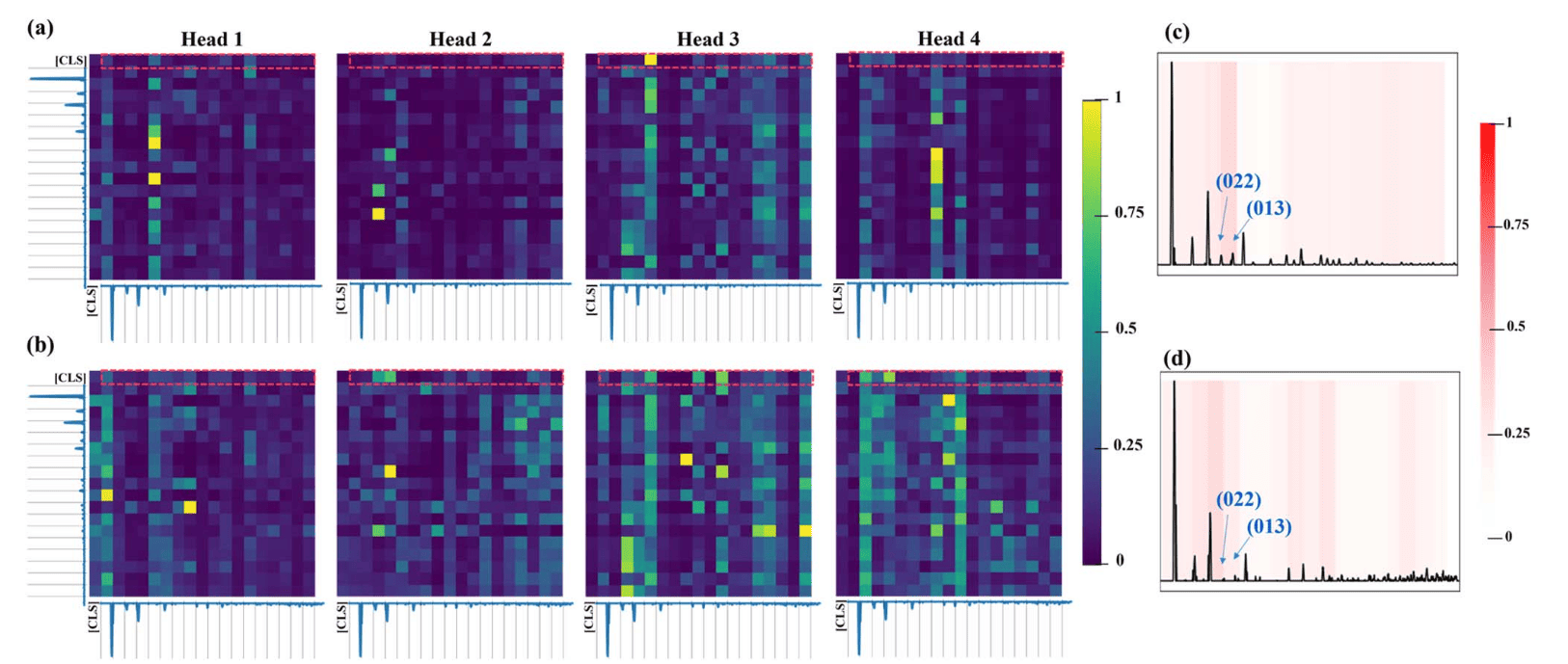}
    \caption{Visualization of attention weights in the final layer of the ViT-XRD model. (a, b) Attention head-specific heatmaps for XRD spectra of ZIF-8 and ZIF-67 show how different heads focus on distinct regions of the spectrum, capturing minor peaks often overlooked by CNNs. (c, d) Attention rollout maps aggregate attention from the [CLS] token, indicating that ViT models emphasize subtle spectral variations such as (022)/(013) peaks, which are critical for distinguishing highly similar materials. Adapted from Chen et al. \cite{chen2024interpretable}}
    \label{fig:chen_attn}
\end{figure}

A challenge in materials science is representation: how to convert physical structures into tokens that a model can understand. Yu et al. \cite{yuunified} address this using the Unified Material Transformer (UMT), which converts 3D crystal structures into voxel-based spatial grids. Unlike Graph Neural Networks (GNNs), which often struggle to capture global symmetry, UMT uses positional encodings to model long-range atomic interactions. Here, trustworthiness appears indirectly: the learned embeddings naturally organize according to periodic table trends, even without supervision. This suggests that the Transformer is capturing meaningful physical patterns rather than only fitting statistical correlations. In material characterization, Transformers also provide an interpretability advantage over CNNs. Chen et al. \cite{chen2024interpretable} show this by using ViT to classify X-ray Diffraction (XRD) and FTIR spectra. CNNs often overfit to the strongest spectral peaks, but the ViT’s self-attention spreads focus across smaller, high-frequency features. As illustrated in Figure \ref{fig:chen_attn}, the attention heads (a, b) and rollout maps (c, d) highlight subtle peaks, such as the $(022)/(013)$ reflections, that are important for distinguishing similar polymorphs like ZIF-8 and ZIF-67. This fine-grained attention helps the model transfer knowledge across different measurement types (from XRD to FTIR), providing a level of robustness that CNNs were unable to achieve.

The highest risks appear in Generative Design, where models suggest new material compositions. In this setting, trustworthiness depends on balancing creativity with physical validity. The BLMM Crystal Transformer by Wei et al. \cite{wei2204crystal} improves trust by framing generation as a \textit{blank-filling} process. Instead of producing a black-box result, the model gradually inserts elements into a template using domain-guided actions such as doping or substitution. This makes the process partly interpretable, chemists can follow each step and check that important physical constraints, like charge neutrality and electronegativity balance, are satisfied throughout the generation. In contrast, models such as MatterGPT \cite{chen2024mattergpt} and the text-to-material approach by Yang and Buehler \cite{yang2021words} focus on scale and multi-objective generation. MatterGPT uses SLICES encoding to generate crystal structures conditioned on properties like band gap, and it can produce designs beyond the training data, which suggests some level of generalization. However, these autoregressive models do not provide formal UQ. When MatterGPT proposes a new crystal, it gives no confidence estimate about its thermodynamic stability. Similarly, the CLIP-based pipeline of Yang and Buehler generates visually reasonable lattice structures from text descriptions, but without validation from a physics simulator, these outputs remain unverified and may represent physically unrealistic \textit{hallucinations}.

\begin{tcolorbox}[colback=orange!5!white,colframe=red!75!black]
\noindent A review of generative material Transformers shows a major gap in validation:
\begin{itemize}
\item \textbf{Implicit Bias:} Benchmark studies by Fu et al. show that different architectures (such as GPT and BERT) tend to favor generating certain types of chemistries over others. Without proper bias analysis, these models may overproduce specific material classes, which can distort and limit the material discovery process.
\item \textbf{Missing UQ:} None of the reviewed generative models provide epistemic uncertainty estimates. In experimental research, synthesizing a prediction without knowing its confidence level can lead to wasted time, effort, and resources.
\item \textbf{Physical Grounding:} Although BLMM applies chemical constraints during generation, most large foundation models do not include physics-informed loss functions. Instead, they rely mainly on matching patterns in the training data.
\end{itemize}
To move forward, future models must combine strong generative ability with reliable, physics-based uncertainty estimation.

\end{tcolorbox}

\subsection{Fluid Flows}
Fluid dynamics is dominated by chaotic turbulence and complex multi-scale interactions, making traditional numerical simulations extremely expensive and time-consuming. Transformers have emerged as promising \textit{Neural Surrogates} in this context, leveraging self-attention to capture long-range correlations and anisotropic features within turbulent flows. However, substituting established numerical solvers with deep learning models introduces notable risks. Unlike classical solvers, which conserve mass and momentum up to discretization errors, a Transformer can \textit{hallucinate} unphysical states, such as negative densities or energy violations, if appropriate constraints are not applied. Therefore, the trustworthiness of these neural surrogates critically depends on their ability to enforce physical consistency while also providing measures of uncertainty for their predictions.

The drive to build foundation models for physics has produced architectures like POSEIDON \cite{herde2024poseidon}, which uses a Scalable Operator Transformer to generalize across different flow regimes. Similarly, Universal Physics Transformers (UPT) \cite{alkin2024universal} unify Eulerian (grid-based) and Lagrangian (particle-based) simulations by encoding both into a shared latent space. These models demonstrate impressive sample efficiency and computational speed, but they largely function as black boxes. They do not inherently enforce physical conservation laws, relying instead on purely data-driven loss functions. More importantly, they lack intrinsic UQ. For example, if POSEIDON encounters an out-of-distribution scenario, such as a Reynolds number outside the training range, it offers no confidence metric to alert engineers that its prediction may be unreliable. Some specialized methods, like Peng et al. \cite{peng2022attention}, introduce inductive bias by enhancing Fourier Neural Operators with spatial attention. This allows the model to adaptively focus on regions of high non-equilibrium, such as turbulent eddies. However, the focus remains on predictive accuracy rather than verification; there is still no guarantee that the attention truly corresponds to physically meaningful energy cascades, leaving trustworthiness in critical applications unaddressed.

To achieve both global awareness and local detail, researchers have explored hybrid architectures. Kang et al. \cite{kang2023new} combine a Vision Transformer as a global encoder with a U-Net as a local refiner to model flow around obstacles. While this approach produces visually accurate flow predictions, there is no investigation into whether the attention heads actually follow meaningful physical structures, such as vortices or boundary layers. A more promising path for trustworthy modeling is AeroDiT \cite{zheng2024aerodit}, which couples Swin-style attention with Diffusion Models for aerodynamic flows. Unlike deterministic regression models, AeroDiT is generative: for a single input, it can produce multiple plausible flow field realizations. This naturally introduces variability across samples, providing an implicit mechanism to capture aleatoric uncertainty and reflecting the intrinsic stochasticity of turbulent flows.

\begin{figure}[htpb]
    \centering
    \includegraphics[width=0.75\textwidth]{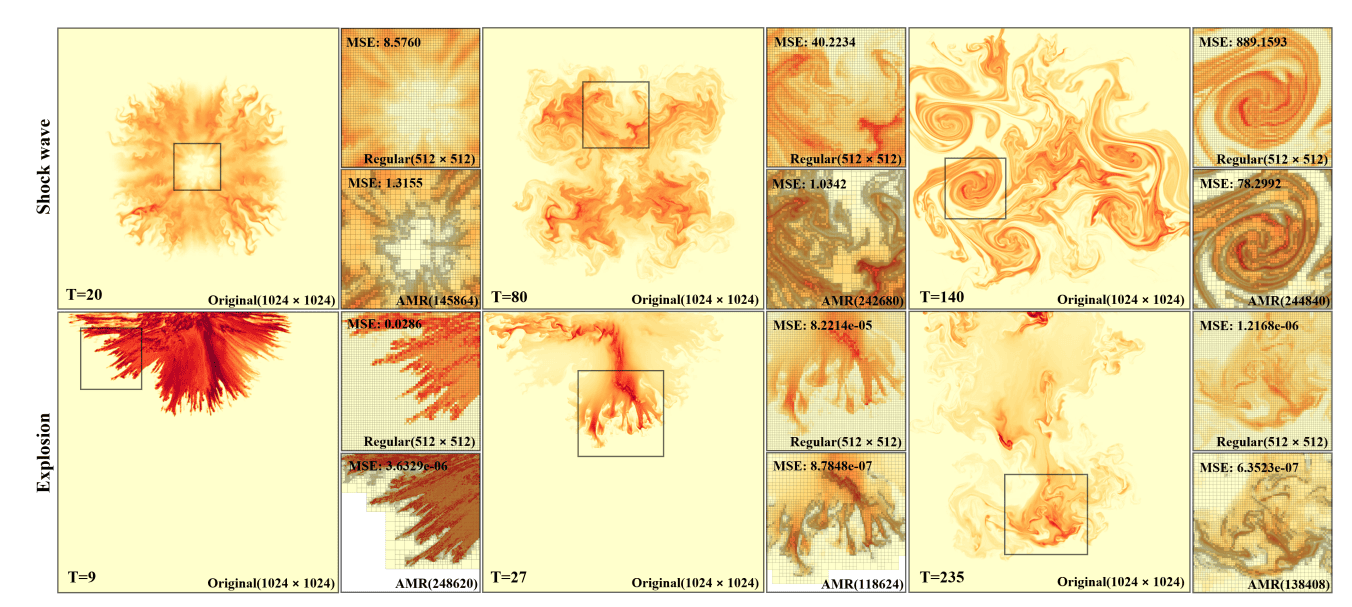}
    \caption{Visualization of Adaptive Mesh Refinement (AMR) tokenization versus regular grid discretization for high-resolution fluid simulations ($1024\times1024$). The AMR approach selectively refines regions with complex flow features, significantly reducing token count while maintaining lower error rates compared to uniform grids. Source: Xu et al. \cite{xu2025amr}.}
    \label{fig:amr_token}
\end{figure}

Trustworthiness also depends on how models manage spatial resolution. Handling fine computational grids is costly, which motivated the development of the AMR-Transformer \cite{xu2025amr}. This model uses a \textit{physically motivated tokenizer} based on Adaptive Mesh Refinement (AMR). Instead of a uniform grid, it allocates dense tokens to regions with complex dynamics (e.g., shock fronts) and sparse tokens to calm areas, reducing computational cost by up to $60\times$. As shown in Figure \ref{fig:amr_token}, the AMR scheme (right) dynamically assigns resources to high-gradient features, achieving the same fidelity as a fully dense grid (left) with far fewer tokens. However, this efficiency introduces a new source of \textit{epistemic uncertainty}: token pruning can create errors. Without explicit error bounds, it is unclear whether the coarsened mesh might miss subtle but important flow instabilities that could trigger turbulence or other critical transitions.

\begin{tcolorbox}[colback=orange!5!white,colframe=red!75!black]
\noindent In safety-critical scenarios, whether simulating airflow over a wing or modeling turbulent coolant in a reactor, understanding what a model doesn’t know is just as important as its raw accuracy. Today’s fluid surrogates are excellent at reconstructing flows but often lack essential trustworthiness features:
\begin{itemize}
 \item \textbf{Uncertainty Quantification:} Models fail to indicate low confidence in turbulent or out-of-distribution regimes, as seen in \textit{POSEIDON} \cite{herde2024poseidon}   and \textit{UPT} \cite{alkin2024universal}.
 \item \textbf{Physical Consistency:} Predictions may violate basic physical laws (e.g., negative density), since purely data-driven losses cannot enforce constraints.
 \item \textbf{Interpretability:} Attention weights are rarely mapped to identifiable flow structures like vortices, a gap highlighted in \textit{Kang et al.} \cite{kang2023new}.
 \end{itemize}
Without these safeguards, even surrogates with high reconstruction accuracy cannot be safely used in automated design loops or control systems.
\end{tcolorbox}

\subsection{Nuclear Science}
In the hierarchy of safety-critical fields, nuclear engineering represents one of the most demanding environments. Whether monitoring reactor thermal hydraulics or controlling plasma in a fusion tokamak, decision systems must operate with near-zero tolerance for error, and uncertainty must be explicitly measured. Transformers are increasingly used in this domain because they can capture long-term patterns in complex sensor data. However, recent studies show a clear divide: some models include built-in safety features such as uncertainty estimation, while others act as highly accurate but opaque predictors, which raises concerns about their safe deployment.

One particularly illustrative example is the application of the Temporal Fusion Transformer (TFT) \cite{li2022long} for accident prognosis in nuclear reactors. This model is used to predict key safety parameters following a loss of coolant accident (LOCA), one of the most serious events in pressurized water reactors. The TFT integrates several critical components: a static covariate encoder, LSTM layers for local dynamics, and a multi-head attention mechanism to capture distant temporal relationships. It also uses quantile regression, which provides prediction intervals rather than point estimates. This feature directly addresses uncertainty in the forecast, an essential consideration for trustworthy AI in nuclear safety systems. Moreover, by including accident type and severity labels as static covariates, the model ensures that contextual knowledge shapes the prediction process. The architecture also includes gating mechanisms that reduce the influence of irrelevant or noisy data, adding a layer of robustness. The multi-horizon forecasting design allows reactor operators to view how the system may evolve over several future time steps, helping them take precautionary action. Another compelling study focuses on predicting disruptions in fusion reactors, particularly tokamaks \cite{spangher2023autoregressive}. These disruptions represent sudden losses of plasma stability and pose a serious threat to reactor integrity. In this case, the researchers adapt a GPT-like autoregressive transformer to a classification task, where the model determines the likelihood of disruption in the near future based on multivariate time-series data. The model benefits from being able to capture long sequences of control parameters, sensor data, and physical measurements, helping it capture early warning signs that simpler models might miss. Techniques such as curriculum learning and state pretraining are introduced to improve performance on harder cases, like disruptions that evolve quickly near the end of a plasma shot. Despite these strengths, the model lacks any formal method for uncertainty estimation or interpretability. Given the extremely high-stakes nature of fusion disruption prediction, the inability to quantify how confident the model is in its predictions, or explain which variables are driving its decisions, is a notable shortcoming.

In a different direction, transformer-based architecture is used to enhance Particle-in-Cell (PIC) \cite{chen2023leveraging} simulations used in plasma physics. These simulations are central to modeling electromagnetic interactions, often within the context of fusion or high-energy nuclear research. The proposed architecture uses attention mechanisms to replace the traditional particle-to-grid interpolation step, which is typically one of the computational bottlenecks in PIC codes. The approach begins with a transformer model trained to output first-order accurate charge densities based on particle positions and grid information. A second transformer then improves this to second-order accuracy, guided by a discriminator network. This hierarchical strategy significantly reduces the data labeling requirements while enhancing precision. However, the model does not explicitly include any features aimed at ensuring interpretability or robustness. There is no uncertainty quantification, no discussion of bias handling, and no visibility into the model’s reasoning. While technically impressive and potentially useful in high-performance computing settings, the model’s applicability in safety-critical environments may be limited by the absence of trust-enabling mechanisms.

In the realm of environmental safety, another team developed a transformer-based framework, NRFormer \cite{lyu2024nrformer}, to forecast nuclear radiation levels across an entire country. Using data from thousands of monitoring stations and corresponding meteorological information, the model aims to provide forecasts of radiation levels with a horizon of 1 to 24 days ahead. This is particularly important in regions vulnerable to radioactive leaks or atmospheric dispersion from nuclear power plants. The NRFormer architecture is specially designed to address the challenges posed by uneven station distribution and the non-stationary nature of radiation levels. It includes a non-stationary temporal attention module, an imbalance-aware spatial attention mechanism, and a radiation propagation prompting strategy. These adaptations allow the model to generalize across geographies and seasons, and also mitigate overfitting to heavily monitored areas. While not as formal as probabilistic methods, the model’s use of attention weights and normalization techniques does introduce a degree of interpretability and robustness. Furthermore, the prompting strategy helps the model incorporate domain knowledge in a structured way, which enhances its trustworthiness, especially in data-scarce or unseen scenarios.

\begin{tcolorbox}[colback=orange!5!white,colframe=red!75!black]
\noindent Taken together, these works show the growing influence of transformers in nuclear-related applications, particularly where time-series or spatial-temporal data are central. However, only a few of them incorporate design choices aimed at enhancing trustworthiness. The study on nuclear reactor accident prognosis stands out for its use of quantile regression, static covariates, and uncertainty intervals, features that are explicitly designed to support decision-making in safety-critical environments. Similarly, the radiation forecasting model, while less formal in its treatment of uncertainty, does attempt to handle distributional imbalance and contextual variation in a way that improves robustness and transparency. In contrast, the PIC simulation and fusion disruption prediction models, focus largely on accuracy and speed without addressing the needs for interpretability, robustness, or fairness. This is a significant limitation. In high-risk domains like nuclear engineering, models that fail to provide confidence estimates or rationale for their predictions risk becoming opaque tools that decision-makers cannot fully rely on. 
\end{tcolorbox}

\subsection{Theorem Proving}
\begin{figure}[htpb]
    \centering
    \includegraphics[width=0.9\textwidth]{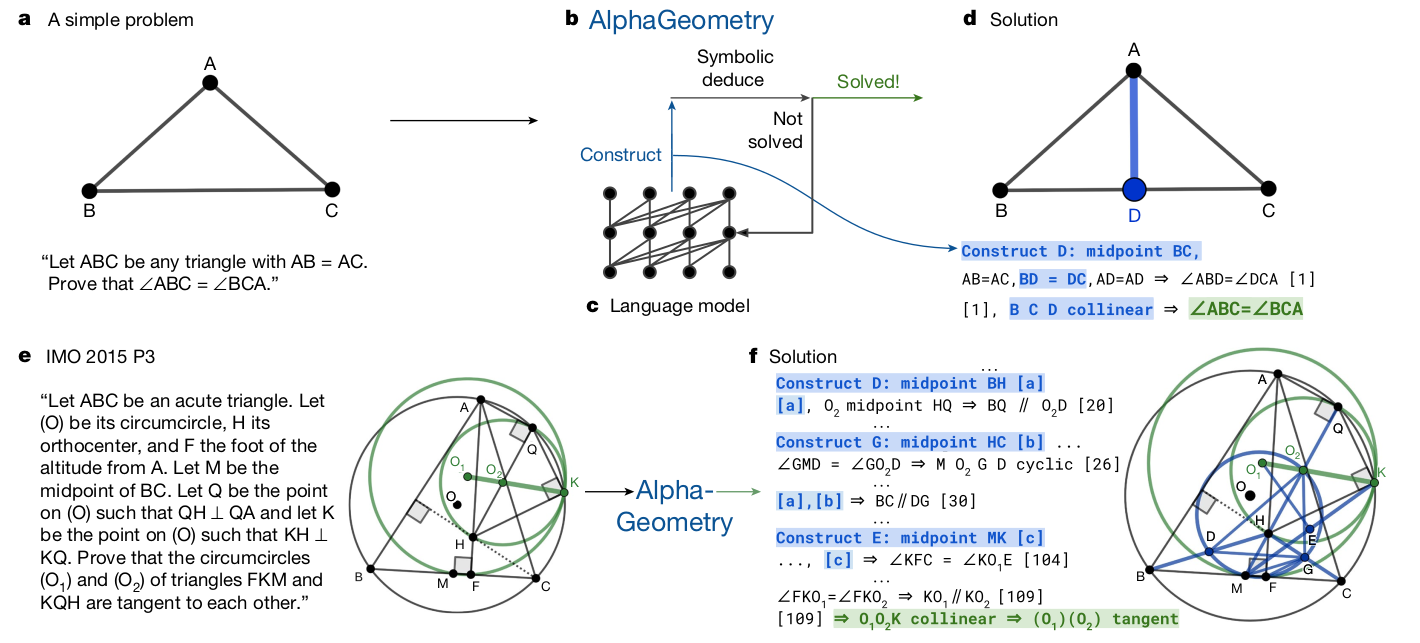}
    \caption{AlphaGeometry's neuro-symbolic loop. (a) A simple geometry problem example. (b) Proof search begins with the symbolic deduction engine. (c) The language model suggests an auxiliary point (blue) when the engine stalls. (d) The engine uses the new point to solve the simple problem. (e) Problem statement for the complex IMO 2015 Problem 3. (f) Solution to IMO 2015 P3, interleaving model suggestions (blue) and symbolic deductions. (Source: Trinh et al.\cite{trinh2024solving}).}
    \label{fig:ag1}
\end{figure}

The application of Transformer architectures has expanded beyond natural language processing into the domain of rigorous mathematical reasoning. While traditional symbolic engines good at logical verification, they suffer from combinatorial explosions when searching for proofs. Recent research demonstrates that Transformers can overcome this limitation by providing \textit{intuitive} guidance, predicting the necessary intermediate steps (such as auxiliary constructions or proof tactics) to guide symbolic search algorithms through infinite search spaces. This is best exemplified by the development of AlphaGeometry and its successors, which combine the pattern-recognition capabilities of Transformers with the precision of symbolic engines. The fundamental innovation lies in a \textit{neuro-symbolic} architecture: the Transformer acts as a \textit{creative guide}, predicting useful auxiliary constructions (like adding a line or point) that infinite search spaces often miss, while a deterministic symbolic engine acts as a rigorous verifier, deducing the logical consequences of those steps. This architecture was first successfully realized in AlphaGeometry (AG1) \cite{trinh2024solving} (overview of AG1 is shown in Figure \ref{fig:ag1}). Addressing the scarcity of human geometric proofs, the authors synthesized 100 million theorems by working backward from random diagrams, creating a massive dataset to train a Transformer from scratch. When presented with a problem, the Transformer suggests an auxiliary construction, which the symbolic engine (Deductive Database for Automated Reasoning) then uses to deduce new facts. If the engine stalls, the loop repeats. This synergy allowed AG1 to solve 25 out of 30 Olympiad-level geometry problems. Building on this foundation, AlphaGeometry2 (AG2) \cite{chervonyi2025gold} achieved an even higher standard, effectively surpassing the average International Mathematical Olympiad (IMO) Gold Medalist. AG2 integrates a Gemini-based language model trained on 300 million synthetic examples and pairs it with a symbolic engine optimized to be two orders of magnitude faster. A key advancement in AG2 is the \textit{Shared Knowledge Ensemble of Search Trees}, a search algorithm that allows multiple reasoning threads to run in parallel and share discovered facts to solving complex \textit{locus} problems involving moving objects and linear equations. Expanding beyond geometry, AlphaProof \cite{hubert2025olympiad} applies similar principles to the broader and more abstract domains of algebra, number theory, and combinatorics within the formal environment of the Lean theorem prover. Unlike the synthetic generation of AG1, AlphaProof employs an AlphaZero-style reinforcement learning loop. It bridges the data gap by \textit{auto-formalizing} millions of natural language math problems into formal Lean code, creating a vast training curriculum. During competition, it utilizes \textit{Test-Time Reinforcement Learning}, where the model actively trains on variants of the specific problem at hand to adapt its strategies in real-time. Together, these systems achieved a historic milestone at the 2024 IMO, solving four out of six problems to reach a Silver Medal level performance, demonstrating that Transformers can indeed master the formal mathematical proof. 

\begin{tcolorbox}[colback=orange!5!white,colframe=red!75!black]
\noindent Theorem proving stands as a unique outlier in the landscape of Transformer applications, offering a standard for moving from probabilistic guessing to verified truth. Unlike fields such as NLP or computer vision, where mistakes may look reasonable and go unnoticed, neuro-symbolic systems like AlphaGeometry address the \textit{hallucination problem} in a fundamental way. In these systems, the Transformer is not treated as a final decision-maker but as a proposal generator. Every idea it produces is checked by a strict symbolic engine, making it impossible for the system to output an incorrect proof. This design creates a strong feedback loop that is missing in standard predictive models. Since errors are detected immediately and objectively, the model can quickly adjust its strategy and try alternative solution paths. The success of this approach provides an important lesson for safety-critical AI. It suggests that trustworthiness may not come from making neural networks perfect, but from combining their creative abilities with domain-specific verification tools, such as physics simulators, code compilers, or formal logic checkers, to ensure reliability by construction.
\end{tcolorbox}

\subsection{Agentic AI}
Agentic AI refers to intelligent systems composed of autonomous, interactive agents that coordinate, communicate, and dynamically adapt to achieve complex goals. Unlike standalone AI agents, which perform narrowly defined tasks, Agentic AI systems exhibit persistent memory, goal decomposition, inter-agent orchestration, and adaptive decision-making. Transformers, especially large language or image models, form the computational backbone of these systems, driving their reasoning, planning, and perception capabilities (Sapkota et al. \cite{sapkota2025ai}). Transformers enable context-aware reasoning and long-horizon planning across diverse modalities. In Agentic AI, transformers often serve dual roles: as reasoning engines that interpret goals and decompose them into subtasks, and as communication engines that manage inter-agent dialogue. Models such as GPT-4 and PaLM are embedded in frameworks like AutoGPT, CrewAI, and LangGraph, where they coordinate complex operations ranging from research automation to robotic control (Sapkota et al. \cite{sapkota2025ai}). These models are orchestrated via meta-agents or distributed planners that implement dynamic workflows using tools like ReAct prompting, Retrieval-Augmented Generation (RAG), where the model augments its internal knowledge by querying an external data source such as a document corpus or database at each step. In multi-agent ecosystems, transformers are enhanced with persistent memory and dynamic task allocation strategies. These features are important in safety-critical applications such as medical diagnostics, autonomous robotics, and disaster response, where situational awareness and real-time coordination are essential. 

\begin{figure}[htpb]
    \centering
    \includegraphics[width=0.75\textwidth]{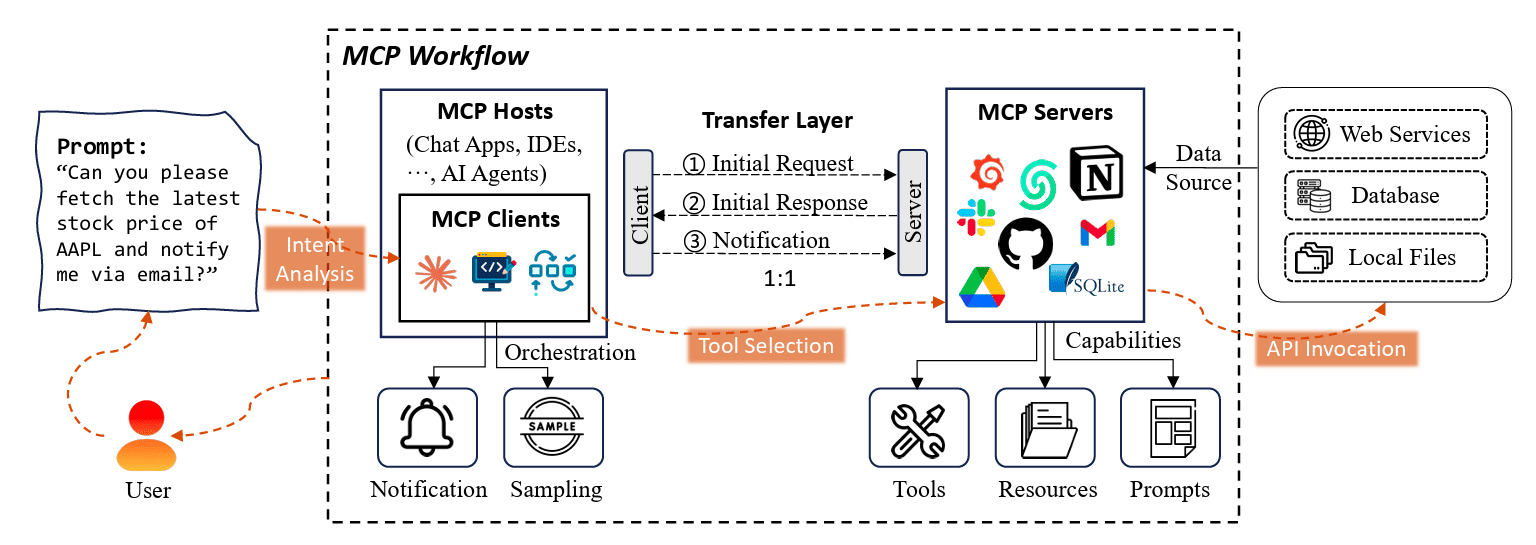}
    \caption{Illustration of the MCP workflow. The user request is analyzed by the MCP client, which selects suitable tools hosted on MCP servers. These servers manage capabilities such as API access, resources, and prompts, facilitating seamless execution and notification. Adapted from Hou et al. \cite{hou2025model}}
    \label{fig:mcp}
\end{figure}

The Model Context Protocol (MCP) provides a standardized infrastructure to enable Agentic AI systems to discover, invoke, and coordinate external tools autonomously (Hou et al. \cite{hou2025model}). MCP facilitates interoperability by abstracting tool-specific APIs into modular, callable units that agents can uses without integration. This decouples model logic from execution details, increasing scalability and adaptability. An MCP server consists three core services: tools, resources, and prompts, allowing AI agents to interact with real-time data, external APIs, or predefined workflows. For example, a financial agent could invoke an MCP-exposed pricing engine, process updates, and coordinate follow-up actions with minimal human oversight. MCP clients manage communication, ensuring secure orchestration, while the transport layer guarantees real-time, reliable data exchange. Figure \ref{fig:mcp} illustrates the MCP workflow: the user issues a natural language request (e.g., '\textit{fetch AAPL stock price and notify via email}'), which is processed by the MCP host (a chat app or agent framework). The MCP client performs intent analysis and routes the task via the transfer layer to the appropriate MCP server, which handles tool execution, data sourcing, and real-time notifications. MCP's security-aware lifecycle, spanning creation, operation, and update further ensures safe deployment in critical environments. During creation, integrity checks prevent backdoors or spoofed servers; During updates, version drift and unauthorized reconfiguration are mitigated via controlled authorization and auditing (Hou et al. \cite{hou2025model}). 

For Agentic AI systems to be viable in safety-critical environments, trustworthiness must be embedded at every level from data pipelines and model behavior to communication protocols and execution logic. Trustworthiness is not a singular property but a collection of interdependent dimensions that together ensure safety, fairness, and robustness.

\begin{itemize}
    \item \textbf{Interpretability:} Attention maps from transformers provide partial transparency, but real-world safety demands deeper interpretability through causal attribution, natural language explanations, and traceable decision pathways.

    \item \textbf{Robustness:} Agentic AI systems must remain functional under uncertainty, noisy inputs, and even adversarial conditions. Approaches like redundant planning agents and adaptive memory modules enhance system resistivity in dynamic or degraded environments.

    \item \textbf{Fairness \& Bias:} Pre-trained transformer models often carry forward societal biases found in their training data. This becomes particularly problematic in domains such as healthcare, law, finance, and human resources, where biased decisions can lead to systemic discrimination. To mitigate harm, techniques like fairness-constrained fine-tuning and demographic auditing must be incorporated.

    \item \textbf{Privacy:} These systems often process sensitive and proprietary data. Mechanisms such as federated learning \cite{mcmahan2017communication}, differential privacy \cite{dwork2008differential}, and Homomorphic Encryption \cite{gentry2009fully} should be standard, especially when AI agents interact with external tools via orchestration protocols like MCP.
    \item \textbf{Security:} Protocols like the MCP must be robust against misuse. Ensuring secure communication channels, access permissions, and runtime monitoring is essential to prevent unauthorized tool access or data leaks.
\end{itemize}

\section{Summary and Discussion}
Transformer architectures have rapidly emerged as foundational models in machine learning, achieving state-of-the-art performance across a broad spectrum of domains, including natural language processing, computer vision, and, more recently, scientific computing. Their remarkable scalability and representational capacity have accelerated their deployment in high-stakes and safety-critical settings, such as healthcare diagnostics, autonomous navigation, climate modeling, materials discovery, and robotics. Yet, as their societal and scientific influence deepens, a fundamental question arises: \emph{Can Transformer models be considered trustworthy for safety-critical and scientific applications?} Despite impressive empirical successes, systematic evaluation of their reliability, transparency, and broader impacts remains incomplete. In this Perspective, we provide a comprehensive and critical assessment of Transformer-based models through the lens of six interrelated dimensions: interpretability, robustness, fairness, privacy, uncertainty quantification, and scientific applicability. By synthesizing recent advances and identifying persistent limitations, we aim to clarify the conditions under which Transformer models can be responsibly deployed; and to delineate the open challenges that must be addressed before such trust can be fully justified.

The work begins by examining interpretability and explainability, which serve as foundational prerequisites for building user trust in any artificial intelligence systems. Attention mechanisms, central to Transformer architectures, provide insights into model reasoning by enabling the visualization of relevance patterns between input elements. While attention heatmaps offer a degree of transparency, their role as faithful explanations remains under active scrutiny. In the visual domain, the emergence of Vision Transformers has enabled global spatial reasoning through self-attention mechanisms. Techniques such as attention rollout and gradient-based saliency maps are discussed as tools for enhancing interpretability in ViT-based pipelines. Morover, this work reviews symbolic and neuro-symbolic Transformers, which integrate formal reasoning and logical priors into Transformer models to improve their transparency and alignment with domain-specific rules. The discussion then shifts to robustness, an important attribute for model reliability in dynamic or adversarial environments. Despite their strong performance on benchmark datasets, Transformer models are demonstrably vulnerable to adversarial attacks, which are carefully crafted perturbations capable of manipulating model predictions. This work details such vulnerabilities in both language and vision modalities and explores existing defenses, including adversarial training, robust optimization techniques, and randomized smoothing. In data-constrained scientific applications, ensuring robustness to noise and out-of-distribution inputs is particularly important. Fairness and bias mitigation constitute the third pillar of trustworthiness. Large pre-trained Transformers, often trained on web-scale corpora, have been shown to perpetuate and sometimes amplify societal biases, including those related to gender, race, and culture. This work synthesizes recent findings in this area and reviews multiple mitigation strategies, such as counterfactual data augmentation, adversarial debiasing, and fairness-aware loss functions. In parallel, the issue of privacy is examined, with emphasis on the risks posed by model memorization and data leakage. Techniques such as differential privacy, federated learning, and knowledge distillation are presented as viable solutions for privacy-preserving training and inference. 

The perspective also highlights emerging directions in scientific machine learning, where the integration of Transformer architectures offers novel opportunities for modeling complex physical systems. In particular, physics-informed Transformers are discussed as promising tools that incorporate domain knowledge and governing equations directly into the learning process. These models are capable of solving forward and inverse problems involving partial differential equations, enabling both data-efficient learning and adherence to physical laws. Furthermore, we examine techniques for uncertainty quantification, including Monte Carlo dropout, Bayesian attention formulations, and ensemble methods, which are essential for model validation in scientific and engineering contexts. To support the systematic evaluation of trustworthy Transformers, this work reviews a range of metrics and benchmarking protocols across interpretability, robustness, fairness, and uncertainty. The authors advocate for the development of domain-specific and standardized evaluation criteria to ensure consistent and reproducible assessments, particularly in safety-critical workflows. To this end, the perspective highlights the applicability of Transformers in a broad range of mission-critical domains, providing in-depth discussions on their use in autonomous systems, medical imaging, drug discovery, robotics, genomics, climate science, materials informatics, and fluid dynamics. In each of these settings, ensuring transparency, robustness to perturbations, and calibrated uncertainty is not merely beneficial but indispensable for safe and ethical deployment. The authors argue that advances in trustworthy Transformer architectures must be aligned with the requirements of these sensitive application domains.

In conclusion, this work presents a holistic and multidisciplinary assessment of ongoing efforts to improve the trustworthiness of Transformer models. As these architectures continue to evolve and expand into real-world systems with significant societal impact, there is a pressing need for the development of Transformer models that are not only accurate but also interpretable, fair, robust, and privacy-aware. This work concludes by outlining several future research directions, including the design of physics-aware attention mechanisms, the incorporation of ethical constraints during training, and the creation of trust-oriented benchmark suites for Transformer-based systems in scientific and industrial domains.

\subsection{Future directions}
The next phase of Transformer research must move beyond performance scaling toward principled trust engineering. Rather than optimizing interpretability, robustness, fairness, privacy, and uncertainty in isolation, future systems should adopt unified multi-objective frameworks that explicitly encode trade-offs among these dimensions. Embedding causal reasoning and counterfactual inference into Transformer architectures represents an important step toward improving robustness under distributional shift. Complementarily, formal verification and certification methods, providing provable guarantees on safety, stability, or bounded behavior, are essential for deployment in safety-critical domains.

Efficiency must not compromise reliability. Trust-preserving compression strategies, including fairness-aware pruning and explainable distillation, should ensure that smaller models retain transparency and robustness. As real-world environments are dynamic, continual evaluation protocols with drift detection and recalibration mechanisms are necessary to sustain trust over time. Human-centered interfaces that enable users to interrogate model reasoning and uncertainty will further support accountable deployment. Expanding trust research into multimodal and multilingual settings is increasingly urgent, particularly as large-scale systems integrate text, vision, and audio modalities. 

In scientific and high-stakes applications, such as medicine, climate science, and aerospace, integrating domain knowledge, physics-informed priors, and rigorous uncertainty quantification will be indispensable. To this end, the field must establish standardized benchmarks and reproducible metrics for trustworthiness, while advancing hybrid neuro-symbolic approaches that combine formal reasoning with deep representation learning.


Ultimately, the challenge is not whether Transformer models can achieve extraordinary accuracy, but whether they can earn justified confidence under real-world constraints. As these systems increasingly shape scientific discovery, policy decisions, and human well-being, the defining question remains: \emph{Do we truly trust Transformers?}

\bibliographystyle{unsrt}
\bibliography{sample}

@article{vaswani2017attention,
  title={Attention is all you need},
  author={Vaswani, Ashish and Shazeer, Noam and Parmar, Niki and Uszkoreit, Jakob and Jones, Llion and Gomez, Aidan N and Kaiser, {\L}ukasz and Polosukhin, Illia},
  journal={Advances in neural information processing systems},
  volume={30},
  year={2017}
}

@article{jagtap2022physics,
  title={Physics-informed neural networks for inverse problems in supersonic flows},
  author={Jagtap, Ameya D and Mao, Zhiping and Adams, Nikolaus and Karniadakis, George Em},
  journal={Journal of Computational Physics},
  volume={466},
  pages={111402},
  year={2022},
  publisher={Elsevier}
}

@article{hu2021extended,
  title={When do extended physics-informed neural networks (XPINNs) improve generalization?},
  author = {Hu, Zheyuan and Jagtap, Ameya D. and Karniadakis, George Em and Kawaguchi, Kenji},
  journal = {SIAM Journal on Scientific Computing},
  volume = {44},
  number = {5},
  pages = {A3158-A3182},
  year = {2022}
}

@article{jagtap2020locally,
  title={Locally adaptive activation functions with slope recovery for deep and physics-informed neural networks},
  author={Jagtap, Ameya D and Kawaguchi, Kenji and Em Karniadakis, George},
  journal={Proceedings of the Royal Society A},
  volume={476},
  number={2239},
  pages={20200334},
  year={2020},
  publisher={The Royal Society}
}

@article{jagtap2023important,
  title={How important are activation functions in regression and classification? A survey, performance comparison, and future directions},
  author={Jagtap, Ameya D and Karniadakis, George Em},
  journal={Journal of Machine Learning for Modeling and Computing},
  volume={4},
  number={1},
  year={2023},
  publisher={Begel House Inc.}
}

@article{jagtap2020extended,
  title={Extended physics-informed neural networks (XPINNs): A generalized space-time domain decomposition based deep learning framework for nonlinear partial differential equations},
  author={Jagtap, Ameya D and Karniadakis, George Em},
  journal={Communications in Computational Physics},
  volume={28},
  number={5},
  year={2020},
  publisher={Brown Univ., Providence, RI (United States)}
}

@article{raissi2019physics,
  title={Physics-informed neural networks: A deep learning framework for solving forward and inverse problems involving nonlinear partial differential equations},
  author={Raissi, Maziar and Perdikaris, Paris and Karniadakis, George E},
  journal={Journal of Computational physics},
  volume={378},
  pages={686--707},
  year={2019},
  publisher={Elsevier}
}

@article{menon2026scientific,
  title={On Scientific Foundation Models: Rigorous Definitions, Key Applications, and a Comprehensive Survey},
  author={Menon, Sidharth S and Mondal, Trishit and Brahmachary, Shuvayan and Panda, Aniruddha and Joshi, Subodh M and Kalyanaraman, Kaushic and Jagtap, Ameya D},
  journal={Neural Networks},
  volume={198},
  pages={108567},
  year={2026},
  publisher={Elsevier}
}

@article{menon2025anant,
  title={Anant-Net: Breaking the Curse of Dimensionality with Scalable and Interpretable Neural Surrogate for High-Dimensional PDEs},
  author={Menon, Sidharth S and Jagtap, Ameya D},
  journal = {Computer Methods in Applied Mechanics and Engineering},
  volume = {447},
  pages = {118403},
  year = {2025},
  publisher={Elsevier}
}

@article{abbasi2025challenges,
  title = {Challenges and advancements in modeling shock fronts with physics-informed neural networks: A review and benchmarking study},
   author = {Jassem Abbasi and Ameya D. Jagtap and Ben Moseley and Aksel Hiorth and P{\aa}l {\O}steb{\o}},
  journal = {Neurocomputing},
  volume = {657},
  pages = {131440},
  year = {2025},
  publisher={Elsevier}
}

@article{jagtap2020adaptive,
  title={Adaptive activation functions accelerate convergence in deep and physics-informed neural networks},
  author={Jagtap, Ameya D and Kawaguchi, Kenji and Karniadakis, George Em},
  journal={Journal of Computational Physics},
  volume={404},
  pages={109136},
  year={2020},
  publisher={Elsevier}
}

@article{jagtap2022deepknn,
  title={Deep Kronecker neural networks: A general framework for neural networks with adaptive activation functions},
  author={Jagtap, Ameya D and Shin, Yeonjong and Kawaguchi, Kenji and Karniadakis, George Em},
  journal={Neurocomputing},
  volume={468},
  pages={165--180},
  year={2022},
  publisher={Elsevier}
}

@article{abbasi2025history,
  title={History-Matching of imbibition flow in fractured porous media Using Physics-Informed Neural Networks (PINNs)},
  author={Abbasi, Jassem and Moseley, Ben and Kurotori, Takeshi and Jagtap, Ameya D and Kovscek, Anthony R and Hiorth, Aksel and Andersen, P{\aa}l {\O}steb{\o}},
  journal={Computer Methods in Applied Mechanics and Engineering},
  volume={437},
  pages={117784},
  year={2025},
  publisher={Elsevier}
}

@article{jagtap2022deepWW,
  title={Deep learning of inverse water waves problems using multi-fidelity data: Application to Serre--Green--Naghdi equations},
  author={Jagtap, Ameya D and Mitsotakis, Dimitrios and Karniadakis, George Em},
  journal={Ocean Engineering},
  volume={248},
  pages={110775},
  year={2022},
  publisher={Elsevier}
}

@article{peyvan2024riemannonets,
  title={Riemannonets: Interpretable neural operators for riemann problems},
  author={Peyvan, Ahmad and Oommen, Vivek and Jagtap, Ameya D and Karniadakis, George Em},
  journal={Computer Methods in Applied Mechanics and Engineering},
  volume={426},
  pages={116996},
  year={2024},
  publisher={Elsevier}
}

@article{Zhang2025BubbleOKAN,
  title = {BubbleOKAN: A physics-informed interpretable neural operator for high-frequency bubble dynamics},
  author = {Yunhao Zhang and Sidharth S. Menon and Lin Cheng and Aswin Gnanaskandan and Ameya D. Jagtap},
  journal = {Computer Methods in Applied Mechanics and Engineering},
  volume = {450},
  pages = {118667},
  year = {2026},
  publisher={Elsevier}
}

@article{goswami2024learning,
  title={Learning stiff chemical kinetics using extended deep neural operators},
  author={Goswami, Somdatta and Jagtap, Ameya D and Babaee, Hessam and Susi, Bryan T and Karniadakis, George Em},
  journal={Computer Methods in Applied Mechanics and Engineering},
  volume={419},
  pages={116674},
  year={2024},
  publisher={Elsevier}
}

@article{vig2019multiscale,
  title={A multiscale visualization of attention in the transformer model},
  author={Vig, Jesse},
  journal={arXiv preprint arXiv:1906.05714},
  year={2019}
}

@article{lu2021learning,
  title={Learning nonlinear operators via DeepONet based on the universal approximation theorem of operators},
  author={Lu, Lu and Jin, Pengzhan and Pang, Guofei and Zhang, Zhongqiang and Karniadakis, George Em},
  journal={Nature machine intelligence},
  volume={3},
  number={3},
  pages={218--229},
  year={2021},
  publisher={Nature Publishing Group UK London}
}

@article{clark2019does,
  title={What does bert look at? an analysis of bert's attention},
  author={Clark, Kevin and Khandelwal, Urvashi and Levy, Omer and Manning, Christopher D},
  journal={arXiv preprint arXiv:1906.04341},
  year={2019}
}

@inproceedings{chefer2021transformer,
  title={Transformer interpretability beyond attention visualization},
  author={Chefer, Hila and Gur, Shir and Wolf, Lior},
  booktitle={Proceedings of the IEEE/CVF conference on computer vision and pattern recognition},
  pages={782--791},
  year={2021}
}

@article{abnar2020quantifying,
  title={Quantifying attention flow in transformers},
  author={Abnar, Samira and Zuidema, Willem},
  journal={arXiv preprint arXiv:2005.00928},
  year={2020}
}

@article{voita2019bottom,
  title={The bottom-up evolution of representations in the transformer: A study with machine translation and language modeling objectives},
  author={Voita, Elena and Sennrich, Rico and Titov, Ivan},
  journal={arXiv preprint arXiv:1909.01380},
  year={2019}
}

@article{lorsung2024physics,
  title={Physics informed token transformer for solving partial differential equations},
  author={Lorsung, Cooper and Li, Zijie and Farimani, Amir Barati},
  journal={Machine Learning: Science and Technology},
  volume={5},
  number={1},
  pages={015032},
  year={2024},
  publisher={IOP Publishing}
}

@inproceedings{hao2023gnot,
  title={Gnot: A general neural operator transformer for operator learning},
  author={Hao, Zhongkai and Wang, Zhengyi and Su, Hang and Ying, Chengyang and Dong, Yinpeng and Liu, Songming and Cheng, Ze and Song, Jian and Zhu, Jun},
  booktitle={International Conference on Machine Learning},
  pages={12556--12569},
  year={2023},
  organization={PMLR}
}

@article{guo2022transformer,
  title={Transformer meets boundary value inverse problems},
  author={Guo, Ruchi and Cao, Shuhao and Chen, Long},
  journal={arXiv preprint arXiv:2209.14977},
  year={2022}
}

@inproceedings{mahmood2021robustness,
  title={On the robustness of vision transformers to adversarial examples},
  author={Mahmood, Kaleel and Mahmood, Rigel and Van Dijk, Marten},
  booktitle={Proceedings of the IEEE/CVF international conference on computer vision},
  pages={7838--7847},
  year={2021}
}

@article{mandal2023biased,
  title={Biased Attention: Do Vision Transformers Amplify Gender Bias More than Convolutional Neural Networks?},
  author={Mandal, Abhishek and Leavy, Susan and Little, Suzanne},
  journal={arXiv preprint arXiv:2309.08760},
  year={2023}
}

@article{nemani2023gender,
  title={Gender bias in transformer models: A comprehensive survey},
  author={Nemani, Praneeth and Joel, Yericherla Deepak and Vijay, Palla and Liza, Farhana Ferdousi},
  journal={arXiv preprint arXiv:2306.10530},
  year={2023}
}

@inproceedings{zhang2024towards,
  title={Towards fairness-aware adversarial learning},
  author={Zhang, Yanghao and Zhang, Tianle and Mu, Ronghui and Huang, Xiaowei and Ruan, Wenjie},
  booktitle={Proceedings of the IEEE/CVF Conference on Computer Vision and Pattern Recognition},
  pages={24746--24755},
  year={2024}
}

@Article{computers13040092,
AUTHOR = {Fantozzi, Paolo and Naldi, Maurizio},
TITLE = {The Explainability of Transformers: Current Status and Directions},
JOURNAL = {Computers},
VOLUME = {13},
YEAR = {2024},
NUMBER = {4},
ARTICLE-NUMBER = {92},
URL = {https://www.mdpi.com/2073-431X/13/4/92},
ISSN = {2073-431X},
DOI = {10.3390/computers13040092}
}

@article{rai2024practical,
  title={A practical review of mechanistic interpretability for transformer-based language models},
  author={Rai, Daking and Zhou, Yilun and Feng, Shi and Saparov, Abulhair and Yao, Ziyu},
  journal={arXiv preprint arXiv:2407.02646},
  year={2024}
}

@article{geneva2022transformers,
  title={Transformers for modeling physical systems},
  author={Geneva, Nicholas and Zabaras, Nicholas},
  journal={Neural Networks},
  volume={146},
  pages={272--289},
  year={2022},
  publisher={Elsevier}
}

@article{sheth2023neurosymbolic,
  title={Neurosymbolic AI--Why, What, and How},
  author={Sheth, Amit and Roy, Kaushik and Gaur, Manas},
  journal={arXiv preprint arXiv:2305.00813},
  year={2023}
}

@article{zhang2024neuro,
  title={Neuro-Symbolic AI: Explainability, Challenges, and Future Trends},
  author={Zhang, Xin and Sheng, Victor S},
  journal={arXiv preprint arXiv:2411.04383},
  year={2024}
}

@article{tan2024transformation,
  title={Transformation-Dependent Adversarial Attacks},
  author={Tan, Yaoteng and Cai, Zikui and Asif, M Salman},
  journal={arXiv preprint arXiv:2406.08443},
  year={2024}
}

@article{shi2020robustness,
  title={Robustness verification for transformers},
  author={Shi, Zhouxing and Zhang, Huan and Chang, Kai-Wei and Huang, Minlie and Hsieh, Cho-Jui},
  journal={arXiv preprint arXiv:2002.06622},
  year={2020}
}

@inproceedings{kim2024exploring,
  title={Exploring adversarial robustness of vision transformers in the spectral perspective},
  author={Kim, Gihyun and Kim, Juyeop and Lee, Jong-Seok},
  booktitle={Proceedings of the IEEE/CVF Winter Conference on Applications of Computer Vision},
  pages={3976--3985},
  year={2024}
}

@article{gesi2024beyond,
  title={Beyond Self-learned Attention: Mitigating Attention Bias in Transformer-based Models Using Attention Guidance},
  author={Gesi, Jiri and Ahmed, Iftekhar},
  journal={arXiv preprint arXiv:2402.16790},
  year={2024}
}

@article{castellon2023dp,
  title={DP-TBART: A Transformer-based Autoregressive Model for Differentially Private Tabular Data Generation},
  author={Castellon, Rodrigo and Gopal, Achintya and Bloniarz, Brian and Rosenberg, David},
  journal={arXiv preprint arXiv:2307.10430},
  year={2023}
}

@article{zhao2023pinnsformer,
  title={Pinnsformer: A transformer-based framework for physics-informed neural networks},
  author={Zhao, Zhiyuan and Ding, Xueying and Prakash, B Aditya},
  journal={arXiv preprint arXiv:2307.11833},
  year={2023}
}

@article{ovadia2024vito,
  title={Vito: Vision transformer-operator},
  author={Ovadia, Oded and Kahana, Adar and Stinis, Panos and Turkel, Eli and Givoli, Dan and Karniadakis, George Em},
  journal={Computer Methods in Applied Mechanics and Engineering},
  volume={428},
  pages={117109},
  year={2024},
  publisher={Elsevier}
}

@article{zhou2024unisolver,
  title={Unisolver: PDE-conditional transformers are universal PDE solvers},
  author={Zhou, Hang and Ma, Yuezhou and Wu, Haixu and Wang, Haowen and Long, Mingsheng},
  journal={arXiv preprint arXiv:2405.17527},
  year={2024}
}

@article{shao2021adversarial,
  title={On the adversarial robustness of vision transformers},
  author={Shao, Rulin and Shi, Zhouxing and Yi, Jinfeng and Chen, Pin-Yu and Hsieh, Cho-Jui},
  journal={arXiv preprint arXiv:2103.15670},
  year={2021}
}

@INPROCEEDINGS{10658195,
  author={Jain, Samyak and Dutta, Tanima},
  booktitle={2024 IEEE/CVF Conference on Computer Vision and Pattern Recognition (CVPR)}, 
  title={Towards Understanding and Improving Adversarial Robustness of Vision Transformers}, 
  year={2024},
  volume={},
  number={},
  pages={24736-24745},
  doi={10.1109/CVPR52733.2024.02336}}

@article{jain2019attention,
  title={Attention is not explanation},
  author={Jain, Sarthak and Wallace, Byron C},
  journal={arXiv preprint arXiv:1902.10186},
  year={2019}
}

@article{qiang2023interpretability,
  title={Interpretability-aware vision transformer},
  author={Qiang, Yao and Li, Chengyin and Khanduri, Prashant and Zhu, Dongxiao},
  journal={arXiv preprint arXiv:2309.08035},
  year={2023}
}

@article{jia2017adversarial,
  title={Adversarial examples for evaluating reading comprehension systems},
  author={Jia, Robin and Liang, Percy},
  journal={arXiv preprint arXiv:1707.07328},
  year={2017}
}

@article{hendrycks2020pretrained,
  title={Pretrained transformers improve out-of-distribution robustness},
  author={Hendrycks, Dan and Liu, Xiaoyuan and Wallace, Eric and Dziedzic, Adam and Krishnan, Rishabh and Song, Dawn},
  journal={arXiv preprint arXiv:2004.06100},
  year={2020}
}

@inproceedings{zhou2022understanding,
  title={Understanding the robustness in vision transformers},
  author={Zhou, Daquan and Yu, Zhiding and Xie, Enze and Xiao, Chaowei and Anandkumar, Animashree and Feng, Jiashi and Alvarez, Jose M},
  booktitle={International conference on machine learning},
  pages={27378--27394},
  year={2022},
  organization={PMLR}
}

@article{bai2021transformers,
  title={Are transformers more robust than cnns?},
  author={Bai, Yutong and Mei, Jieru and Yuille, Alan L and Xie, Cihang},
  journal={Advances in neural information processing systems},
  volume={34},
  pages={26831--26843},
  year={2021}
}

@inproceedings{paul2022vision,
  title={Vision transformers are robust learners},
  author={Paul, Sayak and Chen, Pin-Yu},
  booktitle={Proceedings of the AAAI conference on Artificial Intelligence},
  volume={36},
  number={2},
  pages={2071--2081},
  year={2022}
}

@article{anil2021large,
  title={Large-scale differentially private BERT},
  author={Anil, Rohan and Ghazi, Badih and Gupta, Vineet and Kumar, Ravi and Manurangsi, Pasin},
  journal={arXiv preprint arXiv:2108.01624},
  year={2021}
}

@article{chen2022x,
  title={The-x: Privacy-preserving transformer inference with homomorphic encryption},
  author={Chen, Tianyu and Bao, Hangbo and Huang, Shaohan and Dong, Li and Jiao, Binxing and Jiang, Daxin and Zhou, Haoyi and Li, Jianxin and Wei, Furu},
  journal={arXiv preprint arXiv:2206.00216},
  year={2022}
}

@article{ribeiro2020beyond,
  title={Beyond accuracy: Behavioral testing of NLP models with CheckList},
  author={Ribeiro, Marco Tulio and Wu, Tongshuang and Guestrin, Carlos and Singh, Sameer},
  journal={arXiv preprint arXiv:2005.04118},
  year={2020}
}

@article{goel2021robustness,
  title={Robustness gym: Unifying the NLP evaluation landscape},
  author={Goel, Karan and Rajani, Nazneen and Vig, Jesse and Tan, Samson and Wu, Jason and Zheng, Stephan and Xiong, Caiming and Bansal, Mohit and R{\'e}, Christopher},
  journal={arXiv preprint arXiv:2101.04840},
  year={2021}
}

@article{clark2018think,
  title={Think you have solved question answering? try arc, the ai2 reasoning challenge},
  author={Clark, Peter and Cowhey, Isaac and Etzioni, Oren and Khot, Tushar and Sabharwal, Ashish and Schoenick, Carissa and Tafjord, Oyvind},
  journal={arXiv preprint arXiv:1803.05457},
  year={2018}
}

@article{wallace2019universal,
  title={Universal adversarial triggers for attacking and analyzing NLP},
  author={Wallace, Eric and Feng, Shi and Kandpal, Nikhil and Gardner, Matt and Singh, Sameer},
  journal={arXiv preprint arXiv:1908.07125},
  year={2019}
}

@inproceedings{harikumar2021scalable,
  title={Scalable backdoor detection in neural networks},
  author={Harikumar, Haripriya and Le, Vuong and Rana, Santu and Bhattacharya, Sourangshu and Gupta, Sunil and Venkatesh, Svetha},
  booktitle={Machine Learning and Knowledge Discovery in Databases: European Conference, ECML PKDD 2020, Ghent, Belgium, September 14--18, 2020, Proceedings, Part II},
  pages={289--304},
  year={2021},
  organization={Springer}
}

@article{schaeffer2023emergent,
  title={Are emergent abilities of large language models a mirage?},
  author={Schaeffer, Rylan and Miranda, Brando and Koyejo, Sanmi},
  journal={Advances in Neural Information Processing Systems},
  volume={36},
  pages={55565--55581},
  year={2023}
}

@inproceedings{cheng2025exploring,
  title={Exploring the robustness of in-context learning with noisy labels},
  author={Cheng, Chen and Yu, Xinzhi and Wen, Haodong and Sun, Jingsong and Yue, Guanzhang and Zhang, Yihao and Wei, Zeming},
  booktitle={ICASSP 2025-2025 IEEE International Conference on Acoustics, Speech and Signal Processing (ICASSP)},
  pages={1--5},
  year={2025},
  organization={IEEE}
}

@article{inala2020neurosymbolic,
  title={Neurosymbolic transformers for multi-agent communication},
  author={Inala, Jeevana Priya and Yang, Yichen and Paulos, James and Pu, Yewen and Bastani, Osbert and Kumar, Vijay and Rinard, Martin and Solar-Lezama, Armando},
  journal={Advances in Neural Information Processing Systems},
  volume={33},
  pages={13597--13608},
  year={2020}
}

@INPROCEEDINGS{10031186,
  author={Baran, Joanna and Kocoń, Jan},
  booktitle={2022 IEEE International Conference on Data Mining Workshops (ICDMW)}, 
  title={Linguistic Knowledge Application to Neuro-Symbolic Transformers in Sentiment Analysis}, 
  year={2022},
  volume={},
  number={},
  pages={395-402},
  doi={10.1109/ICDMW58026.2022.00059}}

@article{wang2022behavior,
  title={Behavior cloned transformers are neurosymbolic reasoners},
  author={Wang, Ruoyao and Jansen, Peter and C{\^o}t{\'e}, Marc-Alexandre and Ammanabrolu, Prithviraj},
  journal={arXiv preprint arXiv:2210.07382},
  year={2022}
}

@INPROCEEDINGS{10651426,
  author={Russo, Alessandro Sebastian and Morra, Lia and Lamberti, Fabrizio and Dimasi, Paolo Emmanuel Ilario},
  booktitle={2024 International Joint Conference on Neural Networks (IJCNN)}, 
  title={ESRA: a Neuro-Symbolic Relation Transformer for Autonomous Driving}, 
  year={2024},
  volume={},
  number={},
  pages={1-10},
  doi={10.1109/IJCNN60899.2024.10651426}}

@article{hamilton2024neuro,
  title={Is neuro-symbolic AI meeting its promises in natural language processing? A structured review},
  author={Hamilton, Kyle and Nayak, Aparna and Bo{\v{z}}i{\'c}, Bojan and Longo, Luca},
  journal={Semantic Web},
  volume={15},
  number={4},
  pages={1265--1306},
  year={2024},
  publisher={IOS Press}
}

@article{baheri2025hierarchical,
  title={Hierarchical Neuro-Symbolic Decision Transformer},
  author={Baheri, Ali and Alm, Cecilia O},
  journal={arXiv preprint arXiv:2503.07148},
  year={2025}
}

@article{sankararaman2022bayesformer,
  title={Bayesformer: Transformer with uncertainty estimation},
  author={Sankararaman, Karthik Abinav and Wang, Sinong and Fang, Han},
  journal={arXiv preprint arXiv:2206.00826},
  year={2022}
}

@article{muller2021transformers,
  title={Transformers can do bayesian inference},
  author={M{\"u}ller, Samuel and Hollmann, Noah and Arango, Sebastian Pineda and Grabocka, Josif and Hutter, Frank},
  journal={arXiv preprint arXiv:2112.10510},
  year={2021}
}

@inproceedings{xue2021bayesian,
  title={Bayesian transformer language models for speech recognition},
  author={Xue, Boyang and Yu, Jianwei and Xu, Junhao and Liu, Shansong and Hu, Shoukang and Ye, Zi and Geng, Mengzhe and Liu, Xunying and Meng, Helen},
  booktitle={ICASSP 2021-2021 IEEE International Conference on Acoustics, Speech and Signal Processing (ICASSP)},
  pages={7378--7382},
  year={2021},
  organization={IEEE}
}

@article{xiao2024bayesian,
  title={Bayesian variational transformer: A generalizable model for rotating machinery fault diagnosis},
  author={Xiao, Yiming and Shao, Haidong and Wang, Jie and Yan, Shen and Liu, Bin},
  journal={Mechanical Systems and Signal Processing},
  volume={207},
  pages={110936},
  year={2024},
  publisher={Elsevier}
}

@article{bouallegue2024improving,
  title={Improving medium-range ensemble weather forecasts with hierarchical ensemble transformers},
  author={Bouall{\`e}gue, Zied Ben and Weyn, Jonathan A and Clare, Mariana CA and Dramsch, Jesper and Dueben, Peter and Chantry, Matthew},
  journal={Artificial Intelligence for the Earth Systems},
  volume={3},
  number={1},
  pages={e230027},
  year={2024},
  publisher={American Meteorological Society}
}

@article{hittawe2024time,
  title={Time-series weather prediction in the Red sea using ensemble transformers},
  author={Hittawe, Mohamad Mazen and Harrou, Fouzi and Togou, Mohammed Amine and Sun, Ying and Knio, Omar},
  journal={Applied Soft Computing},
  volume={164},
  pages={111926},
  year={2024},
  publisher={Elsevier}
}

@article{olorunnimbe2024ensemble,
  title={Ensemble of temporal Transformers for financial time series},
  author={Olorunnimbe, Kenniy and Viktor, Herna},
  journal={Journal of Intelligent Information Systems},
  volume={62},
  number={4},
  pages={1087--1111},
  year={2024},
  publisher={Springer}
}

@article{xu2023fine,
  title={Fine-grained visual classification via internal ensemble learning transformer},
  author={Xu, Qin and Wang, Jiahui and Jiang, Bo and Luo, Bin},
  journal={IEEE Transactions on Multimedia},
  volume={25},
  pages={9015--9028},
  year={2023},
  publisher={IEEE}
}

@article{liu2023dmeformer,
  title={Dmeformer: A newly designed dynamic model ensemble transformer for crude oil futures prediction},
  author={Liu, Chao and Ruan, Kaiyi and Ma, Xinmeng},
  journal={Heliyon},
  volume={9},
  number={6},
  year={2023},
  publisher={Elsevier}
}

@article{huang2024trustllm,
  title={Trustllm: Trustworthiness in large language models},
  author={Huang, Yue and Sun, Lichao and Wang, Haoran and Wu, Siyuan and Zhang, Qihui and Li, Yuan and Gao, Chujie and Huang, Yixin and Lyu, Wenhan and Zhang, Yixuan and others},
  journal={arXiv preprint arXiv:2401.05561},
  year={2024}
}

@article{chen2024trustworthy,
  title={Trustworthy, responsible, and safe ai: A comprehensive architectural framework for ai safety with challenges and mitigations},
  author={Chen, Chen and Gong, Xueluan and Liu, Ziyao and Jiang, Weifeng and Goh, Si Qi and Lam, Kwok-Yan},
  journal={arXiv preprint arXiv:2408.12935},
  year={2024}
}

@article{tadi2025trustformer,
  title={Trustformer: A Trusted Federated Transformer},
  author={Tadi, Ali Abbasi and Alhadidi, Dima and Rueda, Luis},
  journal={arXiv preprint arXiv:2501.11706},
  year={2025}
}

@article{latibari2024transformers,
  title={Transformers: A security perspective},
  author={Latibari, Banafsheh Saber and Nazari, Najmeh and Chowdhury, Muhtasim Alam and Gubbi, Kevin Immanuel and Fang, Chongzhou and Ghimire, Sujan and Hosseini, Elahe and Sayadi, Hossein and Homayoun, Houman and Salehi, Soheil and others},
  journal={IEEE Access},
  year={2024},
  publisher={IEEE}
}

@inproceedings{abadi2016deep,
  title={Deep learning with differential privacy},
  author={Abadi, Martin and Chu, Andy and Goodfellow, Ian and McMahan, H Brendan and Mironov, Ilya and Talwar, Kunal and Zhang, Li},
  booktitle={Proceedings of the 2016 ACM SIGSAC conference on computer and communications security},
  pages={308--318},
  year={2016}
}

@inproceedings{xu2022v2x,
  title={V2x-vit: Vehicle-to-everything cooperative perception with vision transformer},
  author={Xu, Runsheng and Xiang, Hao and Tu, Zhengzhong and Xia, Xin and Yang, Ming-Hsuan and Ma, Jiaqi},
  booktitle={European conference on computer vision},
  pages={107--124},
  year={2022},
  organization={Springer}
}

@article{li2203bevformer,
  title={Bevformer: Learning bird’s-eye-view representation from multi-camera images via spatiotemporal transformers. arxiv 2022},
  author={Li, Z and Wang, W and Li, H and Xie, E and Sima, C and Lu, T and Yu, Q and Dai, J},
  journal={arXiv preprint arXiv:2203.17270}
}

@article{jia2025drivetransformer,
  title={Drivetransformer: Unified transformer for scalable end-to-end autonomous driving},
  author={Jia, Xiaosong and You, Junqi and Zhang, Zhiyuan and Yan, Junchi},
  journal={arXiv preprint arXiv:2503.07656},
  year={2025}
}

@article{bjorck2025gr00t,
  title={Gr00t n1: An open foundation model for generalist humanoid robots},
  author={Bjorck, Johan and Casta{\~n}eda, Fernando and Cherniadev, Nikita and Da, Xingye and Ding, Runyu and Fan, Linxi and Fang, Yu and Fox, Dieter and Hu, Fengyuan and Huang, Spencer and others},
  journal={arXiv preprint arXiv:2503.14734},
  year={2025}
}

@article{ji2021dnabert,
  title={DNABERT: pre-trained Bidirectional Encoder Representations from Transformers model for DNA-language in genome},
  author={Ji, Yanrong and Zhou, Zhihan and Liu, Han and Davuluri, Ramana V},
  journal={Bioinformatics},
  volume={37},
  number={15},
  pages={2112--2120},
  year={2021},
  publisher={Oxford University Press}
}

@article{rives2021biological,
  title={Biological structure and function emerge from scaling unsupervised learning to 250 million protein sequences},
  author={Rives, Alexander and Meier, Joshua and Sercu, Tom and Goyal, Siddharth and Lin, Zeming and Liu, Jason and Guo, Demi and Ott, Myle and Zitnick, C Lawrence and Ma, Jerry and others},
  journal={Proceedings of the National Academy of Sciences},
  volume={118},
  number={15},
  pages={e2016239118},
  year={2021},
  publisher={National Academy of Sciences}
}

@article{jumper2021highly,
  title={Highly accurate protein structure prediction with AlphaFold},
  author={Jumper, John and Evans, Richard and Pritzel, Alexander and Green, Tim and Figurnov, Michael and Ronneberger, Olaf and Tunyasuvunakool, Kathryn and Bates, Russ and {\v{Z}}{\'\i}dek, Augustin and Potapenko, Anna and others},
  journal={nature},
  volume={596},
  number={7873},
  pages={583--589},
  year={2021},
  publisher={Nature Publishing Group UK London}
}

@article{maziarka2020molecule,
  title={Molecule attention transformer},
  author={Maziarka, {\L}ukasz and Danel, Tomasz and Mucha, S{\l}awomir and Rataj, Krzysztof and Tabor, Jacek and Jastrzebski, Stanis{\l}aw},
  journal={arXiv preprint arXiv:2002.08264},
  year={2020}
}

@article{zhang2022transformer,
  title={Transformer for gene expression modeling (T-GEM): an interpretable deep learning model for gene expression-based phenotype predictions},
  author={Zhang, Ting-He and Hasib, Md Musaddaqul and Chiu, Yu-Chiao and Han, Zhi-Feng and Jin, Yu-Fang and Flores, Mario and Chen, Yidong and Huang, Yufei},
  journal={Cancers},
  volume={14},
  number={19},
  pages={4763},
  year={2022},
  publisher={MDPI}
}

@article{gao2022earthformer,
  title={Earthformer: Exploring space-time transformers for earth system forecasting},
  author={Gao, Zhihan and Shi, Xingjian and Wang, Hao and Zhu, Yi and Wang, Yuyang Bernie and Li, Mu and Yeung, Dit-Yan},
  journal={Advances in Neural Information Processing Systems},
  volume={35},
  pages={25390--25403},
  year={2022}
}

@article{wu2024landslide,
  title={Landslide mapping based on a hybrid CNN-transformer network and deep transfer learning using remote sensing images with topographic and spectral features},
  author={Wu, Lei and Liu, Rui and Ju, Nengpan and Zhang, Ao and Gou, Jingsong and He, Guolei and Lei, Yuzhu},
  journal={International Journal of Applied Earth Observation and Geoinformation},
  volume={126},
  pages={103612},
  year={2024},
  publisher={Elsevier}
}

@article{mousavi2020earthquake,
  title={Earthquake transformer—an attentive deep-learning model for simultaneous earthquake detection and phase picking},
  author={Mousavi, S Mostafa and Ellsworth, William L and Zhu, Weiqiang and Chuang, Lindsay Y and Beroza, Gregory C},
  journal={Nature communications},
  volume={11},
  number={1},
  pages={3952},
  year={2020},
  publisher={Nature Publishing Group UK London}
}

@article{bao2022application,
  title={Application of transformer models to landslide susceptibility mapping},
  author={Bao, Shuai and Liu, Jiping and Wang, Liang and Zhao, Xizhi},
  journal={Sensors},
  volume={22},
  number={23},
  pages={9104},
  year={2022},
  publisher={MDPI}
}

@article{wang2023seismic,
  title={Seismic facies segmentation via a segformer-based specific encoder--decoder--hypercolumns scheme},
  author={Wang, Zhiguo and Wang, Qiannan and Yang, Yang and Liu, Naihao and Chen, Yumin and Gao, Jinghuai},
  journal={IEEE Transactions on Geoscience and Remote Sensing},
  volume={61},
  pages={1--11},
  year={2023},
  publisher={IEEE}
}

@article{nguyen2024scaling,
  title={Scaling transformer neural networks for skillful and reliable medium-range weather forecasting},
  author={Nguyen, Tung and Shah, Rohan and Bansal, Hritik and Arcomano, Troy and Maulik, Romit and Kotamarthi, Rao and Foster, Ian and Madireddy, Sandeep and Grover, Aditya},
  journal={Advances in Neural Information Processing Systems},
  volume={37},
  pages={68740--68771},
  year={2024}
}

@article{lu2019vilbert,
  title={Vilbert: Pretraining task-agnostic visiolinguistic representations for vision-and-language tasks},
  author={Lu, Jiasen and Batra, Dhruv and Parikh, Devi and Lee, Stefan},
  journal={Advances in neural information processing systems},
  volume={32},
  year={2019}
}

@article{alayrac2022flamingo,
  title={Flamingo: a visual language model for few-shot learning},
  author={Alayrac, Jean-Baptiste and Donahue, Jeff and Luc, Pauline and Miech, Antoine and Barr, Iain and Hasson, Yana and Lenc, Karel and Mensch, Arthur and Millican, Katherine and Reynolds, Malcolm and others},
  journal={Advances in neural information processing systems},
  volume={35},
  pages={23716--23736},
  year={2022}
}

@inproceedings{radford2021learning,
  title={Learning transferable visual models from natural language supervision},
  author={Radford, Alec and Kim, Jong Wook and Hallacy, Chris and Ramesh, Aditya and Goh, Gabriel and Agarwal, Sandhini and Sastry, Girish and Askell, Amanda and Mishkin, Pamela and Clark, Jack and others},
  booktitle={International conference on machine learning},
  pages={8748--8763},
  year={2021},
  organization={PmLR}
}

@inproceedings{li2023blip,
  title={Blip-2: Bootstrapping language-image pre-training with frozen image encoders and large language models},
  author={Li, Junnan and Li, Dongxu and Savarese, Silvio and Hoi, Steven},
  booktitle={International conference on machine learning},
  pages={19730--19742},
  year={2023},
  organization={PMLR}
}

@article{wang2024av,
  title={Av-dit: Efficient audio-visual diffusion transformer for joint audio and video generation},
  author={Wang, Kai and Deng, Shijian and Shi, Jing and Hatzinakos, Dimitrios and Tian, Yapeng},
  journal={arXiv preprint arXiv:2406.07686},
  year={2024}
}

@inproceedings{lin2020audiovisual,
  title={Audiovisual transformer with instance attention for audio-visual event localization},
  author={Lin, Yan-Bo and Wang, Yu-Chiang Frank},
  booktitle={Proceedings of the Asian Conference on Computer Vision},
  year={2020}
}

@article{tan2019lxmert,
  title={Lxmert: Learning cross-modality encoder representations from transformers},
  author={Tan, Hao and Bansal, Mohit},
  journal={arXiv preprint arXiv:1908.07490},
  year={2019}
}

@article{zhang2023meta,
  title={Meta-transformer: A unified framework for multimodal learning},
  author={Zhang, Yiyuan and Gong, Kaixiong and Zhang, Kaipeng and Li, Hongsheng and Qiao, Yu and Ouyang, Wanli and Yue, Xiangyu},
  journal={arXiv preprint arXiv:2307.10802},
  year={2023}
}

@inproceedings{radford2023robust,
  title={Robust speech recognition via large-scale weak supervision},
  author={Radford, Alec and Kim, Jong Wook and Xu, Tao and Brockman, Greg and McLeavey, Christine and Sutskever, Ilya},
  booktitle={International conference on machine learning},
  pages={28492--28518},
  year={2023},
  organization={PMLR}
}

@article{yuunified,
  title={Unified Material Transformer as Scalable Material Property Predictor},
  author={Yu, Juncheng and Zhang, Kaiwei and Li, Haonan}
}

@article{wei2204crystal,
  title={Crystal Transformer: Self-learning Neural Language Model for Generative and Tinkering Design of Materials, arXiv, 2022},
  author={Wei, L and Li, Q and Song, Y and Stefanov, S and Siriwardane, E and Chen, F and Hu, J},
  journal={arXiv preprint arXiv:2204.11953}
}

@article{chen2024mattergpt,
  title={MatterGPT: A generative transformer for multi-property inverse design of solid-state materials},
  author={Chen, Yan and Wang, Xueru and Deng, Xiaobin and Liu, Yilun and Chen, Xi and Zhang, Yunwei and Wang, Lei and Xiao, Hang},
  journal={arXiv preprint arXiv:2408.07608},
  year={2024}
}

@article{chen2024interpretable,
  title={An interpretable and transferrable vision transformer model for rapid materials spectra classification},
  author={Chen, Zhenru and Xie, Yunchao and Wu, Yuchao and Lin, Yuyi and Tomiya, Shigetaka and Lin, Jian},
  journal={Digital Discovery},
  volume={3},
  number={2},
  pages={369--380},
  year={2024},
  publisher={Royal Society of Chemistry}
}

@article{yang2021words,
  title={Words to matter: De novo architected materials design using transformer neural networks},
  author={Yang, Zhenze and Buehler, Markus J},
  journal={Frontiers in Materials},
  volume={8},
  pages={740754},
  year={2021},
  publisher={Frontiers Media SA}
}

@article{peng2022attention,
  title={Attention-enhanced neural network models for turbulence simulation},
  author={Peng, Wenhui and Yuan, Zelong and Wang, Jianchun},
  journal={Physics of Fluids},
  volume={34},
  number={2},
  year={2022},
  publisher={AIP Publishing}
}

@article{alkin2024universal,
  title={Universal physics transformers: A framework for efficiently scaling neural operators},
  author={Alkin, Benedikt and F{\"u}rst, Andreas and Schmid, Simon and Gruber, Lukas and Holzleitner, Markus and Brandstetter, Johannes},
  journal={Advances in Neural Information Processing Systems},
  volume={37},
  pages={25152--25194},
  year={2024}
}

@article{herde2024poseidon,
  title={Poseidon: Efficient foundation models for pdes},
  author={Herde, Maximilian and Raonic, Bogdan and Rohner, Tobias and K{\"a}ppeli, Roger and Molinaro, Roberto and de B{\'e}zenac, Emmanuel and Mishra, Siddhartha},
  journal={Advances in Neural Information Processing Systems},
  volume={37},
  pages={72525--72624},
  year={2024}
}

@article{zheng2024aerodit,
  title={AeroDiT: Diffusion Transformers for Reynolds-Averaged Navier-Stokes Simulations of Airfoil Flows},
  author={Zheng, Hao and Dai, Zhibo and Pan, Biyue and Wang, Chunyang and Zhang, Baiyi and Xiang, Hui and Fan, Dixia},
  journal={arXiv preprint arXiv:2412.17394},
  year={2024}
}

@inproceedings{xu2025amr,
  title={AMR-Transformer: Enabling Efficient Long-range Interaction for Complex Neural Fluid Simulation},
  author={Xu, Zeyi and Liu, Jinfan and Chen, Kuangxu and Chen, Ye and Hu, Zhangli and Ni, Bingbing},
  booktitle={Proceedings of the Computer Vision and Pattern Recognition Conference},
  pages={5804--5813},
  year={2025}
}

@article{kang2023new,
  title={A new fluid flow approximation method using a vision transformer and a U-shaped convolutional neural network},
  author={Kang, Hyoeun and Kim, Yongsu and Le, Thi-Thu-Huong and Choi, Changwoo and Hong, Yoonyoung and Hong, Seungdo and Chin, Sim Won and Kim, Howon},
  journal={AIP Advances},
  volume={13},
  number={2},
  year={2023},
  publisher={AIP Publishing}
}

@article{chowdhery2023palm,
  title={Palm: Scaling language modeling with pathways},
  author={Chowdhery, Aakanksha and Narang, Sharan and Devlin, Jacob and Bosma, Maarten and Mishra, Gaurav and Roberts, Adam and Barham, Paul and Chung, Hyung Won and Sutton, Charles and Gehrmann, Sebastian and others},
  journal={Journal of Machine Learning Research},
  volume={24},
  number={240},
  pages={1--113},
  year={2023}
}

@article{touvron2023llama,
  title={Llama: Open and efficient foundation language models},
  author={Touvron, Hugo and Lavril, Thibaut and Izacard, Gautier and Martinet, Xavier and Lachaux, Marie-Anne and Lacroix, Timoth{\'e}e and Rozi{\`e}re, Baptiste and Goyal, Naman and Hambro, Eric and Azhar, Faisal and others},
  journal={arXiv preprint arXiv:2302.13971},
  year={2023}
}

@article{achiam2023gpt,
  title={Gpt-4 technical report},
  author={Achiam, Josh and Adler, Steven and Agarwal, Sandhini and Ahmad, Lama and Akkaya, Ilge and Aleman, Florencia Leoni and Almeida, Diogo and Altenschmidt, Janko and Altman, Sam and Anadkat, Shyamal and others},
  journal={arXiv preprint arXiv:2303.08774},
  year={2023}
}

@article{dosovitskiy2020image,
  title={An image is worth 16x16 words: Transformers for image recognition at scale},
  author={Dosovitskiy, Alexey and Beyer, Lucas and Kolesnikov, Alexander and Weissenborn, Dirk and Zhai, Xiaohua and Unterthiner, Thomas and Dehghani, Mostafa and Minderer, Matthias and Heigold, Georg and Gelly, Sylvain and others},
  journal={arXiv preprint arXiv:2010.11929},
  year={2020}
}

@inproceedings{liu2021swin,
  title={Swin transformer: Hierarchical vision transformer using shifted windows},
  author={Liu, Ze and Lin, Yutong and Cao, Yue and Hu, Han and Wei, Yixuan and Zhang, Zheng and Lin, Stephen and Guo, Baining},
  booktitle={Proceedings of the IEEE/CVF international conference on computer vision},
  pages={10012--10022},
  year={2021}
}

@inproceedings{he2022masked,
  title={Masked autoencoders are scalable vision learners},
  author={He, Kaiming and Chen, Xinlei and Xie, Saining and Li, Yanghao and Doll{\'a}r, Piotr and Girshick, Ross},
  booktitle={Proceedings of the IEEE/CVF conference on computer vision and pattern recognition},
  pages={16000--16009},
  year={2022}
}

@article{baevski2020wav2vec,
  title={wav2vec 2.0: A framework for self-supervised learning of speech representations},
  author={Baevski, Alexei and Zhou, Yuhao and Mohamed, Abdelrahman and Auli, Michael},
  journal={Advances in neural information processing systems},
  volume={33},
  pages={12449--12460},
  year={2020}
}

@article{hsu2021hubert,
  title={Hubert: Self-supervised speech representation learning by masked prediction of hidden units},
  author={Hsu, Wei-Ning and Bolte, Benjamin and Tsai, Yao-Hung Hubert and Lakhotia, Kushal and Salakhutdinov, Ruslan and Mohamed, Abdelrahman},
  journal={IEEE/ACM transactions on audio, speech, and language processing},
  volume={29},
  pages={3451--3460},
  year={2021},
  publisher={IEEE}
}

@article{gong2021ast,
  title={Ast: Audio spectrogram transformer},
  author={Gong, Yuan and Chung, Yu-An and Glass, James},
  journal={arXiv preprint arXiv:2104.01778},
  year={2021}
}

@article{chen2022beats,
  title={Beats: Audio pre-training with acoustic tokenizers},
  author={Chen, Sanyuan and Wu, Yu and Wang, Chengyi and Liu, Shujie and Tompkins, Daniel and Chen, Zhuo and Wei, Furu},
  journal={arXiv preprint arXiv:2212.09058},
  year={2022}
}

@article{sapkota2025ai,
  title={Ai agents vs. agentic ai: A conceptual taxonomy, applications and challenges},
  author={Sapkota, Ranjan and Roumeliotis, Konstantinos I and Karkee, Manoj},
  journal={arXiv preprint arXiv:2505.10468},
  year={2025}
}

@article{hou2025model,
  title={Model context protocol (mcp): Landscape, security threats, and future research directions},
  author={Hou, Xinyi and Zhao, Yanjie and Wang, Shenao and Wang, Haoyu},
  journal={arXiv preprint arXiv:2503.23278},
  year={2025}
}

@article{li2022long,
  title={A Long-term Dependent and Trustworthy Approach to Reactor Accident Prognosis based on Temporal Fusion Transformer},
  author={Li, Chengyuan and Qiu, Zhifang and Ma, Yugao and Li, Meifu},
  journal={arXiv preprint arXiv:2210.17298},
  year={2022}
}

@article{chen2023leveraging,
  title={Leveraging Neural Networks With Attention Mechanism for High-Order Accuracy in Charge Density in Particle-in-Cell Simulation},
  author={Chen, Jian-Nan and Zhang, Jun-Jie},
  journal={arXiv preprint arXiv:2311.14972},
  year={2023}
}

@article{spangher2023autoregressive,
  title={Autoregressive transformers for disruption prediction in nuclear fusion plasmas},
  author={Spangher, Lucas and Arnold, William and Spangher, Alexander and Maris, Andrew and Rea, Cristina},
  journal={arXiv preprint arXiv:2401.00051},
  year={2023}
}

@article{lyu2024nrformer,
  title={NRFormer: Nationwide Nuclear Radiation Forecasting with Spatio-Temporal Transformer},
  author={Lyu, Tengfei and Han, Jindong and Liu, Hao},
  journal={arXiv preprint arXiv:2410.11924},
  year={2024}
}

@inproceedings{conneau2020unsupervised,
  title={Unsupervised cross-lingual representation learning at scale},
  author={Conneau, Alexis and Khandelwal, Kartikay and Goyal, Naman and Chaudhary, Vishrav and Wenzek, Guillaume and Guzm{\'a}n, Francisco and Grave, Edouard and Ott, Myle and Zettlemoyer, Luke and Stoyanov, Veselin},
  booktitle={Proceedings of the 58th annual meeting of the association for computational linguistics},
  pages={8440--8451},
  year={2020}
}

@article{grattafiori2024llama,
  title={The llama 3 herd of models},
  author={Grattafiori, Aaron and Dubey, Abhimanyu and Jauhri, Abhinav and Pandey, Abhinav and Kadian, Abhishek and Al-Dahle, Ahmad and Letman, Aiesha and Mathur, Akhil and Schelten, Alan and Vaughan, Alex and others},
  journal={arXiv preprint arXiv:2407.21783},
  year={2024}
}

@article{wu2016google,
  title={Google's neural machine translation system: Bridging the gap between human and machine translation},
  author={Wu, Yonghui and Schuster, Mike and Chen, Zhifeng and Le, Quoc V and Norouzi, Mohammad and Macherey, Wolfgang and Krikun, Maxim and Cao, Yuan and Gao, Qin and Macherey, Klaus and others},
  journal={arXiv preprint arXiv:1609.08144},
  year={2016}
}

@article{liu2021deep,
  title={Deep learning for procedural content generation},
  author={Liu, Jialin and Snodgrass, Sam and Khalifa, Ahmed and Risi, Sebastian and Yannakakis, Georgios N and Togelius, Julian},
  journal={Neural Computing and Applications},
  volume={33},
  number={1},
  pages={19--37},
  year={2021},
  publisher={Springer}
}

@inproceedings{mcmahan2017communication,
  title={Communication-efficient learning of deep networks from decentralized data},
  author={McMahan, Brendan and Moore, Eider and Ramage, Daniel and Hampson, Seth and y Arcas, Blaise Aguera},
  booktitle={Artificial intelligence and statistics},
  pages={1273--1282},
  year={2017},
  organization={Pmlr}
}

@inproceedings{dwork2008differential,
  title={Differential privacy: A survey of results},
  author={Dwork, Cynthia},
  booktitle={International conference on theory and applications of models of computation},
  pages={1--19},
  year={2008},
  organization={Springer}
}

@inproceedings{gentry2009fully,
  title={Fully homomorphic encryption using ideal lattices},
  author={Gentry, Craig},
  booktitle={Proceedings of the forty-first annual ACM symposium on Theory of computing},
  pages={169--178},
  year={2009}
}

@article{ferdaus2026towards,
  title={Towards trustworthy AI: a review of ethical and robust large language models},
  author={Ferdaus, Md Meftahul and Abdelguerfi, Mahdi and Loup, Elias and N. Niles, Kendall and Pathak, Ken and Sloan, Steven},
  journal={ACM Computing Surveys},
  volume={58},
  number={7},
  pages={1--43},
  year={2026},
  publisher={ACM New York, NY}
}

@inproceedings{adhikari2020nlp,
  title={Nlp based machine learning approaches for text summarization},
  author={Adhikari, Surabhi and others},
  booktitle={2020 Fourth International Conference on Computing Methodologies and Communication (ICCMC)},
  pages={535--538},
  year={2020},
  organization={IEEE}
}

@article{neerudu2023robustness,
  title={On robustness of finetuned transformer-based nlp models},
  author={Neerudu, Pavan Kalyan Reddy and Oota, Subba Reddy and Marreddy, Mounika and Kagita, Venkateswara Rao and Gupta, Manish},
  journal={arXiv preprint arXiv:2305.14453},
  year={2023}
}

@article{gehman2020realtoxicityprompts,
  title={Realtoxicityprompts: Evaluating neural toxic degeneration in language models},
  author={Gehman, Samuel and Gururangan, Suchin and Sap, Maarten and Choi, Yejin and Smith, Noah A},
  journal={arXiv preprint arXiv:2009.11462},
  year={2020}
}

@article{li2024seeing,
  title={Seeing the forest through the trees: Data leakage from partial transformer gradients},
  author={Li, Weijun and Xu, Qiongkai and Dras, Mark},
  journal={arXiv preprint arXiv:2406.00999},
  year={2024}
}

@article{marks2023geometry,
  title={The geometry of truth: Emergent linear structure in large language model representations of true/false datasets},
  author={Marks, Samuel and Tegmark, Max},
  journal={arXiv preprint arXiv:2310.06824},
  year={2023}
}

@article{lambert2025reinforcement,
  title={Reinforcement learning from human feedback},
  author={Lambert, Nathan},
  journal={arXiv preprint arXiv:2504.12501},
  year={2025}
}

@article{bai2022constitutional,
  title={Constitutional ai: Harmlessness from ai feedback},
  author={Bai, Yuntao and Kadavath, Saurav and Kundu, Sandipan and Askell, Amanda and Kernion, Jackson and Jones, Andy and Chen, Anna and Goldie, Anna and Mirhoseini, Azalia and McKinnon, Cameron and others},
  journal={arXiv preprint arXiv:2212.08073},
  year={2022}
}

@article{dafoe2021cooperative,
  title={Cooperative AI: machines must learn to find common ground},
  author={Dafoe, Allan and Bachrach, Yoram and Hadfield, Gillian and Horvitz, Eric and Larson, Kate and Graepel, Thore},
  journal={Nature},
  volume={593},
  number={7857},
  pages={33--36},
  year={2021},
  publisher={Nature Publishing Group UK London}
}

@inproceedings{devlin2019bert,
  title={Bert: Pre-training of deep bidirectional transformers for language understanding},
  author={Devlin, Jacob and Chang, Ming-Wei and Lee, Kenton and Toutanova, Kristina},
  booktitle={Proceedings of the 2019 conference of the North American chapter of the association for computational linguistics: human language technologies, volume 1 (long and short papers)},
  pages={4171--4186},
  year={2019}
}

@article{brown2020language,
  title={Language models are few-shot learners},
  author={Brown, Tom and Mann, Benjamin and Ryder, Nick and Subbiah, Melanie and Kaplan, Jared D and Dhariwal, Prafulla and Neelakantan, Arvind and Shyam, Pranav and Sastry, Girish and Askell, Amanda and others},
  journal={Advances in neural information processing systems},
  volume={33},
  pages={1877--1901},
  year={2020}
}

@article{trinh2024solving,
  title={Solving olympiad geometry without human demonstrations},
  author={Trinh, Trieu H and Wu, Yuhuai and Le, Quoc V and He, He and Luong, Thang},
  journal={Nature},
  volume={625},
  number={7995},
  pages={476--482},
  year={2024},
  publisher={Nature Publishing Group UK London}
}

@article{chervonyi2025gold,
  title={Gold-medalist performance in solving olympiad geometry with alphageometry2},
  author={Chervonyi, Yuri and Trinh, Trieu H and Ol{\v{s}}{\'a}k, Miroslav and Yang, Xiaomeng and Nguyen, Hoang H and Menegali, Marcelo and Jung, Junehyuk and Kim, Junsu and Verma, Vikas and Le, Quoc V and others},
  journal={Journal of Machine Learning Research},
  volume={26},
  number={241},
  pages={1--39},
  year={2025}
}

@article{hubert2025olympiad,
  title={Olympiad-level formal mathematical reasoning with reinforcement learning},
  author={Hubert, Thomas and Mehta, Rishi and Sartran, Laurent and Horv{\'a}th, Mikl{\'o}s Z and {\v{Z}}u{\v{z}}i{\'c}, Goran and Wieser, Eric and Huang, Aja and Schrittwieser, Julian and Schroecker, Yannick and Masoom, Hussain and others},
  journal={Nature},
  pages={1--3},
  year={2025},
  publisher={Nature Publishing Group UK London}
}

@article{li2020bert,
  title={Bert-attack: Adversarial attack against bert using bert},
  author={Li, Linyang and Ma, Ruotian and Guo, Qipeng and Xue, Xiangyang and Qiu, Xipeng},
  journal={arXiv preprint arXiv:2004.09984},
  year={2020}
}

@inproceedings{jin2020bert,
  title={Is bert really robust? a strong baseline for natural language attack on text classification and entailment},
  author={Jin, Di and Jin, Zhijing and Zhou, Joey Tianyi and Szolovits, Peter},
  booktitle={Proceedings of the AAAI conference on artificial intelligence},
  volume={34},
  number={05},
  pages={8018--8025},
  year={2020}
}

\end{document}